\documentclass{article} 
\usepackage{iclr2027_conference,times}
\usepackage{url}
\usepackage[utf8]{inputenc}
\usepackage[T1]{fontenc}
\usepackage{url}
\usepackage{booktabs}
\usepackage{siunitx}
\usepackage{amsfonts}
\usepackage{nicefrac}
\usepackage{microtype}
\usepackage[dvipsnames]{xcolor}
\usepackage{wrapfig}
\usepackage{tikz}
\usepackage{multicol,multirow}
\usepackage{subcaption}
\usepackage{graphicx}

\usepackage{enumitem}
\usepackage[svgnames,x11names]{xcolor}
\definecolor{tabblue}{HTML}{1F77B4}
\definecolor{taborange}{HTML}{FF7F0E}
\definecolor{tabgreen}{HTML}{2CA02C}
\definecolor{tabred}{HTML}{D62728}
\definecolor{tabpurple}{HTML}{9467BD}
\definecolor{tabbrown}{HTML}{8C564B}
\definecolor{tabpink}{HTML}{E377C2}
\definecolor{tabolive}{HTML}{BCBD22}
\definecolor{tabcyan}{HTML}{17BECF}
\definecolor{darkorange}{HTML}{FF8C00}
\definecolor{brown}{HTML}{663399}
\definecolor{maroon}{HTML}{AF3235}
\definecolor{srmcolor}{HTML}{D85A30}
\definecolor{cluecolor}{HTML}{534AB7}
\definecolor{solutioncolor}{HTML}{1D9E75}
\definecolor{baselinecolor}{HTML}{888780}
\definecolor{RoyalBlue}{rgb}{0,0.22,0.66}
\usepackage{makecell}
\usepackage{lipsum}
\usepackage{hyperref}
\hypersetup{
colorlinks=true,
linkcolor=maroon,
anchorcolor=RoyalBlue,
citecolor=RoyalBlue,
filecolor=RoyalBlue,
urlcolor=magenta
}

\newcommand{\colorbullet}[1]{\textcolor{#1}{$\bullet$}}
\newcommand*{\parallelogramm}{%
  \rlap{\rotatebox{-30}{\rule[.05ex]{.4pt}{.77em}}}%
  \kern.04em%
  \rlap{\kern.36em\raisebox{0.649519052835em}{\rule{.6em}{.4pt}}}%
  \rule{.6em}{.4pt}\kern-.04em%
  \rotatebox{-30}{\rule[.05ex]{.4pt}{.77em}}}
\usepackage{amsmath, amssymb, amsthm, mathtools}
\usepackage{multicol, multirow}
\usepackage{fontawesome5, wasysym, pifont}
\usepackage{svg}
\usepackage{relsize} 
\newcommand{\snowflake}{\raisebox{-0.2ex}{\includegraphics[height=1.7ex]{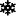}}}
\usepackage{pifont}
\newcommand{\cmark}{\ding{51}}%
\newcommand{\xmark}{\ding{55}}%
\usepackage[ruled,vlined]{algorithm2e} 
\SetKwComment{Comment}{}{} 
\newcommand{\sidecomment}[1]{\Comment{\footnotesize\ttfamily\textcolor{gray}{$\;\;$\##1}}}
\newcommand{\reported}[1]{\textcolor{darkgray}{(\textit{#1})}}

\usepackage{xspace}
\newcommand{\Tangram}{\text{Tangram}\xspace}
\newcommand{\Coldoku}{\text{Coldoku}\xspace}
\newcommand{\Tildoku}{\text{Tildoku}\xspace}
\newcommand{\LogicFace}{\text{LogicFace}\xspace}
\newcommand{\Akari}{\text{Akari}\xspace}
\newcommand{\bulb}[1][1em]{\raisebox{-0.3ex}{\includegraphics[height=#1]{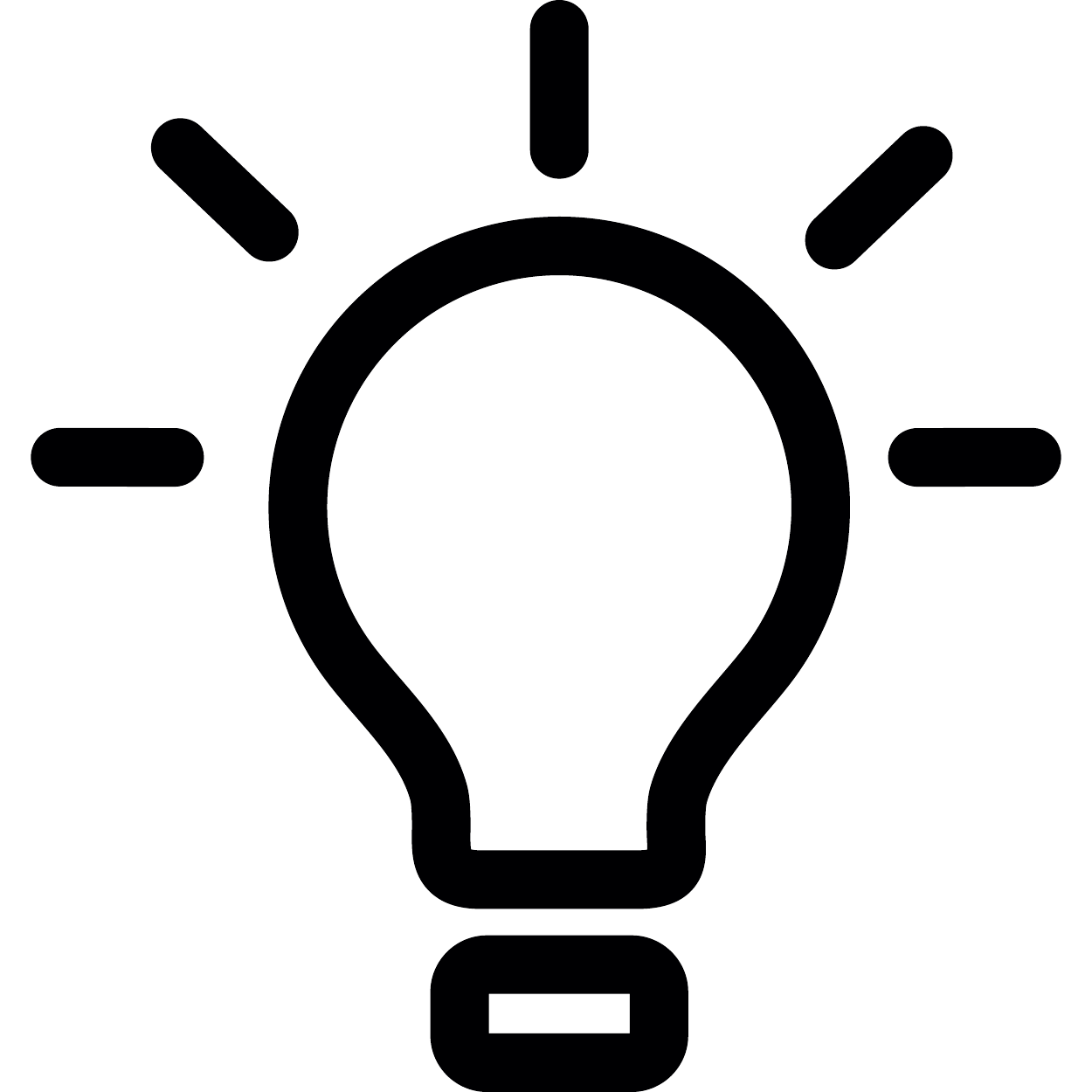}}\xspace}
\newcommand{\relcube}[1][1em]{\raisebox{-0.4ex}{\includegraphics[height=#1]{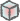}}\xspace}

\definecolor{rscolor}{HTML}{990000}
\definecolor{dscolor}{HTML}{0066CC}

\definecolor{lightyellow}{HTML}{FFF59D}
\newcommand{\light}{
    \raisebox{0.6ex}{\hspace{-2pt}\fcolorbox{black}{lightyellow}{\rule{0pt}{1pt}\rule{1pt}{0pt}}\hspace{-2pt}}
}
\newcommand{\white}{
    \raisebox{0.6ex}{\hspace{-2pt}\fcolorbox{black}{white}{\rule{0pt}{1pt}\rule{1pt}{0pt}}\hspace{-2pt}}
}
\newcommand{\wall}{
    \raisebox{0.6ex}{\hspace{-2pt}\fcolorbox{black}{black}{\rule{0pt}{1pt}\rule{1pt}{0pt}}\hspace{-2pt}}
}

\title{Relational Abstractions for Spatial\\Reasoning with Diffusion Models}

\author{Ana Ezquerro, Ozan Özdenizci \\ Institute of Machine Learning and Neural Computation \\ Graz University of Technology, Austria \\ \texttt{\{ana.ezquerro,oezdenizci\}@tugraz.at}}
\iclrfinalcopy
\begin{document}

\maketitle

\begin{abstract}
Diffusion models excel at image synthesis, but they remain limited in their ability to reliably satisfy structured spatial reasoning constraints.
In conditional data distribution modeling tasks with implicit logical structure, such as puzzles defined by visible clues paired with consistent solutions, state-of-the-art generative models tend to approximate pixel-space distributions without learning the underlying logical rules required for inference.
To address this limitation, we present a novel framework for spatial reasoning with diffusion models that leverages unsupervised object discovery and abstractions of object relations.
We show that the relational knowledge derived from object-centric representations enriches diffusion models with structural primitives, allowing them to effectively guide the generative representation space during both training and inference, and enabling conditional image generation that satisfies reasoning constraints.
Additionally, we introduce a large-scale generative spatial reasoning benchmark with four datasets inspired by human-solvable puzzles.
Our results show that relational abstractions significantly improve reasoning capabilities of diffusion models on a variety of complex reasoning tasks, while enabling robust generalization in out-of-distribution settings\footnote{Code and datasets are available at: \href{https://github.com/anaezquerro/rdm}{https://github.com/anaezquerro/rdm}.}.
\end{abstract}

\section{Introduction}

Diffusion models, including denoising-based~\citep{sohl2015deep,ho-etal-2020-denoising}, score-based~\citep{song-etal-2020-generative,song-etal-2021-score}, and flow-based formulations~\citep{lipman-etal-2023-flow}, have emerged as powerful generative paradigms for learning and high-fidelity sampling from complex data distributions with remarkable success.
Importantly, these models implicitly capture structural and semantic regularities of data, suggesting a promising foundation for reasoning in continuous domains.
While reasoning in language has been extensively explored with large language models, how diffusion models can perform spatial reasoning remains an open question.

Spatial reasoning requires detecting visual patterns, inducing hypotheses, and systematically applying rules to solve complex problems \citep{raven-1983-progressive,fleuret-etal-2011-comparing}. 
Building on the success of deep learning on image data, prior work has explored neural approaches to Abstract Visual Reasoning, where models are trained on tasks such as Raven's Progressive Matrices (RPM)~\citep{barret-etal-2018-measuring} and Bongard-style problems~\citep{nie-etal-2020-bongard}. 
These methods traditionally frame visual reasoning as a discriminative task, requiring models to make a categorical choice rather than synthesizing a solution.
This formulation has been criticized for allowing models to exploit superficial visual heuristics rather than demonstrating genuine, systematic rule inference~\citep{webb-etal-2021-emergent,ye-etal-2025-blinktwice}.
To address this limitation, recent work has shifted toward framing spatial reasoning as a generative problem.
The ARC challenge~\citep{chollet-etal-2019-arc} in particular, requires models to synthesize a complete output grid from scratch.
However, existing approaches typically rely on symbolic program synthesis~\citep{chollet2024arc}, while their application to pixel-level image synthesis remains challenging.
More recently, diffusion models have been explored in structured visual reasoning tasks, such as Sudoku-style image generation, with a particular focus on adapting inference-time sampling strategies to improve solution consistency~\citep{wever-etal-2025-spatial,lee-etal-2025-adaptive}.
In contrast, we argue that effective generative reasoning over complex visual domains requires moving beyond refined pixel-level reconstruction to incorporate higher-level context abstractions within the generation process.

\textbf{Contributions:} 
We introduce RDM (\textit{\underline{R}elational Abstractions for \underline{D}iffusion \underline{M}odels}), a novel framework for spatial reasoning with diffusion models that leverages unsupervised object discovery via Slot Attention~\citep{locatello-etal-2020-object} and abstractions of object relations.
We show that the relational knowledge derived from object-centric embeddings via Abstractors~\citep{mondal-etal-2024-slot,altabaa-etal-2024-abstractors} enhances diffusion models with contextual primitives for both training and inference, enabling high-fidelity image synthesis while adhering to the reasoning constraints.
We introduce a large-scale benchmark consisting of four new challenging datasets inspired by popular human-solvable puzzles for studying spatial reasoning as a conditional generative modeling task, where underlying logical rules must be inferred from training data.
Each dataset contains up to one million paired examples, covering grid-based image completion and multi-context image generation tasks across multiple levels of abstraction.
Our approach improves performance on a variety of complex reasoning tasks, by up to 3$\times$ on the challenging multi-context reasoning problems, while demonstrating strong out-of-distribution generalization.

\section{Related Work}\label{sec:related-work}

\textbf{Diffusion models:} Diffusion models~\citep{ho-etal-2020-denoising,song-etal-2021-score} learn to model a data distribution $q_0(\mathbf{x})$ via a forward process that progressively transforms it into a Gaussian prior $q_1=\mathcal{N}(0,\mathbf{I})$, and train a neural network $p_{\theta}$ to reverse this transformation.
The forward process can be formulated either as a discrete-time Markov chain of Gaussian noising transitions, where the reverse process utilizes a neural network that predicts the added noise at each step $t$ as in denoising models~\citep{ho-etal-2020-denoising}, or as a continuous-time stochastic differential equation (SDE) where noise is injected continuously~\citep{song-etal-2021-score} and the reverse-time SDE dynamics are governed by the score function $\nabla_{\mathbf{x}}\log q_t(\mathbf{x})$, which is estimated by a neural network.

In the continuous-time setting, an alternative formulation based on ordinary differential equations (ODEs) leads to \emph{continuous normalizing flows} (CNFs), which define the forward process via conditional probability paths $p_t(\mathbf{x}\,|\,\mathbf{x}_0)$ for samples $\mathbf{x}_0\sim q_0(\mathbf{x})$.
Specifically, Gaussian paths $p_t(\mathbf{x}\,|\,\mathbf{x}_0)=\mathcal{N}(\mathbf{x}\,|\,\mu_t(\mathbf{x}_0), \sigma_t(\mathbf{x}_0)^2 \mathbf{I})$ across time $t\in[0,1]$ interpolate between the data distribution and a Gaussian prior by setting $\mu_0(\mathbf{x}_0)=\mathbf{x}_0$, $\sigma_0(\mathbf{x}_0)=\sigma_{\min}$, $\mu_1(\mathbf{x}_0)=0$ and $\sigma_1(\mathbf{x}_0)=1$.
Samples along this path can be expressed via the \emph{flow} $\phi_t(\boldsymbol{\epsilon},\mathbf{x}_0)=\mu_t(\mathbf{x}_0) + \sigma_t(\mathbf{x}_0)\boldsymbol{\epsilon}$, a differentiable transformation that maps noise $\boldsymbol{\epsilon}\sim \mathcal{N}(0,\mathbf{I})$ to samples from $p_t(\mathbf{x}\,|\,\mathbf{x}_0)$.
The corresponding conditional \emph{vector field} $v_t(\mathbf{x}\,|\,\mathbf{x}_0) = \sigma'_t(\mathbf{x}_0)/\sigma_t(\mathbf{x}_0)(\mathbf{x}-\mu_t(\mathbf{x}_0)) + \mu'_t(\mathbf{x}_0)$ is approximated by a neural network $p_\theta(\mathbf{x}_t,t)$.
In our experiments, we focus on CNFs trained via \textit{flow matching}~\citep{lipman-etal-2023-flow} (see \S\ref{ap:reasoner_config} for details), although our framework is applicable to all families of diffusion models.

\textbf{Spatial reasoning with generative models:}
Several works explore generative models for visual reasoning.
In RPM, this requires synthesizing a missing panel from scratch, requiring complete understanding of the underlying relational rules.
Earlier solutions primarily used deep latent variable models~\citep{pekar-etal-2020-generating} with structured latent spaces~\citep{shi-etal-2024-towards}, to abstract explicit rules on concepts.
However, these solvers often require handcrafted inductive biases rooted in RPM-specific rules to facilitate abstraction and panel generation.
More recently, diffusion models have been explored for spatial reasoning with sampling-time solutions~\citep{wever-etal-2025-spatial,lee-etal-2025-adaptive}.
Notably, \citet{wever-etal-2025-spatial} proposed SRM based on diffusion forcing~\citep{chen-etal-2024-diffusion-forcing}, to perform spatial reasoning as conditional image generation.
It proposes an incremental sampling strategy based on patch-level uncertainty estimation, exploiting the grid structure of Sudoku-style tasks with state-of-the-art performance, but is inherently tailored to structured grids.

\textbf{Object discovery:}
Object discovery is rooted in the hypothesis that human intelligence decomposes complex visual scenes into discrete, interacting entities~\citep{battaglia-etal-2018-relational,greff-etal-2019-multi}, and studies aim to instill a similar inductive bias into neural networks by moving beyond dense feature maps to interpretable object representations.
\citet{locatello-etal-2020-object} aligned this concept into modern neural networks via \textit{Slot Attention}, a framework that disentangles the feature and positional information of visual components into high-dimensional embeddings, named \emph{slots}, representing the objects captured in the scene.
Object discovery has been also adopted for discriminative reasoning tasks, where relational modules aggregate information across disentangled slots to model object interactions.
Methods such as CoRelNet~\citep{kerg-etal-2022-corelnet}, OCRA~\citep{webb2023systematic} and Abstractors~\citep{altabaa-etal-2024-abstractors} leverage object-centric designs in which learned relations condition a classification head, exploiting the ability of slots to disentangle scene factors under discriminative objectives. 
Recent object-centric diffusion models~\citep{wu-etal-2023-slotdiffusion,jiang2023object} extend these to generative settings, with a focus on downstream vision tasks and compositional image synthesis capabilities.

\section{Relational Abstractions for Diffusion Models}
\label{sec:methodology}

We formalize the spatial reasoning task as modeling the conditional distribution $q(\mathbf{x}_0|\mathbf{x}_c)$, where $\mathbf{x}_c$ denotes a provided problem clue, and $\mathbf{x}_0$ represents a valid solution satisfying all logical constraints.

Our proposed framework comprises two models.
Our first model, the \emph{discoverer}, is a latent variable model $p_\phi$ designed to capture the objects present in the samples from $q(\mathbf{x}^\prime)$, where $\mathbf{x}^\prime\in\{\mathbf{x}_0,\mathbf{x}_c\}$, via disentangled latent representations $\mathbf{z}^\prime$.
Our second model, the \emph{reasoner}, is a diffusion model $p_\theta$ that learns to reverse the trajectory from a Gaussian prior to the target distribution $q(\mathbf{x}_0|\mathbf{x}_c)$, conditioned on the latents $\mathbf{z}^\prime$ provided by the discoverer.
The integration of object-centric latent variables $\mathbf{z}^\prime$ is motivated by the inherent difficulty of learning $q(\mathbf{x}_0|\mathbf{x}_c)$ through a simple parameterization of $p_\theta$.
We propose to enrich the generative modeling process with relational abstractions inferred from object-centric representations obtained via an unsupervised object discovery model $p_\phi(\mathbf{z}^\prime|\mathbf{x}^\prime)$. 

\begin{figure}[t!]\centering
    \includegraphics[width=\linewidth]{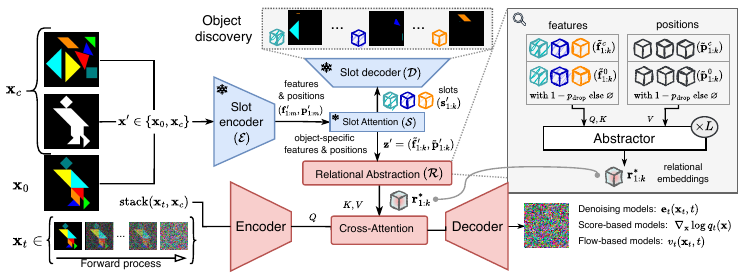}
    \caption{\label{fig:architecture}Illustration of RDM with \textcolor{dscolor}{\emph{Discoverer}} and the \textcolor{rscolor}{\emph{Reasoner}} blocks. During training, the clues ($\mathbf{x}_c$) and solutions $(\mathbf{x}_0)$ are individually fed to the pretrained frozen (\snowflake{}) slot encoder to obtain the corresponding feature ($\mathbf{f}^\prime_{1:m}$) and positional ($\mathbf{p}^\prime_{1:m}$) embeddings. Slot Attention yields the slots ($\mathbf{s}_{1:k}^\prime$) for reconstruction by the slot decoder, and object-specific features and positions, $\smash{\mathbf{z}^\prime=(\tilde{\mathbf{f}}^\prime_{1:k}, \tilde{\mathbf{p}}^\prime_{1:k})}$ for relational abstractions. The relational abstraction module consists of $L$ abstractor layers that optimize the relational embeddings (\relcube), which are introduced into the U-Net bottleneck through a cross-attention layer. Since the solution is unavailable at inference time, its disentangled representations  ($\tilde{\mathbf{f}}^0_{1:k}, \tilde{\mathbf{p}}^0_{1:k}$) are suppressed by a default learnable embedding ($\varnothing$) with probability $p_\text{drop}$.}\vspace{-.2cm}
\end{figure}

\subsection{The \emph{Discoverer}: Unsupervised Slot-based Object Discovery}\label{subsec:discoverer}

The objective of the \emph{discoverer} is to extract informative latent representations $\mathbf{z}^\prime$ via the encoding $p_\phi(\mathbf{z}^\prime|\mathbf{x}^\prime)$ of a latent variable model, while simultaneously disentangling the constituent objects within $\mathbf{x}_0$ and $\mathbf{x}_c$.
We use \textit{slot attention}~\citep{locatello-etal-2020-object} to achieve this, which enables unsupervised learning of disentangled object embeddings from visual data via a slot-based autoencoder.

\textbf{Slot attention:}
The slot encoder maps input visual features into a shared latent space, where $k$ trainable vectors (the \emph{slots}) compete to explain parts of the input via an iterative attention process.
These $k$ slots are then passed to a spatial broadcast decoder to reconstruct the input $\mathbf{x}^\prime$.
By enforcing this reconstruction bottleneck, the model learns to partition the scene into $k$ distinct, permutation-invariant latent variables that capture the essential objects for the downstream reasoning task.

Let $\mathbf{x}\in\mathbb{R}^{\texttt{H}\times \texttt{W}\times \texttt{c}}$ denote an input image, and $(\mathcal{E}, \mathcal{S}, \mathcal{D})$ denote the slot encoder, slot attention and slot decoder modules of the discoverer.
The slot encoder $\mathcal{E}:\mathbb{R}^{\texttt{H}\times \texttt{W}\times \texttt{c}}\to (\mathbb{R}^{m\times d},\mathbb{R}^{m\times d})$ is a feature extractor that yields $m$ features ($\mathbf{f}_{1:m}$) and positional embeddings ($\mathbf{p}_{1:m}$) in a $d$-dimensional latent space.
The slot attention module $\mathcal{S}:(\mathbb{R}^{m\times d},\mathbb{R}^{m\times d})\to\mathbb{R}^{k\times d}$ learns to iteratively update $k$ randomly initialized slots $\mathbf{s}_{1:k}$, using $\eta$ recurrent cross-attention iterations over $\mathbf{f}_{1:m}$ and $\mathbf{p}_{1:m}$ to bind object-centric information to the slots.
The final attention matrix is then used to compute feature- and position-specific embeddings ($\smash{\tilde{\mathbf{f}}_{1:k}}$ and $\smash{\tilde{\mathbf{p}}_{1:k}}$) for each slot, which will be considered as $\mathbf{z}=(\tilde{\mathbf{f}}_{1:k};\tilde{\mathbf{p}}_{1:k})$ (see Algorithm~\ref{alg:slot-attention}). 
The slot decoder $\mathcal{D}:\mathbb{R}^{k\times d} \to \mathbb{R}^{\texttt{H}\times \texttt{W}\times \texttt{c}}$ reconstructs the input by processing each slot independently through transposed convolutions to produce slot-specific reconstructions $\hat{\mathbf{x}}_i\in\mathbb{R}^{\texttt{H}\times \texttt{W}\times \texttt{c}}$ and associated spatial masks $\boldsymbol{\alpha}_i$ for $i=1,\ldots,k$.
The final image $\hat{\mathbf{x}}$ is computed as the weighted sum of $\{\hat{\mathbf{x}}_i\}_{i=1}^k$ with $\text{softmax}_{i=1...k}(\boldsymbol{\alpha}_i)$ (see \S\ref{ap:discoverer_config} for further details).

\textbf{Scaling the number of slots:} 
One of the main limitations of the original slot decoder is the computational overhead of the latent bottleneck when $k$ increases.
Previous works have typically restricted $k\leq 10$ due to the memory overhead of performing independent high-resolution convolutions and subsequent weighted averaging for every slot~\citep{singh-etal-2022-illiterate,singh-etal-2022-simple,mondal-etal-2023-learning}. 

\begin{wrapfigure}{r}{0.40\textwidth}\centering
    \vspace{-14pt}
    \hspace{2em}
    \begin{algorithm}[H]
      \caption{\label{alg:scalable-decoder}Scalable slot decoder}
      \SetKwInOut{Input}{Input}
      \Input{$\mathbf{s}_{1:k}\in \mathbb{R}^{k \times d}$, $\mathbf{g}\in\mathbb{R}^{\texttt{h}\times\texttt{w}\times d}$}
      $\tilde{\boldsymbol\alpha}\leftarrow\text{softmax}(\mathbf{g}\cdot\mathbf{s}^\top_{1:k})$ \sidecomment{[h,w,$k$]}
      $\tilde{\mathbf{s}} \leftarrow \sum_{i=1}^k \tilde{\boldsymbol{\alpha}}_i \cdot \mathbf{s}_i$\sidecomment{[h,w,$d$]}
      \Return $ \mathcal{D}(\tilde{\mathbf{s}})$
    \end{algorithm}
    \vspace{-10pt}
\end{wrapfigure}
To scale our discoverer to a large number of slots, we introduce the \textit{scalable slot decoder} in Algorithm \ref{alg:scalable-decoder} that significantly reduces the memory demand during the backward pass.
It relies on an additional learnable positional grid $\mathbf{g}$ that activates specific slots with specific locations in the bottleneck of the model.
Rather than decoding each slot into a full-resolution image before weighting, our scalable decoder first estimates which slots are relevant to specific locations by computing the dot product between the positional grid $\mathbf{g}$ and the slot embeddings $\mathbf{s}_{1:k}$, yielding $\tilde{\boldsymbol{\alpha}}$ after softmax normalization.
Subsequently, the slot information is weighted by $\tilde{\boldsymbol{\alpha}}$ and projected onto the grid, producing a single feature map $\tilde{\mathbf{s}}$, where slot information is isolated on specific positions of the grid.
This aggregated representation is then passed through the remaining decoder layers to generate a unified reconstruction of the input image.
Effectively, introducing this learnable positional grid creates a computationally more efficient bottleneck during optimization.

\subsection{The \emph{Reasoner}: Relational Object Abstractions as Diffusion Guidance}
\label{subsec:reasoner}

The \emph{reasoner} is a diffusion model conditioned on the clue $\mathbf{x}_c$ and the embeddings $\mathbf{z}^\prime=(\tilde{\mathbf{f}}_{1:k}^\prime;\tilde{\mathbf{p}}_{1:k}^\prime)$ obtained by the discoverer, where $\prime\in\{0,c\}$.
We formalize generative modeling of $q(\mathbf{x}_0|\mathbf{x}_c)$ through a diffusion guidance framework~\citep{ho-etal-2022-classifier} based on embeddings learned by a relational abstraction module ($\mathcal{R}$).
We use a single parameterization of $p_\theta$ to learn both the \emph{solution-guided} distribution $p_\theta(\mathbf{x}_0|\mathbf{x}_c, \mathcal{R}(\mathbf{z}^c,\mathbf{z}^0))$, and \emph{unguided} distribution $p_\theta(\mathbf{x}_0|\mathbf{x}_c,\mathcal{R}(\mathbf{z}^c))$.
This resembles the training setup of classifier-free guidance (CFG).
Unlike CFG, however, RDM does not combine the guided and unguided distributions for sampling, since the solution is unavailable at inference time.

\textbf{Relational abstraction:} 
This module $\smash{\mathcal{R}:(\mathbb{R}^{K\times d}, \mathbb{R}^{K\times d})\to \mathbb{R}^{K\times d}}$ processes the latents $\smash{(\tilde{\mathbf{f}}_{1:k}^*,\tilde{\mathbf{p}}^*_{1:k})}$ with \textit{Slot Abstractors}~\citep{mondal-etal-2024-slot}, where $*=(c,0)$ and $K\geq 2k$ varies with the number of clues and solutions.  
It consists of interleaved cross-attention, self-attention and feed-forward blocks with residual connections, designed to hierarchically capture inter-slot semantics and positional relations by sequentially refining the positional signal ($\smash{\tilde{\mathbf{p}}^*_{1:k}}$) with query-key feature projections ($\smash{\tilde{\mathbf{f}}^*_{1:k}}$) into \textit{relational embeddings} $\mathbf{r}^*_{1:k}$ (see \S\ref{ap:abstractor_config} for details).
After $L$ abstractor blocks, this module returns $\smash{\mathbf{r}^*_{1:k}=\mathcal{R}(\tilde{\mathbf{f}}_{1:k}^*,\tilde{\mathbf{p}}^*_{1:k})}$, encoding object relations.
This ensures that the generative process will be guided not only by individual object features, but also by their contextual interactions.

\textbf{Conditional generative modeling:} 
During training, the disentangled representations of the solution ($\mathbf{z}^0$) are nullified with probability $p_\text{drop}$ to simulate the unguided distribution $p_\theta(\mathbf{x}_0|\mathbf{x}_c,\mathcal{R}(\mathbf{z}^c))$ used at inference; and maintained with probability $(1-p_\text{drop})$ to learn the solution-guided distribution $\smash{p_\theta(\mathbf{x}_0|\mathbf{x}_c,\mathcal{R}(\mathbf{z}^c,\mathbf{z}^0))}$.
By modeling both distributions with a single parameterization $p_\theta$, the reasoner learns: (1) the relations between the disentangled factors of the clue and the solution ($\mathbf{r}^*_{1:k}$) and (2) to generate the solution $\mathbf{x}_0$ at inference-time by relying only on the clue information ($\mathbf{x}_c$, $\mathbf{z}^c$).
Training can be either based on noise prediction~\citep{ho-etal-2020-denoising},  score matching~\citep{song-etal-2021-score}, or flow matching~\citep{lipman-etal-2023-flow} (which we use in our experiments, see \S\ref{ap:reasoner_config} for details).

\section{Spatial Reasoning Benchmark}

We present a novel benchmark comprising four datasets with up to one million images each, inspired by classic logic puzzles, where underlying logical rules are implicit in the conditional data distribution and must be inferred from training data.
We categorize them into two paradigms: (i) \textit{grid-based image completion} tasks: \Akari and \Coldoku, where reasoning is framed as inpainting under global and local spatial constraints, and (ii) \textit{multi-context image generation}: \Tangram and \LogicFace, where a latent rule must be inferred from context images to synthesize a consistent solution.

\subsection{Grid-based Image Completion Tasks}

\begin{wrapfigure}{r}{0.64\textwidth}\centering
    \vspace{-15pt}
    \begin{subfigure}[t]{0.24\linewidth}\small
        \caption{\label{fig:akari-clue}Clue}
        \includegraphics[width=\linewidth]{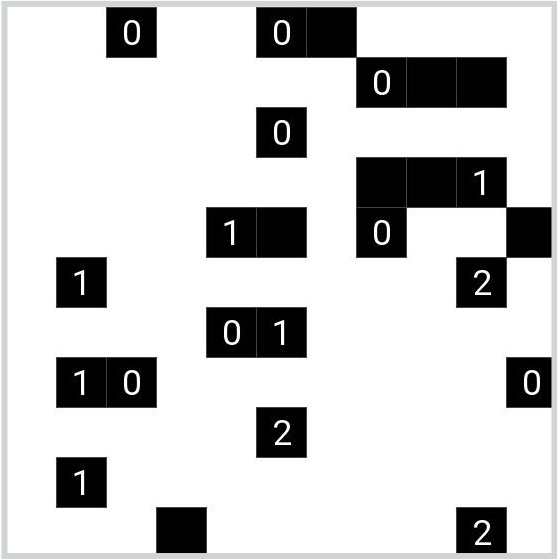}
    \end{subfigure}
    \begin{subfigure}[t]{0.24\linewidth}\small
        \caption{\label{fig:akari-solution}Valid \ding{51}}
        \includegraphics[width=\linewidth]{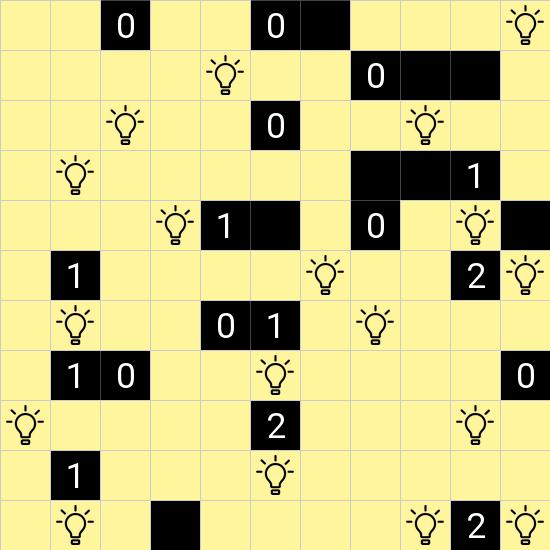}
    \end{subfigure}
    \begin{subfigure}[t]{0.49\linewidth}\small
        \caption{\label{fig:akari-invalid}Invalid \ding{55}}
        \includegraphics[width=0.49\linewidth]{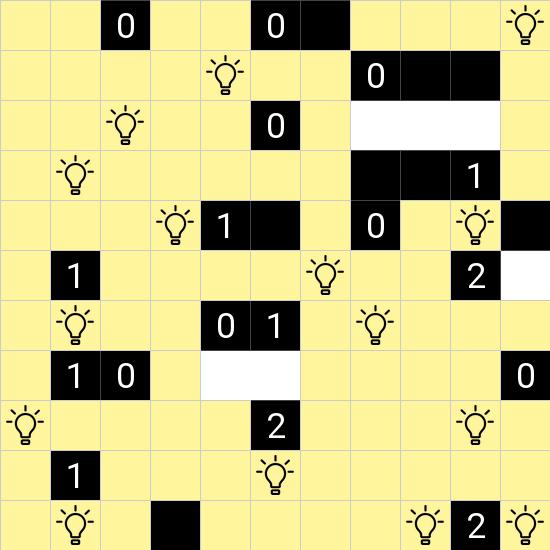}
        \includegraphics[width=0.49\linewidth]{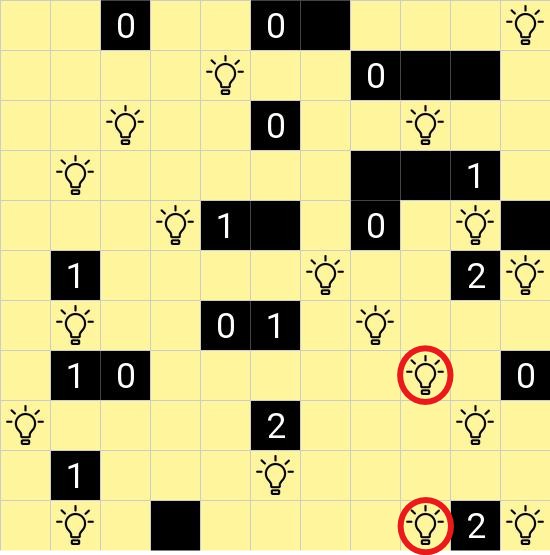}
    \end{subfigure}
    \vspace{-3pt}
    \caption{\label{fig:example-akari}\Akari as a grid-based image completion puzzle.}
    \vspace{-10pt}
\end{wrapfigure}

\textbf{\Akari dataset:}
\Akari is a binary-determination puzzle played on a rectangular grid of \textit{empty cells} and \textit{walls}. 
The goal is to place light bulbs in empty cells such that every cell is illuminated while ensuring no two bulbs share a line of sight.
Walls obstruct light and may be numbered with a digit ($0$-$4$), specifying the exact number of bulbs in its four orthogonally adjacent cells.
Figure \ref{fig:example-akari} illustrates
these constraints, where the \textit{clue} (a) defines wall and digit placements. 
A valid solution (b) illuminates all cells without bulb interference, satisfying the wall clues.
Invalid configurations (c) fail if cells remain unlit or bulbs conflict.
We generated a dataset of one million $11\times 11$ \Akari puzzles with randomized wall configurations.
Unlike traditional \Akari, which requires a unique solution, we allow for multiple valid bulb arrangements to evaluate the model's ability to generate novel, correct solutions.
Additional details on how this dataset is generated can be found in \S\ref{ap:akari}.

\textbf{\Coldoku (\textit{Colored Sudoku}) dataset:} 
Sudoku is a logic-based puzzle played on a $9\times 9$ grid, divided into nine $3\times 3$ subgrids.
The objective is to fill the grid with digits from 1 to 9 such that each column, row, and subgrid does not contain repeated digits. 
The puzzle setter provides a partially completed grid as the \textit{clue}, from which a well-posed solution can be derived.
Recent works~\citep{wever-etal-2025-spatial} used MNIST-based
\begin{wrapfigure}{r}{0.72\textwidth}\centering
    \vspace{-14pt}
    \begin{subfigure}[t]{0.24\linewidth}\small 
        \caption{\label{fig:sudoku}MNIST-Sudoku}
        \includegraphics[width=\linewidth]{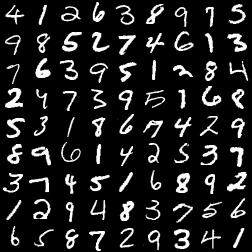}
    \end{subfigure}
    \begin{subfigure}[t]{0.24\linewidth}\small 
        \caption{\label{fig:coldoku-colors}Colors}
        \includegraphics[width=\linewidth]{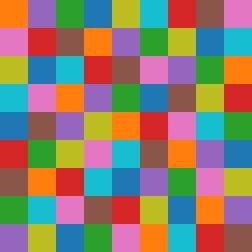}
    \end{subfigure}
    % \\
    \begin{subfigure}[t]{0.24\linewidth}\small 
        \caption{\label{fig:coldoku-clue}\Coldoku}
        \includegraphics[width=\linewidth]{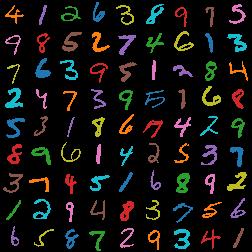}
    \end{subfigure}
    \begin{subfigure}[t]{0.24\linewidth}\small 
        \caption{\label{fig:coldoku-solution}Clues selection}
        \includegraphics[width=\linewidth]{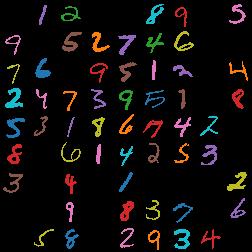}
    \end{subfigure}
    \vspace{-3pt}
    \caption{\label{fig:example-coldoku}\Coldoku puzzle, which combines Sudoku with a color arrangement that also satisfies Sudoku rules.
    \textit{Tableau 10} palette is used, which is designed to have a large perceptual difference between colors.
    }
    \vspace{-10pt}
\end{wrapfigure}
variants of this task for generative modeling.
We extend this in \Coldoku, a colored variant of this puzzle, where we again have a $9\times 9$ grid satisfying the Sudoku rules (Figure \ref{fig:sudoku}), and add an extra layer of complexity by coloring the digits such that there are no repeated colors in each row, column and subgrid (Figures \ref{fig:coldoku-colors}).
\Coldoku requires solving two different Sudokus simultaneously, one considering the digits and the other considering colors.
More details can be found in \S\ref{ap:coldoku}.

\subsection{Multi-context Image Generation Tasks}

\textbf{\Tangram dataset:}
\Tangram is a dissection puzzle consisting of seven flat shapes (\textit{tans}): a square ($\square$), a parallelogram ($\parallelogramm$), two small triangles ($\triangle^S\times 2$), a medium triangle ($\triangle^M$) and two large triangles ($\triangle^L\times 2$), which can be arranged to form figures without overlapping. 
\begin{wrapfigure}{r}{0.61\columnwidth}\centering
    \vspace{-10pt}
    \begin{subfigure}[t]{0.24\linewidth}
        \caption{\label{fig:tangram-pieces}Tans}\small
        \includegraphics[width=\textwidth]{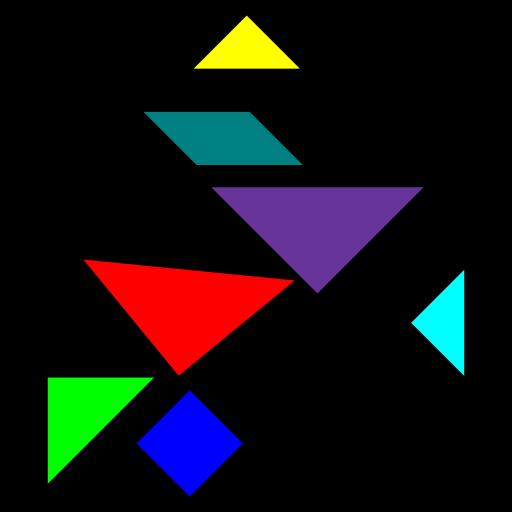}
    \end{subfigure}
    \begin{subfigure}[t]{0.24\linewidth}\small
        \caption{\label{fig:tangram-layout}Silhouette}
        \includegraphics[width=\textwidth]{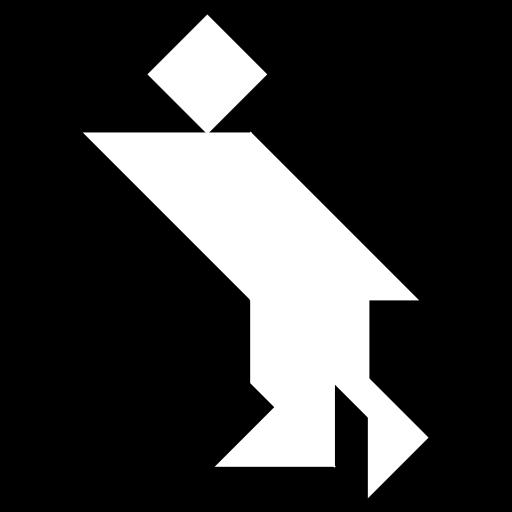}
    \end{subfigure}
    \begin{subfigure}[t]{0.24\linewidth}\small
        \caption{\label{fig:tangram-solution}Valid \ding{51}}
        \includegraphics[width=\textwidth]{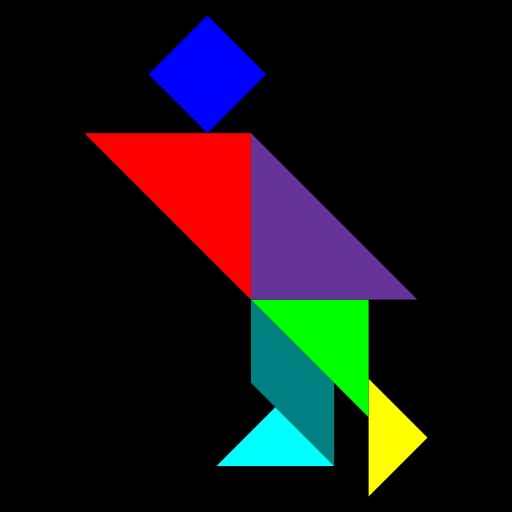}
    \end{subfigure}
    \begin{subfigure}[t]{0.24\linewidth}\small 
        \caption{\label{fig:tangram-invalid}Invalid \ding{55}}
        \includegraphics[width=\textwidth]{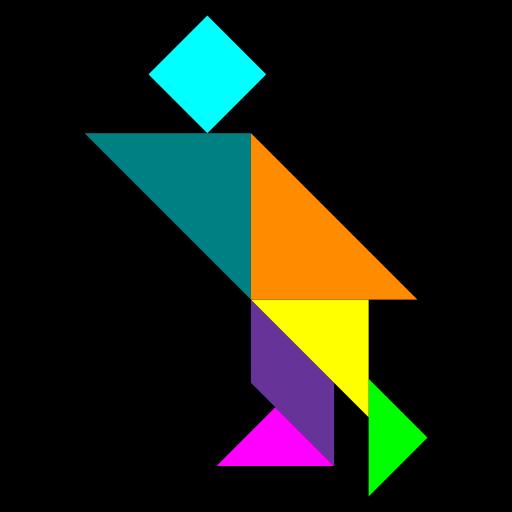}
    \end{subfigure}
    \vspace{-3pt}
    \caption{\label{fig:example-tangram}\Tangram example. A valid solution uses all pieces matching the colors and shapes, given the (a) and (b) as \textit{clues}.}
    \vspace{-10pt}
\end{wrapfigure}
The goal is to assemble all the pieces to match a given \textit{silhouette}, such that no piece overlaps with other, and all pieces of the puzzle are used (Figure \ref{fig:example-tangram}).
\Tangram requires reasoning over shapes, sizes, rotations and displacements of the pieces to project the final solution, reflecting forms of geometric reasoning studied in humans \citep{bohning-etal-1997-using,simpson-etal-2019-link}.
We rely on the Kilogram dataset~\citep{ji-etal-2022-abstract} consisting of 1,013 uncolored \Tangram solutions, and introduce the vision task of generating a valid solution (Figure \ref{fig:tangram-solution}) by using an input pattern (Figure \ref{fig:tangram-layout}) and a set of colored tans (Figure \ref{fig:tangram-pieces}). 
We generate different tan views and solutions by rotating the puzzle and randomly assigning distinct colors to tans (see \S\ref{ap:tangram} for details, as well as a discussion on \Tangram task complexity in \S\ref{ap:reasoner_results}).

\textbf{\LogicFace (\textit{Logical Operators on Facial Features}) dataset:}
We present a challenging logic-based generative modeling problem that requires abstraction and reasoning over real images.
Specifically, \LogicFace defines a conditional generative modeling task for human faces, where certain facial features are combined through hidden logical operators that must be inferred from training examples.
\begin{wrapfigure}{r}{0.5\textwidth}\centering
    \vspace{-14pt}
    \begin{subfigure}[t]{0.32\linewidth}\centering\small
        \caption{\label{fig:logicface-A}$A$}
        \includegraphics[width=\linewidth]{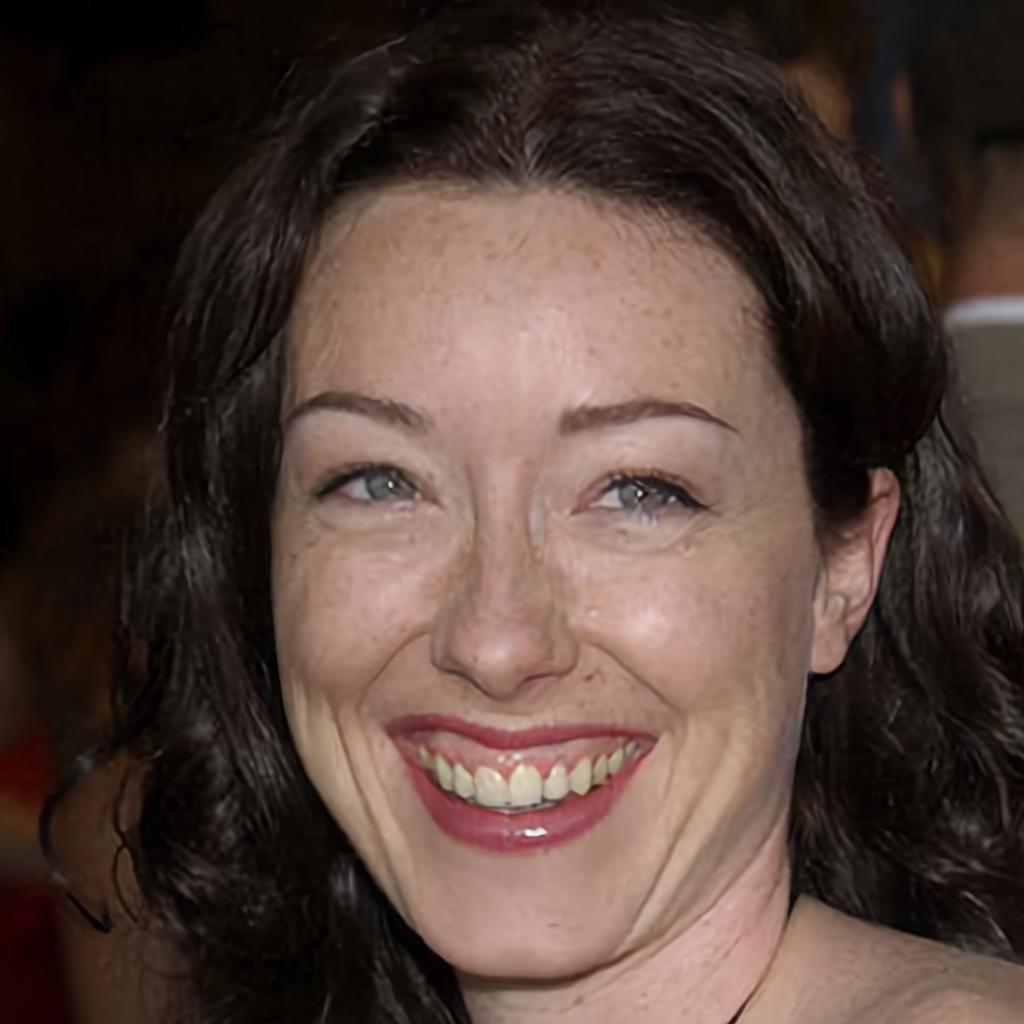}
        $\boldsymbol{a} = [1, 0, 0, 0]$ 
    \end{subfigure}
    \begin{subfigure}[t]{0.32\linewidth}\centering\small 
        \caption{\label{fig:logicface-B}$B$}
        \includegraphics[width=\linewidth]{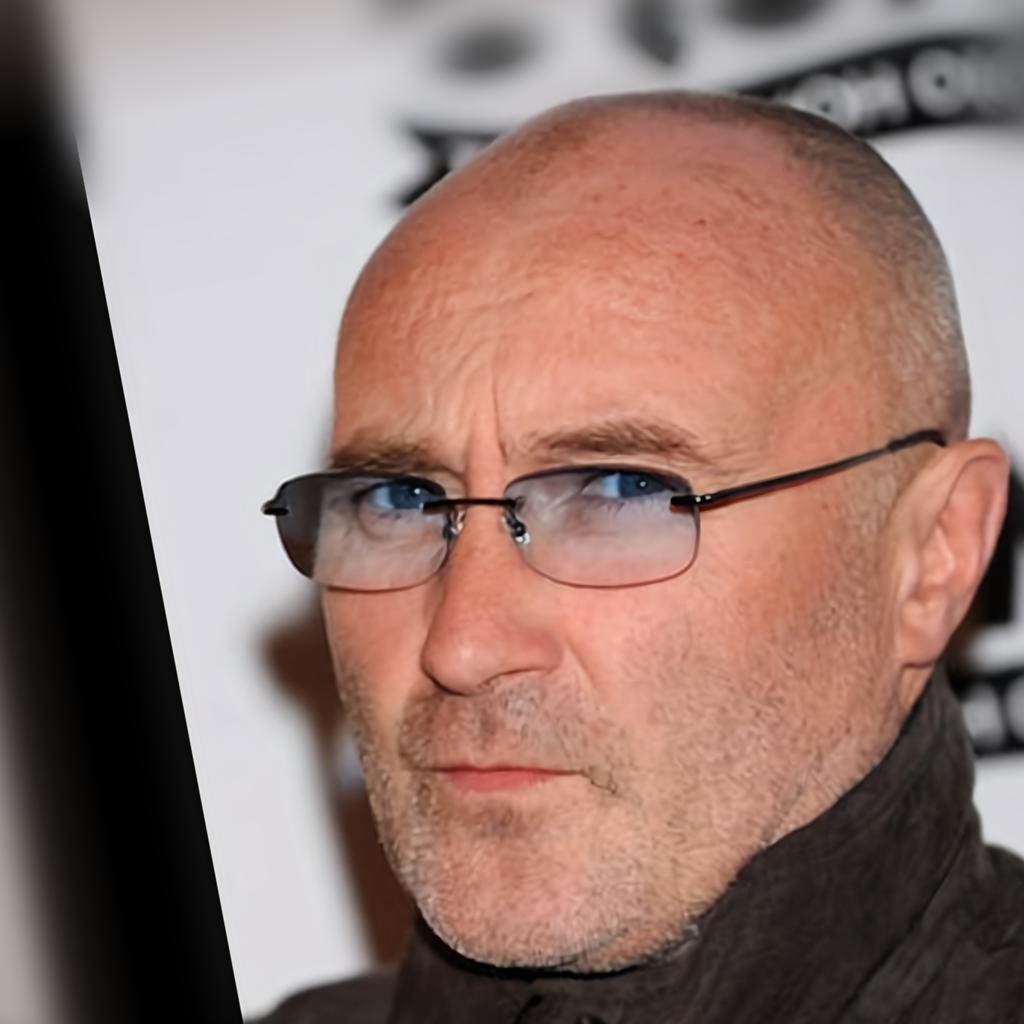}
        $\boldsymbol{b} = [0, 1, 1, 0]$
    \end{subfigure}
    \hfill 
    \begin{subfigure}[t]{0.32\linewidth}\centering\small  
        \caption{\label{fig:logicface-S}$S$}
        \includegraphics[width=\linewidth]{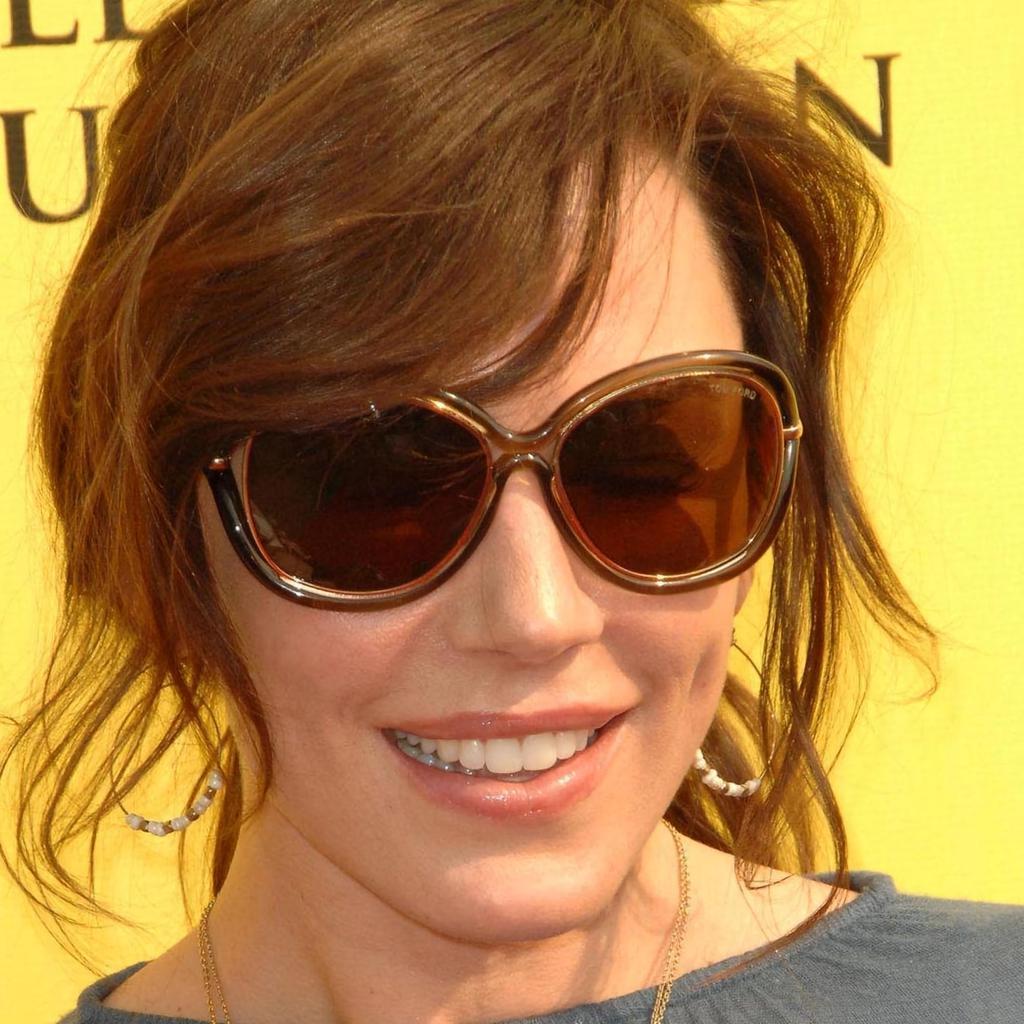}
        $\boldsymbol{s} =[1, 0, 1, 1]$
    \end{subfigure}
    \vspace{-3pt}
    \caption{\label{fig:example-logicface}\LogicFace example with 4 features: \textit{smiling} (\texttt{OR}), \textit{male} (\texttt{AND}), \textit{glasses} (\texttt{XOR}), \textit{young} (\texttt{IMPLIES}). Features from $A$ and $B$ are combined in $S$ via the corresponding logic operators (e.g., \textit{glasses}: $A=0$, $B=1$, $S=1$, satisfying \texttt{XOR}).}
    \vspace{-10pt}
\end{wrapfigure}
Figure \ref{fig:example-logicface} shows an example, where each data point is composed of three images from CelebA-HQ \citep{lee-etal-2020-maskgan} associated with a subset of $\kappa$ binary features retrieved from its original annotations. 
We consider the first two images (\textit{clues)} with the arguments $A$ and $B$, and the third as the \textit{solution} $S$.
Each image has a binary feature vector $\boldsymbol{a}$, $\boldsymbol{b}$, $\boldsymbol{s}\in\{0,1\}^\kappa$, and the dataset contains triples for which, for a given element-wise logical operator $f_j$, the relation $f_j(\boldsymbol{a}_j, \boldsymbol{b}_j) = \boldsymbol{s}_j$ holds.
We set $\kappa=4$ for the attributes to represent the feature vector of each image and a pool of 4 logical operators, each one assigned to one feature (see \S\ref{ap:logicface} for details).

\section{Experiments}
\label{sec:experiments}

\textbf{Model configurations:} 
For the discoverer, we use a convolutional autoencoder where the number of slots ($k$), slot dimension ($d$) and bottleneck resolution $(\texttt{h}, \texttt{w})$ are adapted to the specific task (see Table~\ref{tab:dataset-stats}).  The slot encoder maps the input into $m$ feature and positional embeddings. Slot attention performs $\eta=3$ recurrent iterations and feeds $k$ slots to the decoder.
For \Tangram and \LogicFace we chose $k=7$ and $k=6$, respectively.
For \Coldoku and \Akari, where the true number of objects scales with the number of cells, i.e., 81 and 121 respectively, we propose two configurations: one where $k$ coincides with the number of cells, hence requiring reconstruction with our scalable slot decoder (Algorithm \ref{alg:scalable-decoder}), and the other with $k=16$, such that each slot integrates information from multiple cells.
For experiments with $k\in\{6,7,16\}$ we use the original slot decoder~\citep{locatello-etal-2020-object}. 
For the reasoner, we use a U-Net architecture with an attention-based bottleneck, where relational embeddings learned in $L=6$ abstractor layers are incorporated via cross-attention~\citep{chen-etal-2023-training}.
To study the impact of model size, we conduct experiments with small (\texttt{S}), medium (\texttt{M}) and large (\texttt{L}) backbone U-Net architectures.
Further architectural details are explained in \S\ref{ap:RDM}.

\textbf{Conditioning on relational abstractions:}
Our framework can guide diffusion models via abstraction and conditioning on information from two types of inputs during training: the clue (via $\mathbf{z}^c$) and the solution (via $\smash{\mathbf{z}^0}$).
During training, we drop the condition $\smash{\mathbf{z}^0}$ with probability $\smash{p_\text{drop}=0.5}$\footnote{We observed that smaller $\smash{p_\text{drop}}$ values degrade estimation of the unconditional distribution of interest at inference time, although recommended in previous works~\citep{ho-etal-2022-classifier}. We found the best balance at $\smash{p_\text{drop}=0.5}$, hypothesizing that it allows to evenly learn both the \textit{solution-guided} and \textit{unguided} distributions.} to disable the conditioning derived from the solution, since $\mathbf{x}_0$ will not be available at inference time.

We explore three different conditioning settings (alongside $\mathbf{x}_c$), where relational embeddings are injected via cross-attention in the denoiser bottleneck: (i) \textit{clue}: conditioning on $\mathcal{R}(\mathbf{z}^c)$ during training and inference, (ii) \textit{solution}: $\mathcal{R}(\mathbf{z}^0)$ guides training and is dropped with $p_\text{drop}$, and inference is unconditioned on any relational embedding, and (iii) \textit{full}: both $\mathbf{z}^c$ and $\smash{\mathbf{z}^0}$ (dropped with $p_\text{drop}$) are used during training, and only $\mathbf{z}^c$ is used at inference time.

\textbf{Training details:}
We pretrained the slot autoencoder independently for each task based on a simple reconstruction loss (\S\ref{ap:discoverer_config}), providing images of clues and solutions as training samples.
We assess the quality of the reconstruction and thus the slot embeddings on a validation set (see \S\ref{ap:results} for the final results and disentanglement visualizations).
For the reasoner, we use flow matching to create the diffusion trajectory as detailed in \S\ref{ap:reasoner_config}, where the mean and variance change linearly with $t$.
We perform deterministic sampling from the conditional distribution by solving the reverse-time ODE using $\tau$ discretization steps.
All generative models were trained with the same optimization settings.

\textbf{Baseline comparisons:}
We use a capacity-matched U-Net as baseline (Base) by augmenting the hidden dimensions to match the same model size as RDM, accommodating for the additional parameters of the \textit{Discoverer} and the Abstractor stack.
The Base model is trained to estimate $q(\mathbf{x}_0|\mathbf{x}_c)$ also via flow matching, with the main difference of not incorporating relational abstractions.
We also trained this architecture with diffusion forcing~\citep{chen-etal-2024-diffusion-forcing} for comparisons to SRM~\citep{wever-etal-2025-spatial} with \textit{parallel} and \textit{sequential} sampling strategies, where the latter additionally requires learning uncertainty at training time to perform patch-based autoregressive denoising.
We also performed experiments on MNIST-Sudoku and Counting Polygons benchmarks for completeness~\citep{wever-etal-2025-spatial}, and ablations with non-object-centric feature guidance (details and results in \S\ref{ap:additional_experiments}).

\textbf{Model evaluations:} 
We designed task-specific neural evaluators (see \S\ref{ap:datasets} for details), and used 5,000 held-out samples per task that served as test sets.
We use the overall accuracy (\textsc{acc}) as the main metric in all datasets.
For \Coldoku, we also report individual digit (\textsc{dgt}) and color (\textsc{clr}) accuracies across three difficulty levels: \textit{Easy} (50-60 clues), \textit{Medium} (40-50 clues) and \textit{Hard} (30-40 clues).
For \Akari, we also measure the bulb accuracy (\textsc{blb}), which only evaluates the correctness of bulb allocations by ignoring the numbered walls, and evaluate both in an \textit{in-distribution (ID)} regime where the wall density ratio matches the training set, and \textit{out-of-distribution (OoD)} regimes with extreme wall densities: \textit{low} (5-10\%) and \textit{high} (35-40\%).
For \Tangram, we measure \textsc{shape} accuracy, which only measures the correct partition of the silhouette, and the f\textsubscript{1}-score per tan type.
For \LogicFace, we measure accuracy at feature level (proportion of correctly generated attributes).
We report all results with $\tau=100$ unless specified otherwise (see \S\ref{ap:results} for $\tau=10$ and varying model sizes).

\begin{table}[t]\footnotesize
    \caption{\label{tab:grid-results}Experimental results on \Akari and \Coldoku datasets.}\vspace{-.2cm}
    \setlength{\tabcolsep}{2.8pt}
    \renewcommand{\arraystretch}{1}
    \resizebox{\linewidth}{!}{
    \begin{tabular}{llcccccc|ccccccccc}
        \toprule
        && \multicolumn{6}{c|}{\Akari} & \multicolumn{9}{c}{\Coldoku} \\
        &&\multicolumn{2}{c}{\textit{in-dist.}} & \multicolumn{2}{c}{\textit{low}} & \multicolumn{2}{c|}{\textit{high}} & \multicolumn{3}{c}{\textit{Easy (50-60)}}&  \multicolumn{3}{c}{\textit{Medium (40-50)}} & \multicolumn{3}{c}{\textit{Hard (30-40)}} \\ 
        && \textsc{blb} & \underline{\textsc{acc}} & \textsc{blb}& \underline{\textsc{acc}} & \textsc{blb} & \underline{\textsc{acc}} & \textsc{dgt} & \textsc{clr} & \underline{\textsc{acc}} & \textsc{dgt} & \textsc{clr} & \underline{\textsc{acc}} & \textsc{dgt} & \textsc{clr} & \underline{\textsc{acc}} \\
        \midrule
        \multicolumn{2}{l}{Base} & 86.02 & 80.04 & 43.62 & 39.48 & 92.22 & 88.92 & 93.53 & 94.73 & 89.00 & 61.81 & 78.56 & 49.06 & 11.38 & 25.31 & 3.88 \\
        \midrule 
        \multicolumn{5}{l}{SRM~\citep{wever-etal-2025-spatial}} \\ 
        \multicolumn{2}{l}{\textit{parallel}} & 35.82 & 18.80 & 1.60 & 0.94 & 54.18 & 33.34 & 94.67 & 96.74 & 91.06 & 61.02 & 78.53 & 51.01 & 12.17 & 25.91 & 4.00  \\
        \multicolumn{2}{l}{\textit{sequential}} & 00.00 & 00.00 & -- &  -- & -- & -- & 93.05 & 96.70 & 92.37 & \textbf{82.21} & 83.61 & 68.35 & \textbf{40.95} & 36.66 & \textbf{20.78} \\
        \midrule 
        \multicolumn{3}{l}{RDM$^*$} \\
        \parbox[t]{3pt}{\multirow{3}{*}{\rotatebox[origin=c]{90}{\scriptsize \textit{16 slots}}}}
        & \textit{clue} & 88.30 & 82.06 & \bf 88.30 & 72.48 & 92.90 & 88.94   & 98.06 & 98.50 & 96.66 & 78.18 & 82.20 & 65.62 & 21.40 & 25.64 & \phantom{0}7.76\\
        & \textit{solution} & 86.10 & 78.68 & 82.21 & 70.73 & 92.51 & 87.96   & 96.80 & 96.18 & 93.16 & 71.80 & 67.36 & 50.68 & 16.64 & 12.74 & \phantom{0}3.34\\
        & \textit{full} & \bf 89.44 & 80.42 & 83.32 & \bf 75.06 & \bf 93.80 & 87.48     & \textbf{98.14} & \textbf{99.30} & \textbf{97.48} & 80.12 & \textbf{88.22} & \textbf{71.64} & 22.76 & \textbf{36.88} & 12.58\\
        \parbox[t]{3pt}{\multirow{3}{*}{\rotatebox[origin=c]{90}{\scriptsize \textit{>80 slots}}}}
        & \textit{clue} & 88.88 & \bf 83.12 & 83.10 & 54.58 & 93.62 & \bf 89.56     & 96.82 & 94.88 & 91.98 & 79.18 & 63.40 & 47.12 & 16.46 & 11.76 & \phantom{0}2.86 \\
        & \textit{solution} & 86.36 & 80.28 & 81.12 & 74.58 & 92.37 & 88.94           & 96.82 & 95.26 & 92.38 & 72.04 & 67.24 & 50.86 & 17.24 & 12.94 & \phantom{0}3.36 \\
        & \textit{full}  & 85.84 & 77.34 & 77.30 & 56.20 & 92.76 & 87.40               & 91.12 & 99.18 & 90.36  & 51.86 & 86.26 & 46.14 & 6.46 & 32.06 & \phantom{0}2.94\\
        \bottomrule           
    \end{tabular}}
    $^*$Results of RDM\textsuperscript{\texttt{[S]}} for \Akari, and RDM\textsuperscript{\texttt{[L]}} with $\tau=10$ for \Coldoku.
    \vspace{-5pt}
\end{table}

\section{Experimental Results}\label{sec:results}

We demonstrate the performance of our models on the proposed benchmark in Tables \ref{tab:grid-results} and \ref{tab:tangram-logicface-results}, where RDM outperforms the baselines in \Akari, \Tangram and \LogicFace, while only lagging behind SRM in the hardest level of \Coldoku.
We present extended results and output examples in \S\ref{ap:results}. 

\textbf{Results on \Akari:}\label{subsec:akari-analysis}
RDM consistently outperforms baseline models across all distribution regimes and demonstrates superior generalization capabilities under the most challenging (\textit{low}) regime.
Under the in-distribution (ID) evaluation, RDM achieves 83.12 accuracy with 121 slots, compared to 80.04 for standard diffusion.
We observe a performance margin between 16 and 121 slots (\textit{clue}: 83.12 vs 82.06), potentially because the slot count allows the discoverer to freely learn a bijective mapping between slots and cells, enhancing the reasoner to better exploit bulb placement.
In the \textit{high} regime, all models benefit from an increased wall ratio, which bounds the solution space, localizes constraints, and reduces ambiguity. 
Conversely, the \textit{low} regime imposes harder reasoning demands as the solution space expands beyond the training distribution.
Here we find evidence of how relational abstractions capture the underlying task rules rather than just imitating pixel-wise distributions, and show that disentanglement with $k=16$ strongly aids out-of-distribution generalization in contrast to $k=121$, which slightly overfits the training set.
While Base model plummets (in-distribution (Table~\ref{tab:grid-results}): 80.04 $\to$ \textit{low}: 39.48), RDM \textit{clue} with 16 slots maintains consistency (in-distribution: 82.06 $\to$ \textit{low}: 72.48).

Our evaluation of SRM reveals that diffusion forcing does not universally benefit grid-based puzzles.
While SRM excels in MNIST-Sudoku~\citep{wever-etal-2025-spatial}, where patch-based noise levels combined with uncertainty-based sampling align with optimal Sudoku-solving heuristics, its strategy fails to generalize to \Akari, where reasoning variables are not evenly distributed across cells.
During training, diffusion forcing yields a vast combinatorial space of noise levels that mostly affect illuminated cells, damaging convergence (\textsc{acc}: 18.80).
At inference, SRM \textit{sequential} prematurely prioritizes denoising empty (yellow-illuminated) cells due to their persistent low uncertainty, resulting in performance collapse, where the critical task of bulb placement is deferred until the final sampling states.

\textbf{Results on \Coldoku:}
Our results from Table~\ref{tab:grid-results} show that RDM generally outperforms all baselines in color accuracy and overall accuracy at the easy and medium levels of \Coldoku, and SRM performs better in the hard regime. The best performance was achieved with 16 slots (\textsc{acc}: 97.48 \textit{easy}, 71.64 \textit{medium} and 12.58 \textit{hard}), whereas the \textit{full} 81-slot configuration caused \textsc{acc} on the easy level to drop to 90.36.
We hypothesize that the primary bottleneck in \Coldoku is not the slot embeddings themselves, but rather the relational abstraction module ($\mathcal{R}$) shared between the clue and the solution as input with a high number of slots $k=81$, which requires the unique $\mathcal{R}$ to reconcile complex positional and feature relations from two distinct distributions.
\Coldoku involves a diverse set of objects (multi-colored digits) in varying positions, while \Akari presents a much smaller object vocabulary, where the complexity is exclusively positional.

\begin{table}[t]\footnotesize\centering
    \caption{\label{tab:tangram-logicface-results}Experimental results on \Tangram and \LogicFace datasets.}\vspace{-.2cm}
    \setlength{\tabcolsep}{4pt}
    \renewcommand{\arraystretch}{1}
    \resizebox{\linewidth}{!}{
    \begin{tabular}{lccccccc|cccccc}
        \toprule
        & \multicolumn{7}{c|}{\Tangram} & \multicolumn{6}{c}{\LogicFace}\\
        & \parallelogramm  &  $\square$ & $\triangle^S$ &  $\triangle^M$ & $\triangle^L$ &  \textsc{shape} & \underline{\textsc{acc}}  & \makecell[c]{\textit{smiling} \\ (\texttt{OR})} & \makecell[c]{\textit{male} \\(\texttt{AND})} & \makecell[c]{\textit{glasses} \\(\texttt{XOR})} & \makecell[c]{\textit{young} \\ (\texttt{IMP.})} & \textsc{avg} & \underline{\textsc{acc}} \\ 
         \midrule
         Base & 23.06 & 24.22 & 42.82 & 41.50 & 55.60 & 57.72 & 15.46 & 54.60 & 76.52 & 67.34 & 38.90 & 57.84 & 11.24\\
         \midrule
         \multicolumn{5}{l}{SRM~\citep{wever-etal-2025-spatial}\footnotemark} \\
         \textit{parallel} & -- & -- & --& -- & -- & --& -- & 37.00 & 38.16 & 49.26 & 39.44 & 40.97 & 2.60 \\
         \textit{sequential} & -- & -- & --& -- & -- & --& -- & 25.04 & 58.38 & 50.58 & 73.86 & 51.94 & 5.70 \\
         \midrule 
         \multicolumn{3}{l}{RDM\textsuperscript{\texttt{[S]}}} \\ 
         \textit{clue} & 36.71 & 36.96 & 59.59 & 60.33 & 63.29 & 58.41 & 34.37 & \bf 90.19 & \bf 97.00 & \bf 97.26 & \bf 87.27 & \bf 92.93 & \bf  74.74 \\
         \textit{solution} & 41.12 & 43.94  & 56.49 & 53.06 & 59.94 & 60.25 & 38.57 & 64.87 & 73.20 & 50.16 & 67.78 & 64.00 & 16.27 \\
         \textit{full} & \bf 49.75 & \bf 49.98 & \bf 61.55 & \bf 61.25 & \bf 65.00 & \bf 65.12 & \bf  46.76 & 83.16 & 94.00 & 94.76 & 79.81 & 87.93 & 60.76 \\
         \bottomrule 
    \end{tabular}}
    \flushleft
    \vspace{-5pt}
\end{table}
\footnotetext{Diffusion forcing led to highly unstable training in \Tangram, where both SRM \textit{parallel} and \textit{sequential} were unable to outperform Base (similar to \LogicFace). These findings also align with~\citet{wever-etal-2025-spatial}.}

\textbf{\Tildoku (\textit{Tilted and Colored Sudoku)}:}
We observed that SRM performs well on hard grid-based puzzles with sequential, cell-wise reasoning.
However, its reliance on the grid's geometry and patch structure introduces an implicit task-aligned bias.
To challenge this in a more general scenario, we introduce \Tildoku, where \Coldoku images are randomly tilted (see \S\ref{ap:coldoku} for details), disrupting the alignment between grid cells and image coordinates.
Thus, the spatial structure now does not necessarily align with the grid-informed reasoning mechanism that SRM utilizes.

\begin{wrapfigure}{r}{0.55\linewidth}\centering\footnotesize
    \vspace{-8pt}
    \caption{\label{fig:tildoku-results}Performance of RDM\textsuperscript{[\texttt{L}]} in \Tildoku.}\vspace{-.2cm}
    \hspace{-6pt}
    \begin{subfigure}[c]{0.37\linewidth}
    \includegraphics[width=\linewidth]{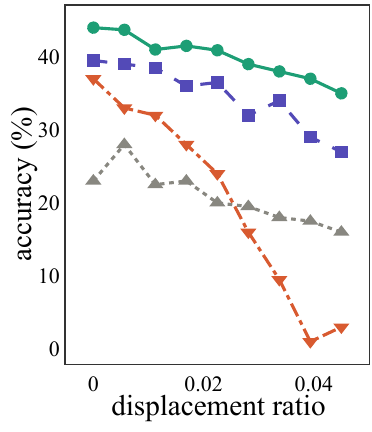}
    \end{subfigure}
    \hfill
    \begin{subfigure}[c]{0.6\linewidth}
    \vspace{-10pt}
    \setlength{\tabcolsep}{3pt}
    \begin{tabular}{lccc}
        \toprule
         \textit{Medium} & \textsc{dgt}  &  \textsc{clr} & \underline{\textsc{acc}} \\
         \midrule
         \colorbullet{baselinecolor} Base  & 22.62 & 45.92 & 12.48 \\
         \colorbullet{srmcolor} SRM (\textit{seq.})  & 24.31 & 21.31 & 7.08 \\
         \midrule 
         \colorbullet{cluecolor} \textit{clue} & 33.02 & 52.84 & 19.18 \\ 
         \colorbullet{solutioncolor} \textit{solution} & \bf 43.70 & \bf 60.12 & \bf 28.40\\
         \textit{full} & 22.24 & 31.96 & 8.46 \\ 
         \bottomrule
    \end{tabular}
    \end{subfigure}
    \vspace{-10pt}
\end{wrapfigure}
Figure \ref{fig:tildoku-results} shows results on \Tildoku (\textit{Medium}, $\tau=10$) for different displacement ratios at test time (left), and their average (right), where RDM demonstrates more robust reasoning (\textsc{acc}: 28.40 averaged across various displacement ratios with the \textit{solution} approach).
Notably, SRM significantly worsens when increasing the tilting degree at inference time (orange line in the plot), resulting in lower average metrics.
This is due to the fact that the learned uncertainty is only consistent when patches cover objects from a stable distribution (e.g., when each patch covers exactly one digit), and it becomes unstable when a patch covers half of a digit or none. Figure \ref{fig:tildoku-results} (left) demonstrates this effect, where SRM outperforms the baseline at low tilting degrees, but abruptly collapses when digits do not align within fixed patches (see Table~\ref{tab:ap-tildoku-results} for extended results).

\textbf{Results on \Tangram:}
Table~\ref{tab:tangram-logicface-results} shows that RDM significantly surpasses baselines across all metrics on \Tangram.
Notably, RDM (\textit{full}) reaches three times the baseline performance, achieving \textsc{acc}: 46.76, compared to baseline's 15.46 (see Figure~\ref{fig:tangram-generations} for visual results).
In contrast to \Akari and \Coldoku, we observe a considerable improvement when incorporating the solution into the abstractor stack (\textit{clue}: 34.37 $\to$ \textit{solution}: 38.57 $\to$ \textit{full}: 46.76).
This behavior arises from how relational features differ between the clue and solution distributions.
RDM (\textit{clue}) only extracts primitives from the piece view (colors and size proportions), mainly boosting accuracy in color assignment (Base: 15.46 vs \textit{clue}: 34.37), but maintaining the shape accuracy close to the baseline (Base: 57.72 vs \textit{clue}: 58.41).
Adding information from the solution view, which provides the explicit silhouette partition, boosts the shape and overall accuracy to 65.12 and 46.76 respectively, showing that the relational stack effectively captures the structural dependencies between these two views to guide reasoning.

\textbf{Results on \LogicFace:}
Our results in Table~\ref{tab:tangram-logicface-results} show superior performance of RDM.
In particular, clue-conditioned models achieve the highest accuracies (\textit{clue}: 74.74 and \textit{full}: 60.76, see Figure~\ref{fig:logicface-generations} for generated solutions), whereas standard diffusion models perform poorly (\textsc{acc}: 11.24).
Unlike \Akari, \Coldoku and \Tangram, where baselines remained competitive, \LogicFace presents the unique challenge of simultaneously modeling a complex real-image distribution, encoding high-level features of multiple input arguments (e.g., gender, presence of glasses), and inferring logical operations to generate a consistent target image.
Baseline models prioritize generating realistic human faces at the expense of satisfying the logical rules (see \S\ref{ap:results} for image quality metrics), effectively ignoring any relationship between the inputs and targets.
In contrast, RDM captures the underlying rules for all operators, as evidenced by accuracies that exceed trivial strategies, such as the model defaulting to a single mode (e.g., only generating a smiling female face).

\textbf{Impact of relational abstractions and evidence of rule learning:}
In an ablation study, we isolated the impact of \textit{relational abstractions} by evaluating a \textit{slot-based} diffusion guidance approach, i.e., $p_{\theta}(\mathbf{x}_0|\mathbf{x}_c,\mathbf{s}^c_{1:k})$ at training and inference, similar to previously proposed slot-conditioned diffusion models~\citep{jiang2023object}.
Our results in Table~\ref{tab:slot-diffusion-comparison} revealed that explicit relational abstractions are critical for complex reasoning, beyond purely slot-based diffusion guidance alone (see \S\ref{ap:additional_experiments}).

Finally, besides our rule-consistent image synthesis evaluations, we also investigated whether the relational abstraction module learns embeddings that capture the logical rules required for reasoning.
Across controlled experiments, we found evidence that these embeddings (1) yield near-optimal solution-conditioned inference at test time, (2) can help improve reasoning without the raw clue image conditioning, and (3) remain linearly separable between valid versus rule-violating puzzles (see \S\ref{ap:additional_experiments} for further details on our empirical evidence of rule learning experiments).

\section{Discussion}
\label{sec:discussion}

We study a key limitation of image diffusion models regarding their emphasis on pixel-level fidelity in conditional data distribution modeling, which limits their ability to capture logical structures required for spatial reasoning.
We introduce RDM as a novel framework that enhances diffusion models with unsupervised object-centric relational abstractions of the structural regularities in the training data.
While RDM does not incorporate explicit symbolic computation modules, it leverages learned relational abstractions that move diffusion models toward more symbolic forms of reasoning.

We further present a large-scale spatial reasoning benchmark comprising diverse challenging tasks, on which our approach significantly improves reasoning-driven image generation capabilities.
We show that harnessing abstractions from different task components (clues and solutions) enhances reasoning both when applied during training alone for diffusion guidance (\textit{solution}), and when used jointly during training and inference (\textit{clue} and \textit{full}).
Our work does not aim to position any single variant as the preferred recipe, but instead demonstrates how object-centric relational abstractions can guide generative spatial reasoning with a task-agnostic approach.
Experiments on grid-based image completion tasks revealed that current task-informed uncertainty-driven inference-time reasoning approaches~\citep{wever-etal-2025-spatial} do not reliably scale across diverse reasoning tasks, where the failure modes became more pronounced in \Tildoku under distorted grid structures.
Nevertheless, our method is in principle compatible with any additional sampling-time reasoning approach~\citep{wever-etal-2025-spatial}.

\textbf{Limitations:} 
Our approach relies on object discovery to extract relational patterns, utilizing neural architectures with inherently finite scalability.
While these models might not scale to ``in-the-wild'' scenarios in unconstrained scenes with thousands of entities, we argue that it still remains aligned with the scope of human reasoning, where the number of objects is naturally bounded by cognitive constraints.
Nevertheless, more powerful relational vision representations may enable better scalability in natural scenes, making open-world visual reasoning an important direction for future research.
We also note that the neural-based automated evaluators used in our benchmarks may be prone to minor estimation errors due to their own performance. 
However, the comparisons presented remain fair and valid as the identical evaluation pipeline applied consistently for all models.

\subsection*{Reproducibility statement}

We provide detailed descriptions of the model architectures, training configurations and hyperparameters of the experiments reported in the paper, both partly in the main text, and in \S\ref{ap:datasets} and \S\ref{ap:RDM}.
Our code and datasets are available at: \href{https://github.com/anaezquerro/rdm}{https://github.com/anaezquerro/rdm}.

\subsubsection*{Acknowledgments}
This work was supported by the ELLIS Unit Graz and the Graz Center for Machine Learning (GraML).
This research was funded in whole by the Austrian Science Fund (FWF) [10.55776/COE12].
The computational results presented have been achieved in part using the Austrian Scientific Computing (ASC) infrastructure.

\bibliography{bibliography}
\bibliographystyle{iclr2027_conference}

\newpage
\appendix

\section{Details on the Spatial Reasoning Benchmark Datasets}\label{ap:datasets}

\begin{wraptable}{r}{0.55\columnwidth} 
    \centering\scriptsize
    \vspace{-10pt} 
    \setlength{\tabcolsep}{4pt}
    \begin{tabular}{@{}c|cccc@{}}
        \toprule
        \multicolumn{1}{c}{}& \Akari & \Coldoku & \Tangram & \LogicFace  \\
        \midrule
        $\texttt{H} \times \texttt{W}$   & $198 \times 198$ & $180 \times 180$ & $128 \times 128$ & $128 \times 128$ \\
        $\texttt{h} \times \texttt{w}$ & $11 \times 11$     & $9 \times 9$     & $8 \times 8$     & $8 \times 8$     \\
        $k$            & 121 / 16           & 81 / 16             & 7             & 6               \\
        \# samples     & 1M          & 1M              & 1,013                & 30,000            \\
        neural evaluator & \xmark  & \cmark & \xmark & \cmark \\ 
        \bottomrule
    \end{tabular}
    \caption{\label{tab:dataset-stats}General statistics: image  ($\texttt{H} \times \texttt{W}$) and bottleneck ($\texttt{h} \times \texttt{w}$) resolution; number of slots ($k$) and unique samples (\# samples); and neural evaluation.}
    \vspace{-30pt}
\end{wraptable}
In this section, we provide a detailed description of our spatial reasoning datasets, and motivate their use for evaluating generative  models on challenging visual tasks.
Our primary goal in introducing this novel spatial reasoning benchmark is to move beyond simple pattern matching by requiring generative models to satisfy discrete geometric and logical reasoning constraints.

\subsection{\Akari dataset}\label{ap:akari}

\Akari (also known as \textit{Light Up}) is a binary-determination puzzle played on a rectangular grid of empty cells (\white) and walls (\wall).
The goal is to place light bulbs (\bulb) in empty cells such that every cell is illuminated (\light) while ensuring no pair of bulbs shine on one another, i.e., no two bulbs share a line of sight, horizontally or vertically, unobstructed by a wall.
Walls act as obstacles to light and may serve as \textit{numbered cells}.
These cells are annotated with a digit from 0 to 4, indicating the exact number of bulbs that must be placed in the four adjacent (orthogonal) cells.

We generated a dataset of one million $11\times 11$ \Akari puzzles\footnote{\Akari grids can be of arbitrary sizes, but most commonly $11\times 11$ or $13\times 13$. We opted for $11\times 11$ simply to balance the image resolution with our computational resources.}.
Our generation process follows a constraints-first approach.
We first distributed walls across 20\%--30\% of the grid, then allocated bulbs according to the core rules of the puzzle, and finally converted 50\% of the walls into clued cells by annotating them with the count of their adjacent bulbs, leaving the remaining walls unnumbered.

\begin{figure}[h]\centering
    \includegraphics[width=0.19\linewidth]{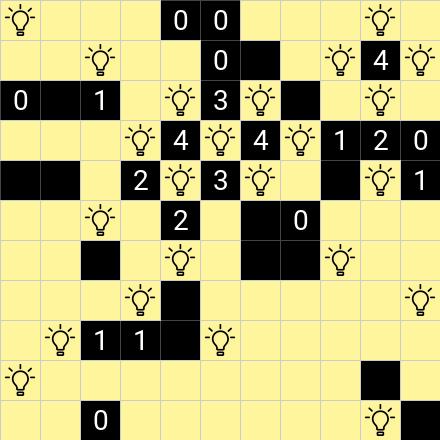}
    \includegraphics[width=0.19\linewidth]{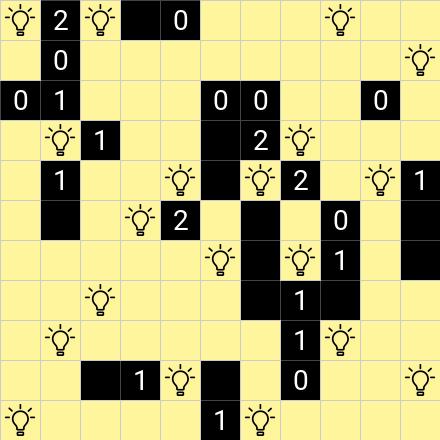}
    \includegraphics[width=0.19\linewidth]{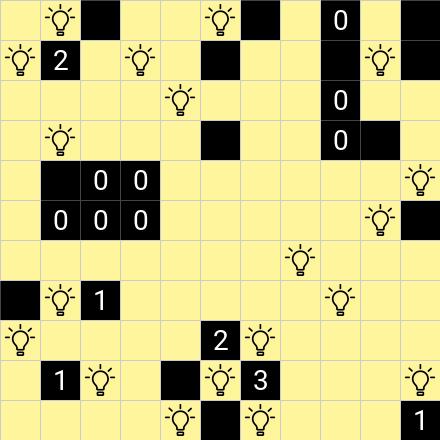}
    \includegraphics[width=0.19\linewidth]{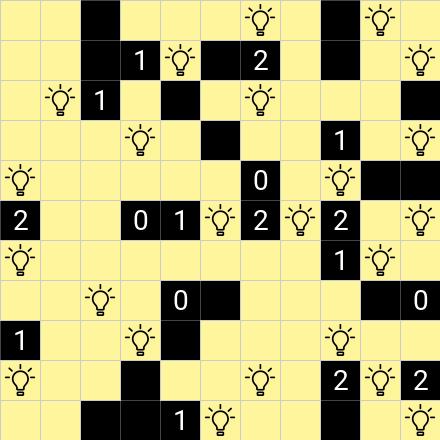}
    \includegraphics[width=0.19\linewidth]{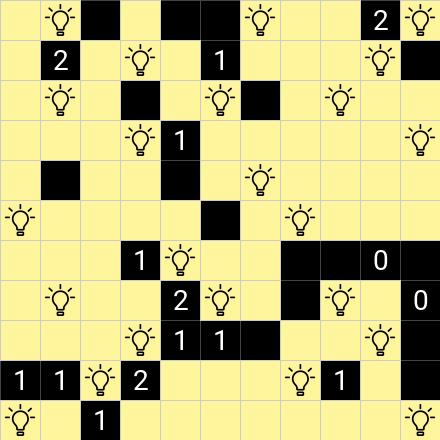}
    \\  
    \includegraphics[width=0.19\linewidth]{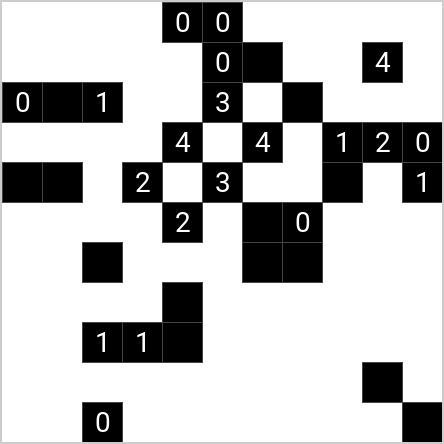}
    \includegraphics[width=0.19\linewidth]{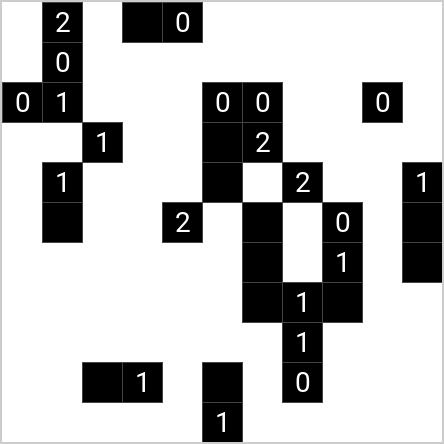}
    \includegraphics[width=0.19\linewidth]{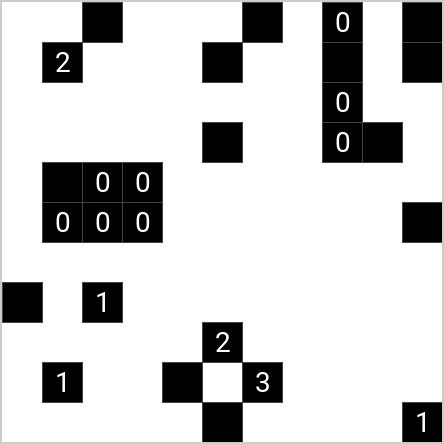}
    \includegraphics[width=0.19\linewidth]{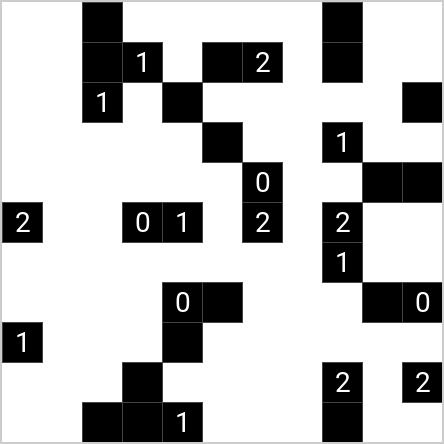}
    \includegraphics[width=0.19\linewidth]{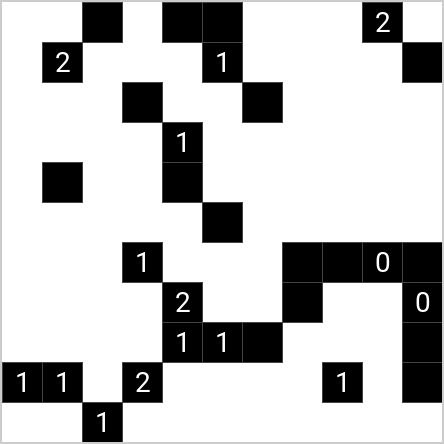}

    \caption{\Akari samples from the training set.}
\end{figure}

\textbf{\Akari properties:} 
\Akari is an NP-complete problem \citep{sun-etal-2021-shedding}, a complexity derived from the interplay between line-of-sight constraints and numerical wall requirements.
Without clued cells, the problem reduces to polynomial time.
Solving typically involves local constraint satisfaction, such as prioritizing bulbs for high-value clues or identifying bottleneck empty cells that can only be illuminated by a single potential bulb placement.

From a generative perspective, ensuring a unique solution elevates the task to a higher-order complexity ($\smash{\sum_2^P}$ or ASP classes \citep{elik-etal-2009-comparing}).
Standard algorithms rely on backtracking solvers: a candidate grid is produced and iteratively modified with additional wall constraints if more than one solution is found.
To better evaluate model novelty and rule abstraction, our dataset permits multiple valid arrangements, allowing novel solutions that may differ from training annotations.

\textbf{Evaluation of \Akari:} 
We evaluate generated images with a deterministic pipeline that individually classifies each cell as a bulb (\bulb) or illuminated background (\light) via mean squared error against target patches.
A sequential algorithm then verifies the arrangement against puzzle constraints.
We use five metrics: (i) \textit{accuracy}, requiring all constraints to be satisfied, (ii) \textit{bulb accuracy}, a relaxed metric ignoring numbered wall constraints, (iii) \textit{unlit count}, the count of unlit empty cells, (iv) \textit{distance}, as the absolute error between a wall's digit and its actual adjacent bulb count, and (v) \textit{overlap}, as the count of bulbs sharing a line of sight.

\subsection{\Coldoku: The Colored Sudoku dataset}\label{ap:coldoku}

Sudoku is a combinatorial number-placement puzzle played on a $9\times 9$ grid, which is further partitioned into nine $3\times 3$ subgrids (also known as \emph{blocks} or \emph{boxes}).
In its classical formulation, the objective is to populate the grid with digits from 1 to 9 such that each row, column, and subgrid contains every digit exactly once.
Various extensions of the puzzle have been proposed, ranging from larger grid dimensions to the addition of supplementary constraints.
The complexity of solving Sudoku puzzles has been formally classified as NP-complete \citep{takayuki-etal-2003-complexity}.
While it has been mathematically proven that a minimum of 17 initial clues is necessary for a unique solution, human-solvable puzzles typically provide 20 to 35 fixed digits to guide the player through the remaining cells.

We introduce on a \emph{multi-constraint} Sudoku variant that couples the classical digit-placement logic with a coloring challenge, named \Coldoku.
This task effectively requires the model to simultaneously solve a secondary Sudoku-style constraint involving nine distinct colors.
We relied on the MNIST-Sudoku dataset released by~\citet{wever-etal-2025-spatial} as our primary source for 1M solved Sudoku configurations.
Following their methodology, we train an MNIST classifier to populate the grid cells with handwritten digits, thereby generating a realistic handwritten view of the Sudoku grid.
To introduce the additional layer of complexity, we randomly pair each Sudoku with another independent Sudoku.
The second Sudoku determines the color assignment for each digit, effectively creating a \Coldoku puzzle where both digit and color placements must satisfy Sudoku constraints (Figure \ref{fig:example-coldoku}).
Our dataset uses the \textit{Tableau 10} palette, which is designed to have a large perceptual difference between colors (blue \colorbullet{tabblue}, orange \colorbullet{taborange}, green \colorbullet{tabgreen}, red \colorbullet{tabred}, purple \colorbullet{tabpurple}, brown \colorbullet{tabbrown}, pink \colorbullet{tabpink}, olive \colorbullet{tabolive} and cyan \colorbullet{tabcyan}.
By cross-combining these digit and color arrangements, the resulting dataset spans a massive state space of $10^{12}$ (1M$^2$) unique puzzle configurations.

\textbf{\Coldoku properties:}
\Coldoku inherits the formal logical properties of the classical Sudoku puzzle, as it effectively requires the joint resolution of two orthogonal Sudoku instances (one in the digit domain and one in the color domain) given the same set of initial clues.
While a human solver would likely adopt a sequential strategy, first solving the digit arrangement before addressing the color-based constraints, a neural network must perform these operations as a simultaneous multi-objective task.

The properties of uniqueness and complexity (NP-completeness) are preserved for both the digit and color layers.
Consequently, \Coldoku represents an inherently more challenging benchmark for generative models, as it requires the latent reasoning module to maintain two independent sets of global constraints and resolve them within a single inference pass.

\begin{figure}\centering
    \begin{subfigure}[t]{0.19\linewidth}
        \includegraphics[width=\linewidth]{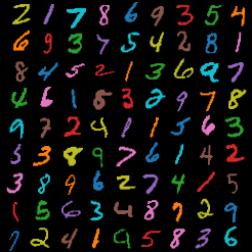}
        \caption{0.01}
    \end{subfigure}
    \begin{subfigure}[t]{0.19\linewidth}
        \includegraphics[width=\linewidth]{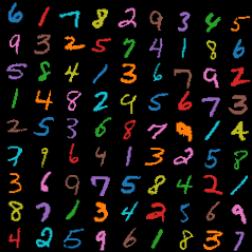}
        \caption{0.03}
    \end{subfigure}
    \begin{subfigure}[t]{0.19\linewidth}
        \includegraphics[width=\linewidth]{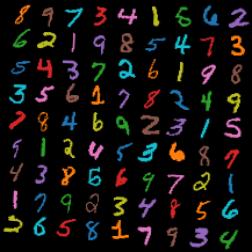}
        \caption{0.05}
    \end{subfigure}
    \begin{subfigure}[t]{0.19\linewidth}
        \includegraphics[width=\linewidth]{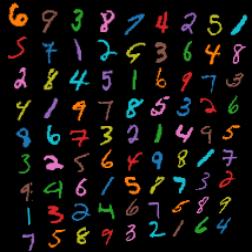}
        \caption{0.07}
    \end{subfigure}
    \begin{subfigure}[t]{0.19\linewidth}
        \includegraphics[width=\linewidth]{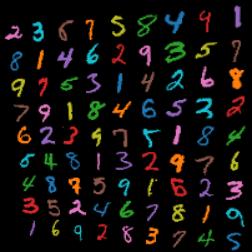}
        \caption{0.09}
    \end{subfigure}
    \caption{\label{fig:tildoku-examples}Visualization of \Tildoku \textit{solution} samples with different tilting ratios.}
\end{figure}

\textbf{\Tildoku (\textit{Tilted and Colored Sudoku}):}
To evaluate RDM in a more realistic scenario, we introduce \Tildoku, a variant of \Coldoku where samples are randomly tilted within a bounded angular range (see Figure \ref{fig:tildoku-examples}).
While \Tildoku maintains the underlying logical rules and reasoning complexity of \Coldoku, it introduces a visual challenge: digit completion now requires the model to detect the boundaries of a varying puzzle geometry.
To ensure the task remains approachable, we limited the tilting distortion to the 15\% of corner displacements. 
During training, each \Coldoku sample is subjected to a random tilt within these bounds before being processed by the model.
For evaluation, the associated affine matrix is utilized to reverse the transformation, allowing the predicted output to be verified by the \Coldoku evaluator.

\textbf{Evaluation of \Coldoku:} 
We assess the generative quality of \Coldoku samples using a two-stage pipeline.
First, we train a CNN classifier on $28 \times 28$ colored MNIST images ($>99\%$ test accuracy).
To ensure precise classification, these training images are quantized such that each digit contains only its target color centroid against a black background.
During evaluation, each generated $9 \times 9$ grid is partitioned into 81 individual cells and processed by the classifier to identify the digit-color pairs.
Consistent with \citet{wever-etal-2025-spatial}, we mitigate the impact of perception errors on our reasoning framework by constructing the input \Coldoku views exclusively from correctly classified MNIST samples. 
Following the perception stage, we independently verify the row, column, and subgrid constraints for both the digit and color arrangements.
We report three primary metrics: (i) \textit{digit accuracy}, ratio of grids where the 81 predicted digits satisfy all Sudoku constraints, (ii) \textit{color accuracy}, ratio of grids where the 81 predicted colors satisfy all Sudoku constraints, and (iii) \textit{accuracy}, joint ratio of grids where both the digit and color arrangements are simultaneously valid.

\begin{figure}\centering
    \includegraphics[width=0.19\linewidth]{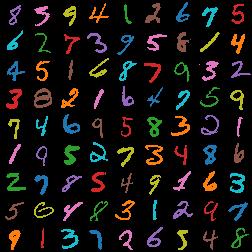}
    \includegraphics[width=0.19\linewidth]{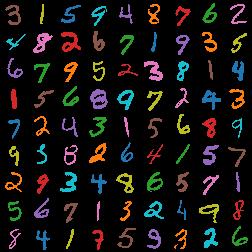}
    \includegraphics[width=0.19\linewidth]{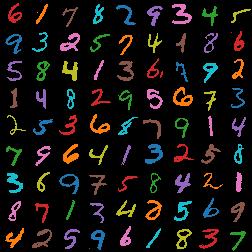}
    \includegraphics[width=0.19\linewidth]{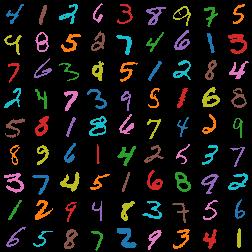}
    \includegraphics[width=0.19\linewidth]{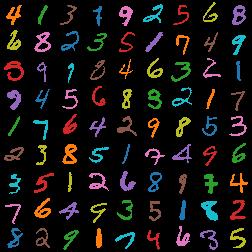}
    \\
    \includegraphics[width=0.19\linewidth]{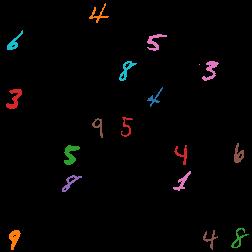}
    \includegraphics[width=0.19\linewidth]{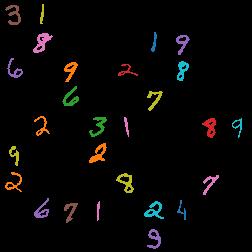}
    \includegraphics[width=0.19\linewidth]{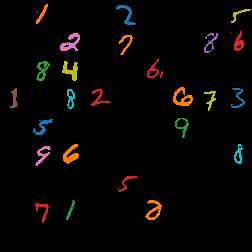}
    \includegraphics[width=0.19\linewidth]{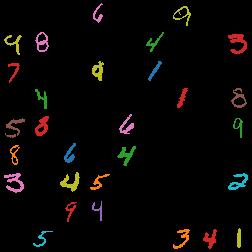}
    \includegraphics[width=0.19\linewidth]{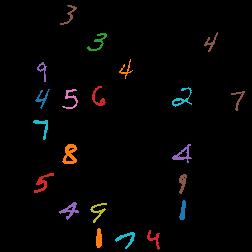}
    \caption{Visualization of \Coldoku samples from the training set (top: \textit{solution}, bottom: \textit{clue}).}
\end{figure}

\subsection{\Tangram dataset}\label{ap:tangram}

\begin{figure}[h]\centering
    \includegraphics[width=0.24\linewidth]{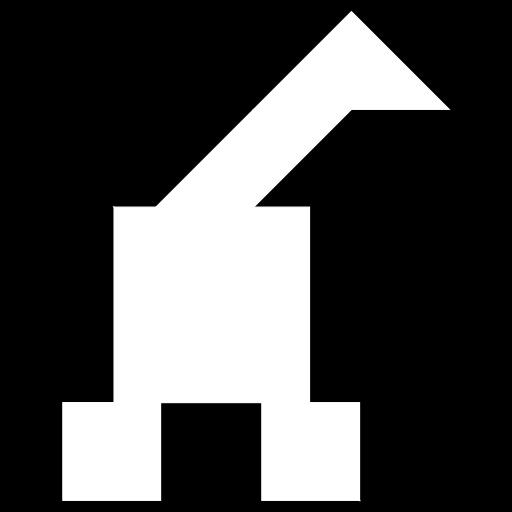}
    \includegraphics[width=0.24\linewidth]{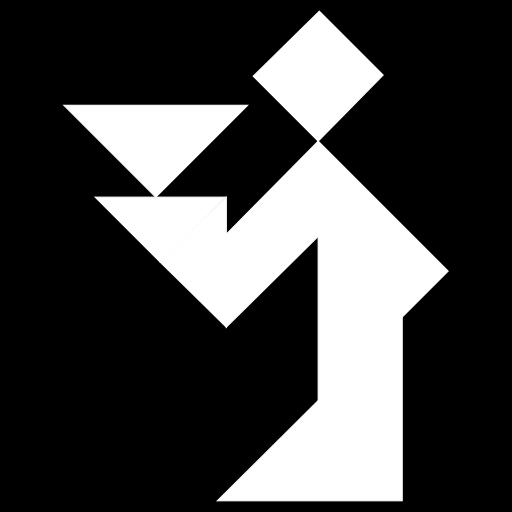}
    \includegraphics[width=0.24\linewidth]{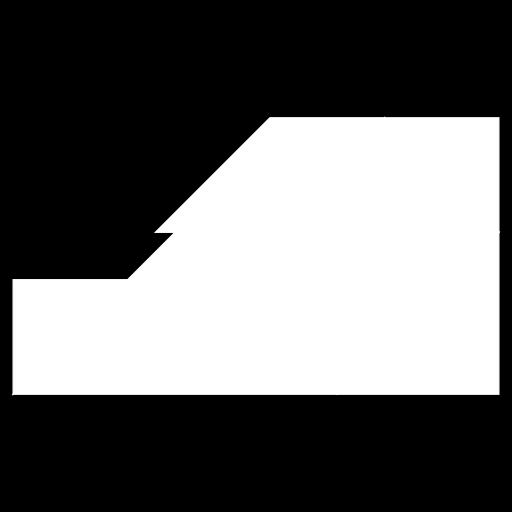}
    \includegraphics[width=0.24\linewidth]{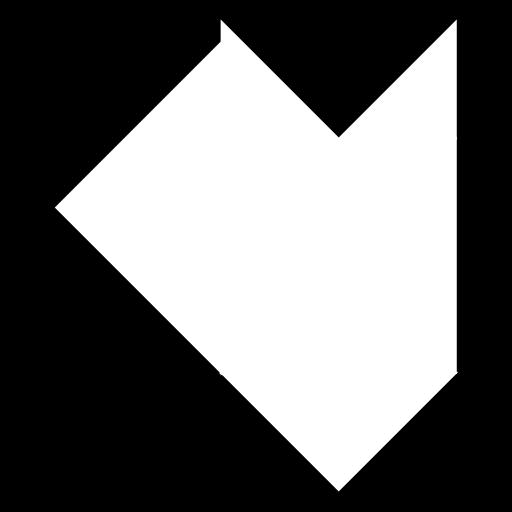}
    \caption{\label{fig:tangram-comparison}Tangram layouts ordered by reasoning difficulty. From easy configurations (left) containing salient geometric cues that simplify piece identification to complex silhouettes (right) characterized by high structural ambiguity, where multiple valid piece arrangements may exist for the same layout.}
\end{figure}

Tangram is a classic dissection puzzle consisting of seven flat shapes, called \emph{tans}, which must be arranged to form a specific target figure, or \emph{silhouette} (see Figure \ref{fig:tangram-layout}). 
The seven pieces (Figure \ref{fig:tangram-pieces}), comprising two large ($\triangle^L\times 2$), one medium ($\triangle^M$) and two small ($\triangle^S\times 2$) right triangles, one square ($\square$), and one parallelogram (\parallelogramm), remain constant across all puzzles. 
The core challenge of the Tangram task lies in the latent inference that is required to map these seven discrete pieces onto a silhouette, where individual piece boundaries are not visible. 
A valid solution must satisfy three primary constraints: (i) all seven tans must be used, (ii) no two pieces may overlap or occupy the same spatial coordinates, and (iii) the final arrangement must maintain the area and size proportions\footnote{All tans maintain strict area ratios relative to the smallest triangle: the medium triangle, square, and parallelogram has twice the area of a small triangle, while each large triangle has four times the area.} of the provided layout.
Many Tangram layouts exhibit high visual ambiguity, i.e., cases where internal boundaries are not easily identifiable, making them a rigorous test for visual reasoning and spatial reconstruction, even for human solvers (Figure \ref{fig:tangram-comparison}). 

We rely on the SVG annotations from the Kilogram dataset \citep{zhang-etal-2021-abstract}, which provides 1,013 uncolored, unique Tangram solutions.
Each sample consists of an SVG file containing precise polygon coordinates for the final silhouette.
We extended this dataset by randomly creating 9,200 shuffles to use as input clue (Figure \ref{fig:tangram-pieces}). 
To color the pieces and the solution, we randomly sample with replacement a pool of nine colors: red \colorbullet{red}, green \colorbullet{green}, blue \colorbullet{blue}, yellow \colorbullet{yellow}, magenta \colorbullet{magenta}, cyan \colorbullet{cyan}, purple \colorbullet{brown}, orange \colorbullet{darkorange} and teal \colorbullet{teal}, and match the color of each piece of the shuffled view to a piece of the same type in the solution. 

While the original Kilogram dataset was designed to evaluate the semantic associations between silhouettes and language, recent works have begun addressing the underlying spatial reasoning task \citep{yamada-etal-2025-tangan, zong-etal-2026-tangramsr}.
However, existing approaches such as TANGAN \citep{yamada-etal-2025-tangan} treat the problem as a standard generative task using Generative Adversarial Networks (GANs) to segment a silhouette into pieces.
Such methods may overlook the discrete constraints of the puzzle (e.g., that a specific set of seven pieces with fixed area ratios must be utilized), and ignore the actual reasoning on the piece proportions to simply recreate a shapes. 

To bridge this gap and enforce an actual human-like reasoning process, we augment the dataset by assigning distinct colors to each of the seven tans.
This modification requires the model to perform joint spatial and logical reasoning: it must not only partition the silhouette correctly but also match the specific geometry of each piece (e.g., the large triangle vs. the medium triangle) to its corresponding color and position. 
This ensures that the \emph{reasoner} is explicitly accounting for the size relationships and global constraints of the puzzle rather than relying on local shape generation. 
Additionally, this initialization (Figure \ref{fig:tangram-pieces}-\ref{fig:tangram-layout}) also resembles how humans solve the Tangram puzzle: from a random view of separated tans (the \emph{shuffle}), \emph{reasoning} over the color and proportions to fill the layout. 
Our perception is robust to different shuffles of the same seven pieces, as well as rotations of the silhouette.

\begin{figure}\centering
    \includegraphics[width=0.15\linewidth]{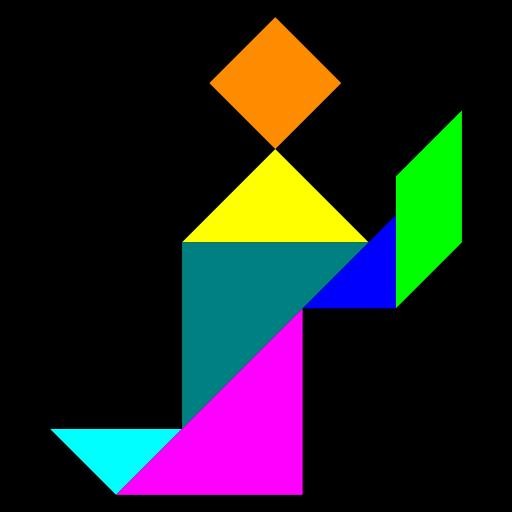}
    \includegraphics[width=0.15\linewidth]{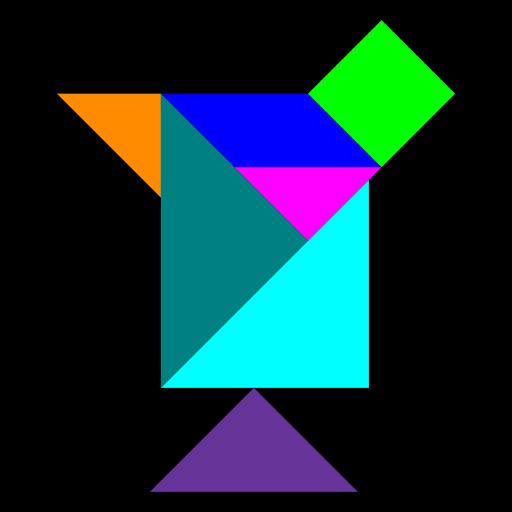}
    \includegraphics[width=0.15\linewidth]{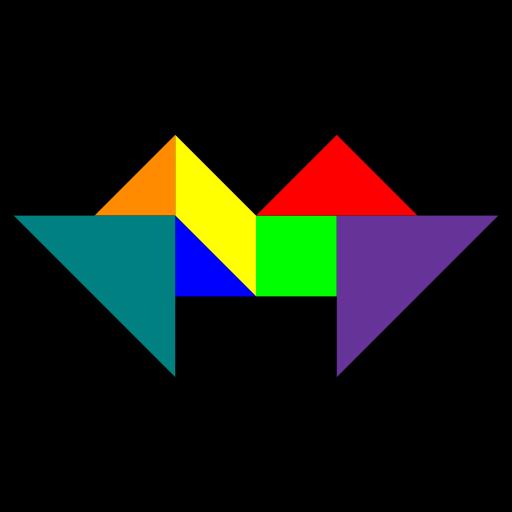}
    \includegraphics[width=0.15\linewidth]{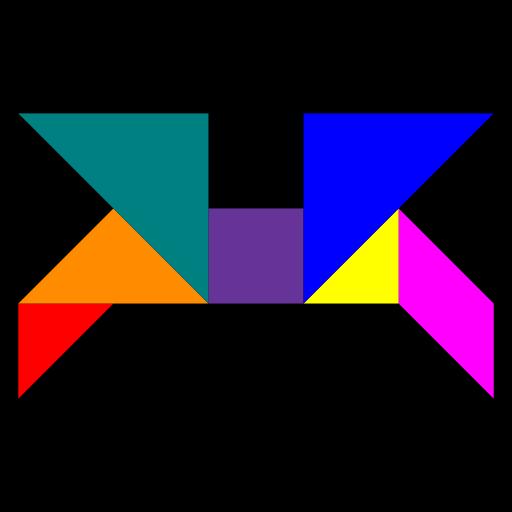}
    \includegraphics[width=0.15\linewidth]{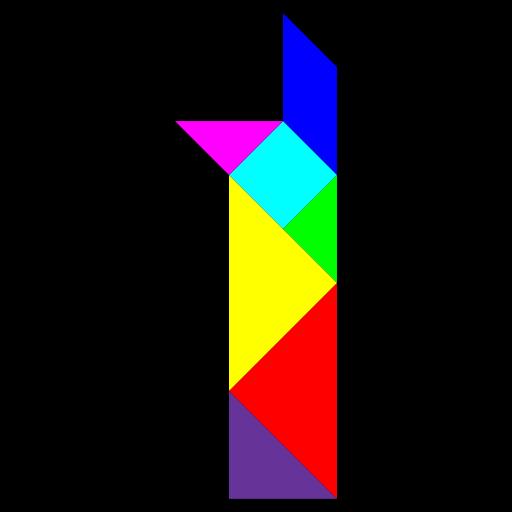}
    \includegraphics[width=0.15\linewidth]{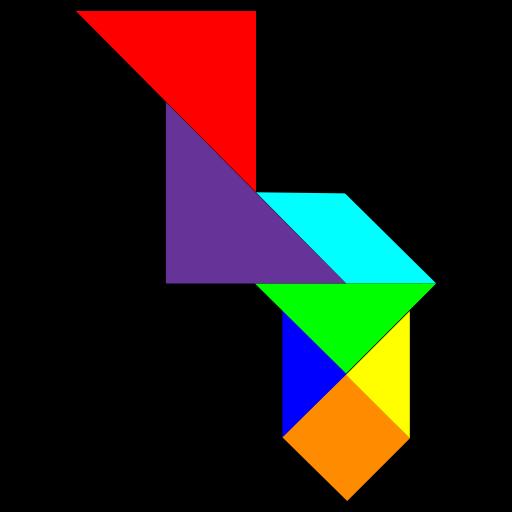}
    \\
    \includegraphics[width=0.15\linewidth]{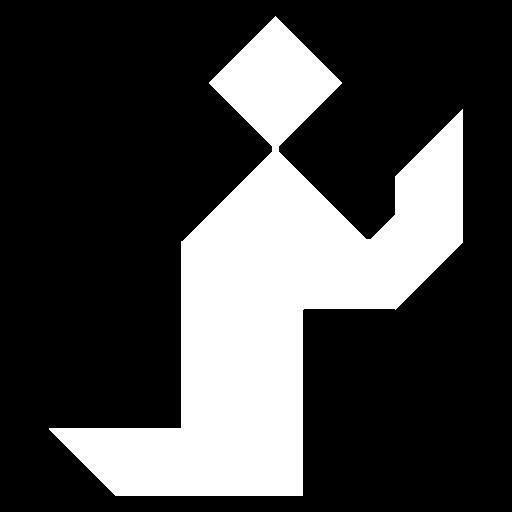}
    \includegraphics[width=0.15\linewidth]{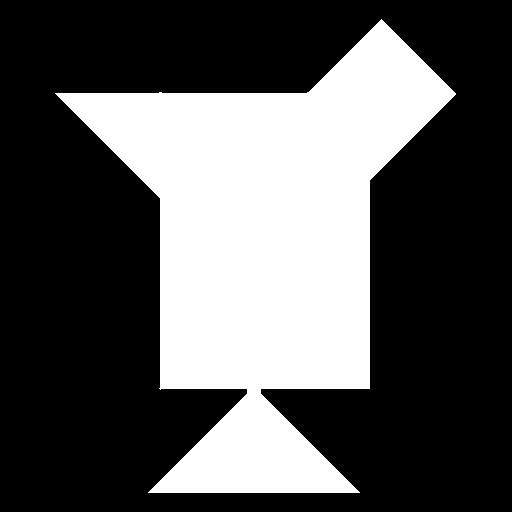}
    \includegraphics[width=0.15\linewidth]{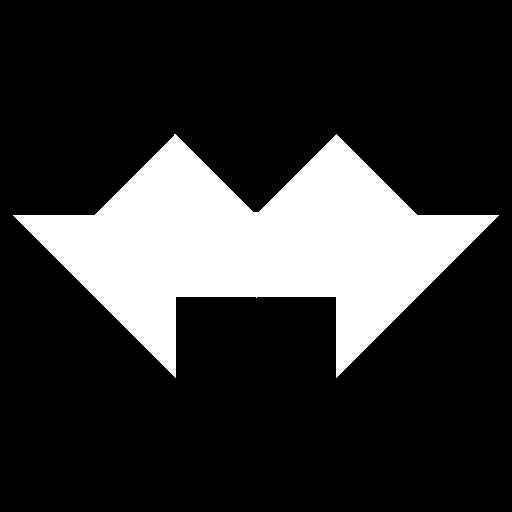}
    \includegraphics[width=0.15\linewidth]{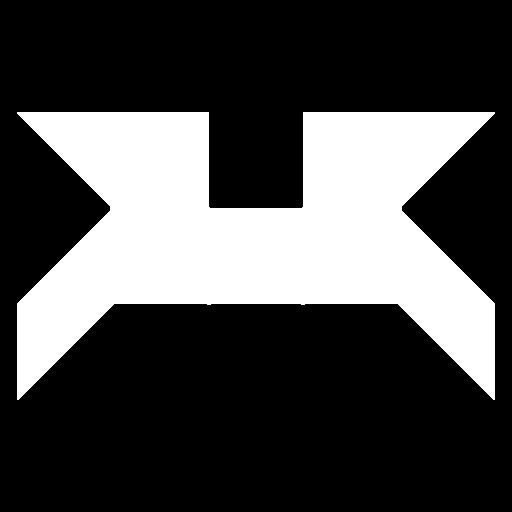}
    \includegraphics[width=0.15\linewidth]{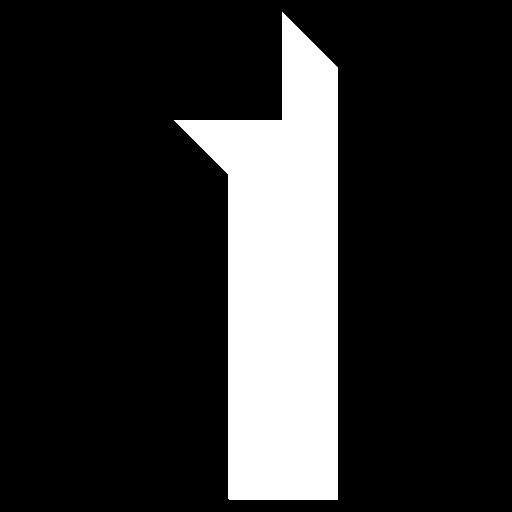}
    \includegraphics[width=0.15\linewidth]{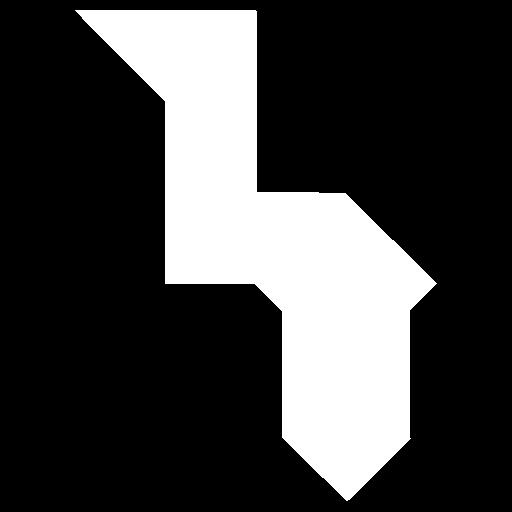}
    \\
    \includegraphics[width=0.15\linewidth]{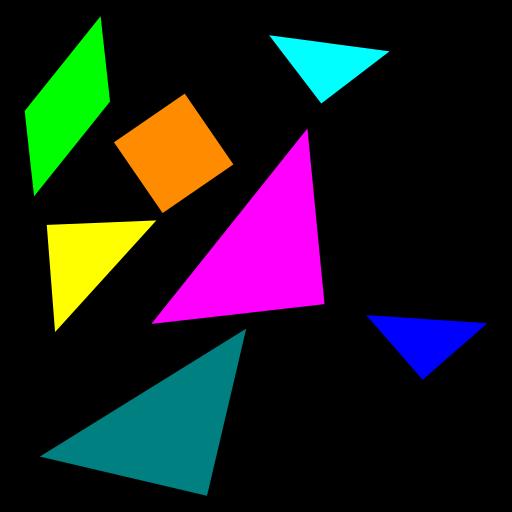}
    \includegraphics[width=0.15\linewidth]{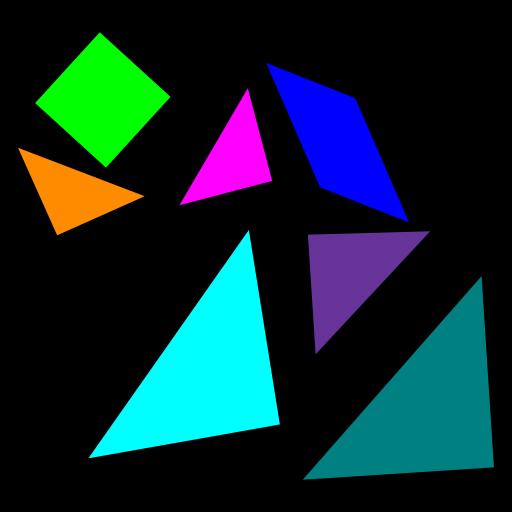}
    \includegraphics[width=0.15\linewidth]{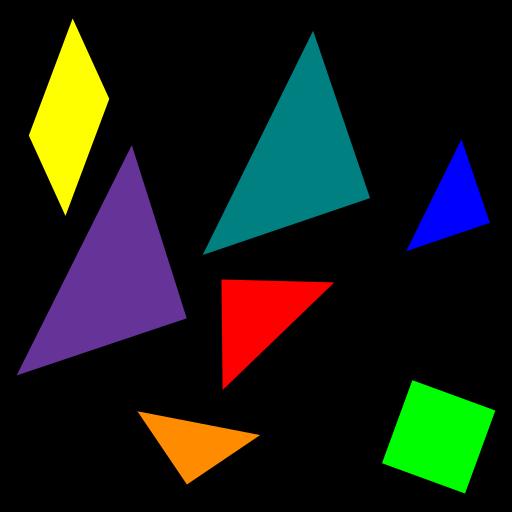}
    \includegraphics[width=0.15\linewidth]{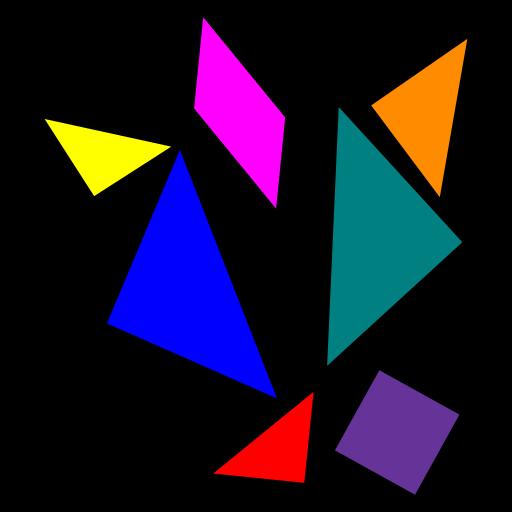}
    \includegraphics[width=0.15\linewidth]{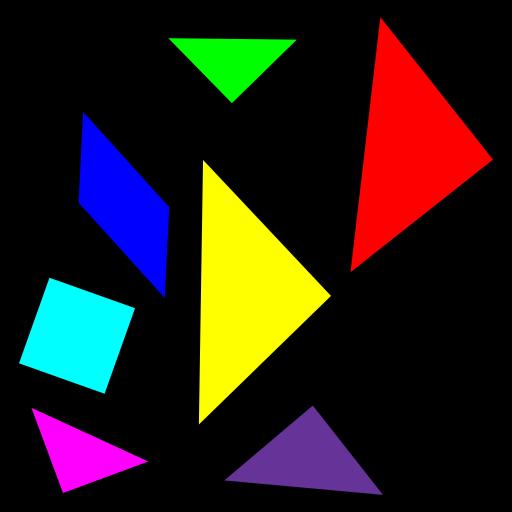}
    \includegraphics[width=0.15\linewidth]{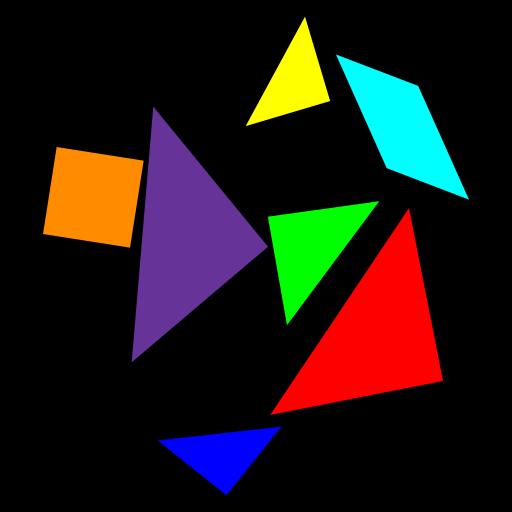}
    \caption{Visualization of \Tangram samples from the training set.}
\end{figure}

\textbf{Evaluation of \Tangram:} 
We use a deterministic evaluation pipeline to assess the correctness of the generated solutions with our diffusion models.
First, we quantize the color space by mapping each pixel to the nearest of 10 predefined color centroids (9 tan colors plus the black background).
We then leverage the input layout to define the ground-truth boundaries of the silhouette.
If the pixel-level mismatch between the generated image and the ground-truth layout exceeds 5\%, the sample is immediately assigned zero accuracy.

Subsequently, for samples passing the initial silhouette layout check, we use the \href{https://opencv.org/}{OpenCV} library to perform contour detection and polygonal approximation.
Based on the extracted polygons, we evaluate the sample against two metrics: 
\begin{enumerate}
    \item Shape accuracy (correctly-shaped): if the polygons maintain the correct area and proportions and the set of extracted shapes matches the original seven-tan composition.
    \item Accuracy (correctly-colored): building upon soft accuracy, if the color of each placed piece matches the identity of the corresponding tan provided in the input piece view.
\end{enumerate}

In Figure \ref{fig:tangram-invalid} of the main text, we show an example of a solution that would be classified with 100\% shape accuracy but with zero standard accuracy.

\subsection{\LogicFace: Logic Operators on Facial Features}\label{ap:logicface}

We challenge our spatial reasoning framework to abstract real-image patterns, and introduce \LogicFace as a structured logical puzzle where facial features must satisfy specific combinations based on logical operators.
The \LogicFace puzzle consists of three images, sorted in a row (Figure \ref{fig:example-logicface}), where the first two are the arguments ($A$ and $B$) and the third one is the answer ($S$).
Each image contains a clear centered human face, characterized with $\kappa$ binary attributes, named as $\boldsymbol{a},\boldsymbol{b},\boldsymbol{s}\in\{0,1\}^\kappa$ for $A$, $B$ and $S$, respectively.
For each attribute, the puzzle fixes an implicit logical operator $f_j:\{0,1\}\times \{0,1\} \to \{0,1\}$ such that the statement $f_j(\boldsymbol{a}_j,\boldsymbol{b}_j)=\boldsymbol{s}_j$ holds.

We used the face images and annotations from the CelebA-HQ dataset \citep{lee-etal-2020-maskgan} and selected $\kappa=4$ binary attributes: \textit{smiling}, \textit{male}, \textit{glasses} and \textit{young} to characterize each input image, with the \texttt{OR}, \texttt{AND}, \texttt{XOR} and \texttt{IMPLIES} logical operators.
These attributes and operators were fixed for all images of the dataset.
During training, the first and second panels are used as clues, while the last panel is corrupted to task the reasoner to learn how to generate facial attributes such that the logical operators hold.

\begin{wraptable}{r}{0.45\linewidth}
    \vspace{-12pt}
    \footnotesize\centering
    \setlength{\tabcolsep}{4pt}
    \begin{tabular}{ccc ccc ccc ccc}
        \toprule 
        \multicolumn{3}{c}{\texttt{OR}} & \multicolumn{3}{c}{\texttt{AND}} &\multicolumn{3}{c}{\texttt{XOR}} & \multicolumn{3}{c}{\texttt{IMPLIES}}\\
        \cmidrule(l{.5em}r{.5em}){1-3}\cmidrule(l{.5em}r{.5em}){4-6}\cmidrule(l{.5em}r{.5em}){7-9}\cmidrule(l{.5em}r{.5em}){10-12}
        \multicolumn{1}{c}{$A$} & \multicolumn{1}{c}{$B$} & \multicolumn{1}{c}{$S$} & \multicolumn{1}{c}{$A$} & \multicolumn{1}{c}{$B$} & \multicolumn{1}{c}{$S$} & \multicolumn{1}{c}{$A$} & \multicolumn{1}{c}{$B$} & \multicolumn{1}{c}{$S$} & \multicolumn{1}{c}{$A$} & \multicolumn{1}{c}{$B$} & \multicolumn{1}{c}{$S$} \\
        \midrule
         0 & 0 & 0 & 0 & 0 & 0 &  0 & 0 & 0 &  0 & 0 & 1\\ 
         0 & 1 & 1 & 0 & 1 & 0 & 0 & 1 & 1 & 0 & 1 & 1 \\
         1 & 0 & 1 & 1 & 0 & 0 & 1 & 0 & 1 & 1 & 0 & 0 \\
         1 & 1 & 1 & 1 & 1 & 1 & 1 & 1 & 0 & 1 & 1 & 1 \\
         \bottomrule
    \end{tabular}
    \caption{Logical operators between $A$ and $B$, yielding $S$. Since there are four operator and each operator has four valid statements, \LogicFace has $256$ possible configurations.}\label{tab:logic-statements}
    \vspace{-10pt}
\end{wraptable}

\begin{figure}\centering\small 
    \begin{subfigure}{0.485\linewidth}
        \begin{subfigure}{0.32\linewidth}
            \includegraphics[width=\linewidth]{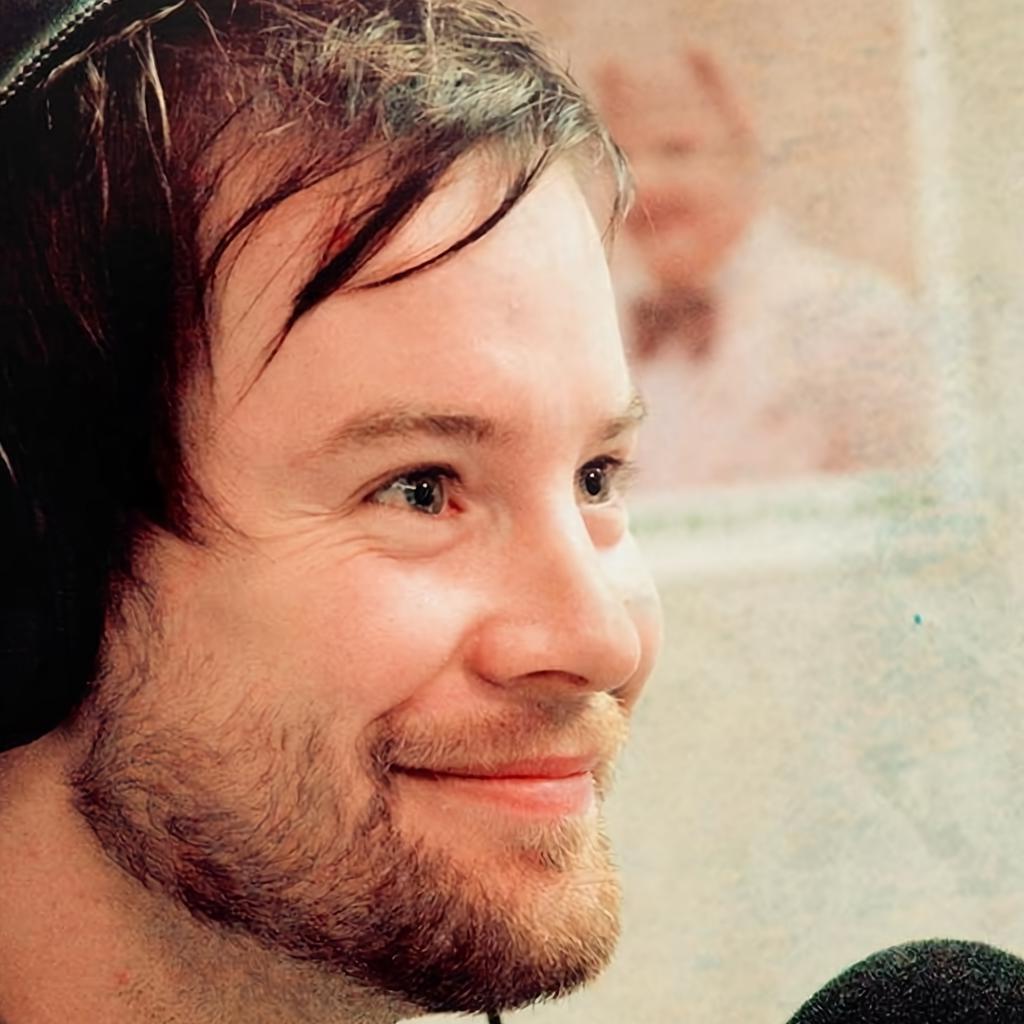}
            $\boldsymbol{a} = [0, 1, 0, 0]$
        \end{subfigure}
        \begin{subfigure}{0.32\linewidth}
            \includegraphics[width=\linewidth]{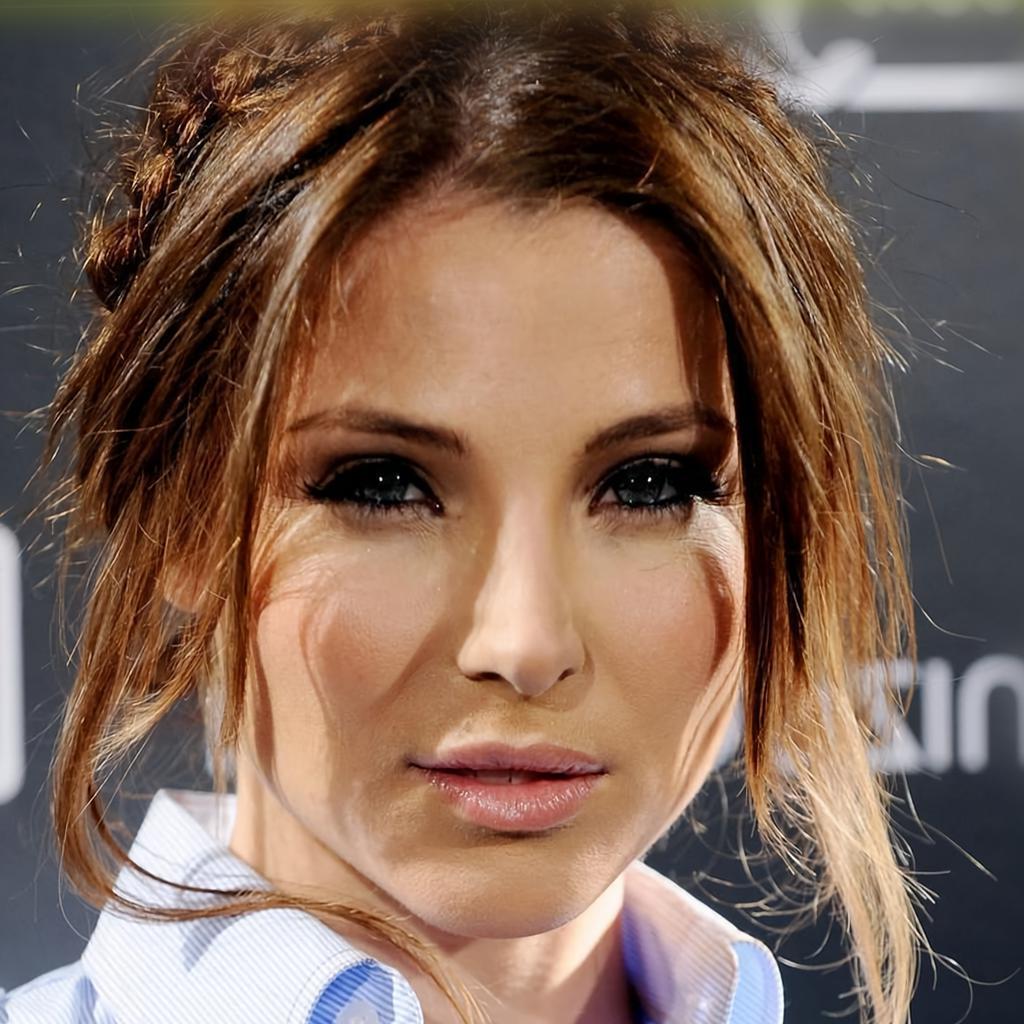}
            $\boldsymbol{b}=[0,0,0,1]$
        \end{subfigure}
        \begin{subfigure}{0.32\linewidth}
            \includegraphics[width=\linewidth]{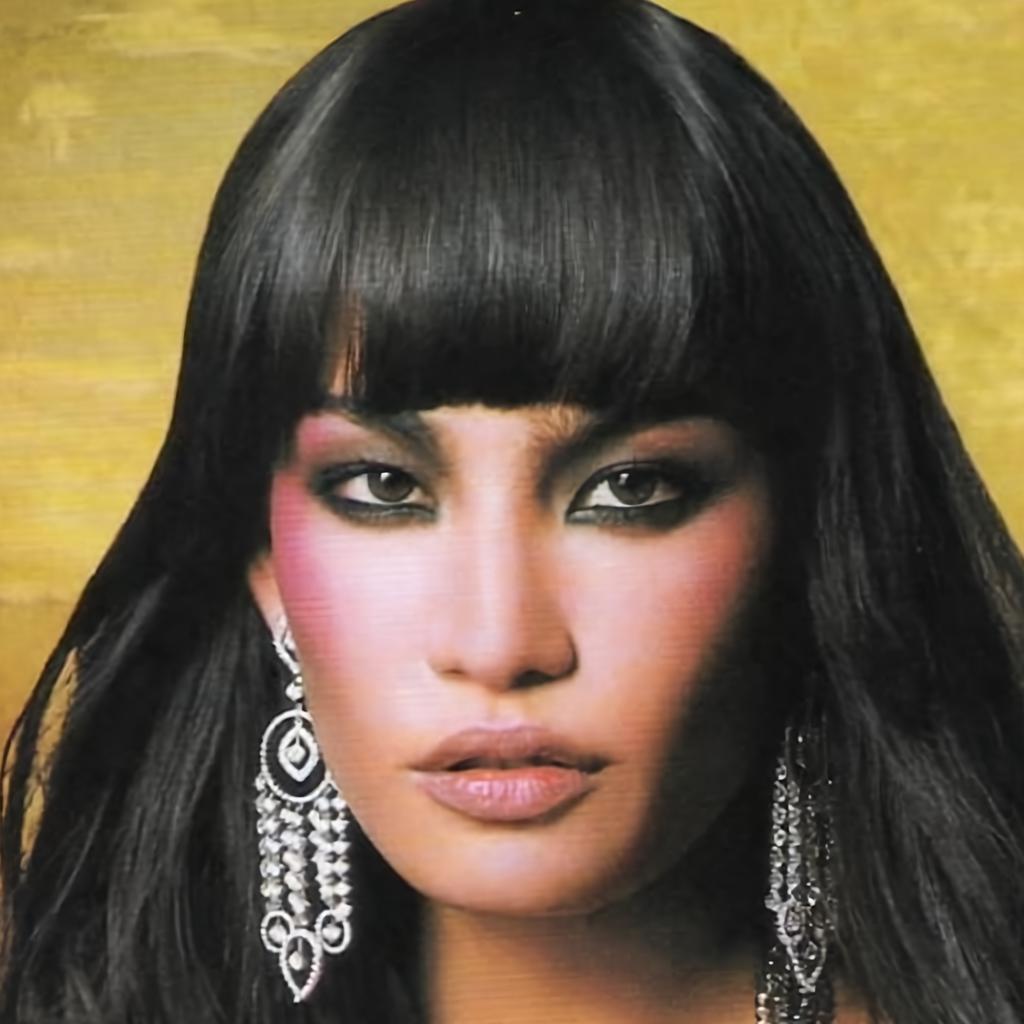}
            $\boldsymbol{s}=[0,0,0,1]$
        \end{subfigure}
    \end{subfigure}\hfill
    \begin{subfigure}{0.485\linewidth}
        \begin{subfigure}{0.32\linewidth}
            \includegraphics[width=\linewidth]{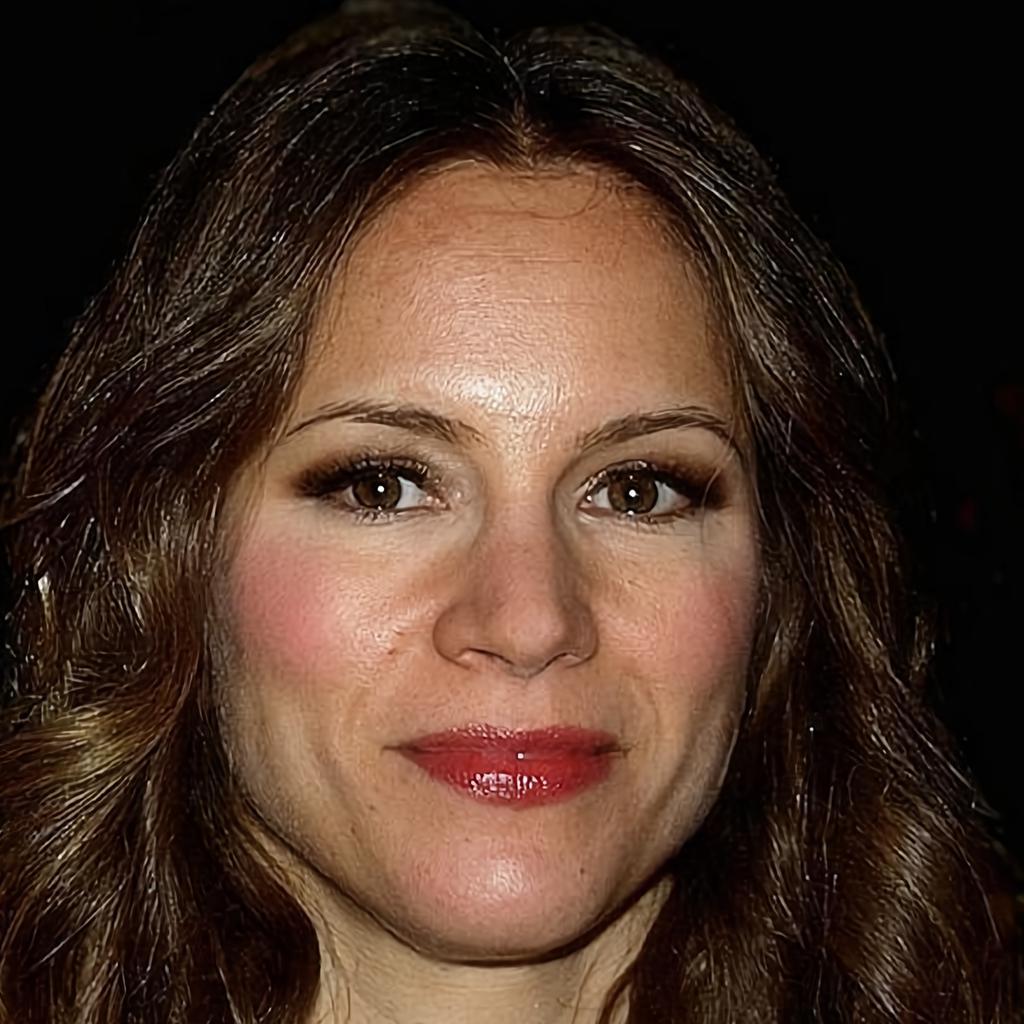}
            $\boldsymbol{a} = [0, 0, 0, 0]$
        \end{subfigure}
        \begin{subfigure}{0.32\linewidth}
            \includegraphics[width=\linewidth]{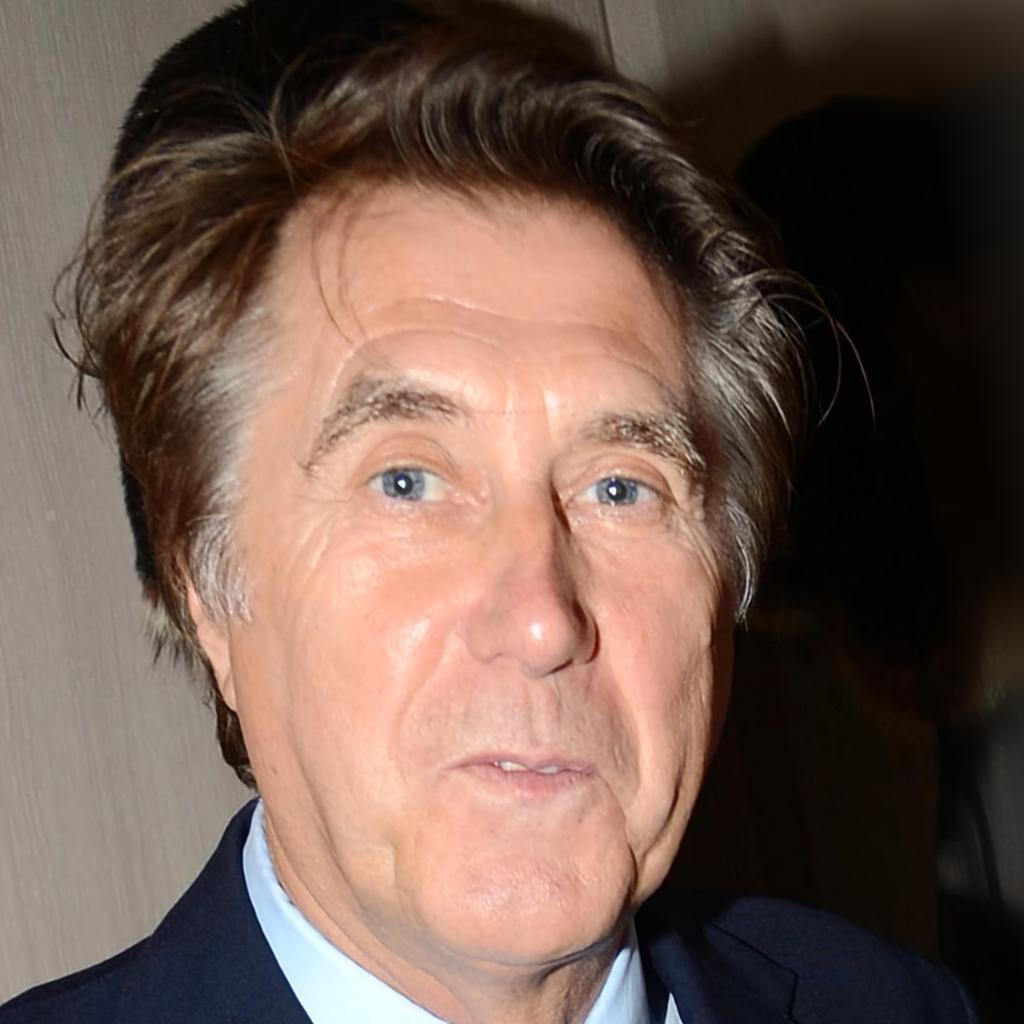}
            $\boldsymbol{b}=[0,1,0,0]$
        \end{subfigure}
        \begin{subfigure}{0.32\linewidth}
            \includegraphics[width=\linewidth]{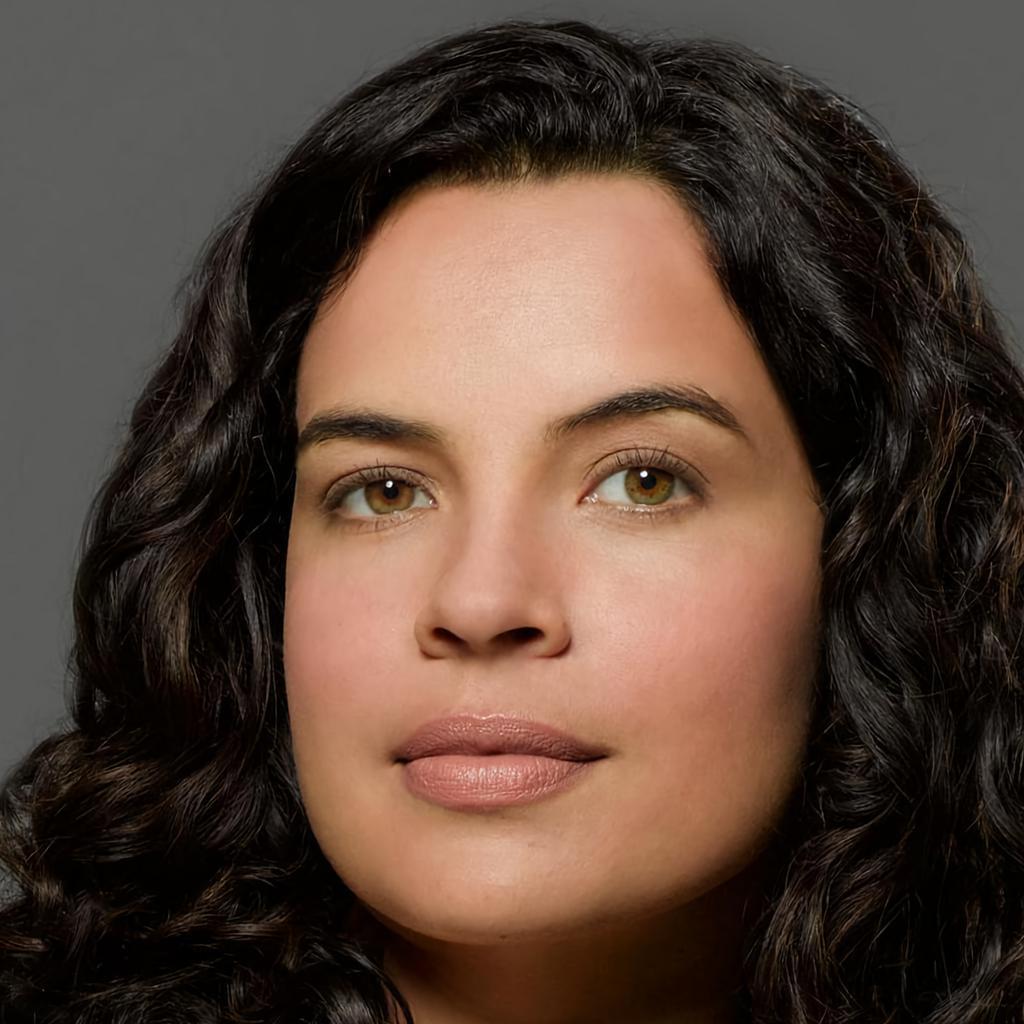}
            $\boldsymbol{s}=[0,0,0,1]$
        \end{subfigure}
    \end{subfigure}
    \\\vspace{0.25cm}
    \begin{subfigure}{0.485\linewidth}
        \begin{subfigure}{0.32\linewidth}
            \includegraphics[width=\linewidth]{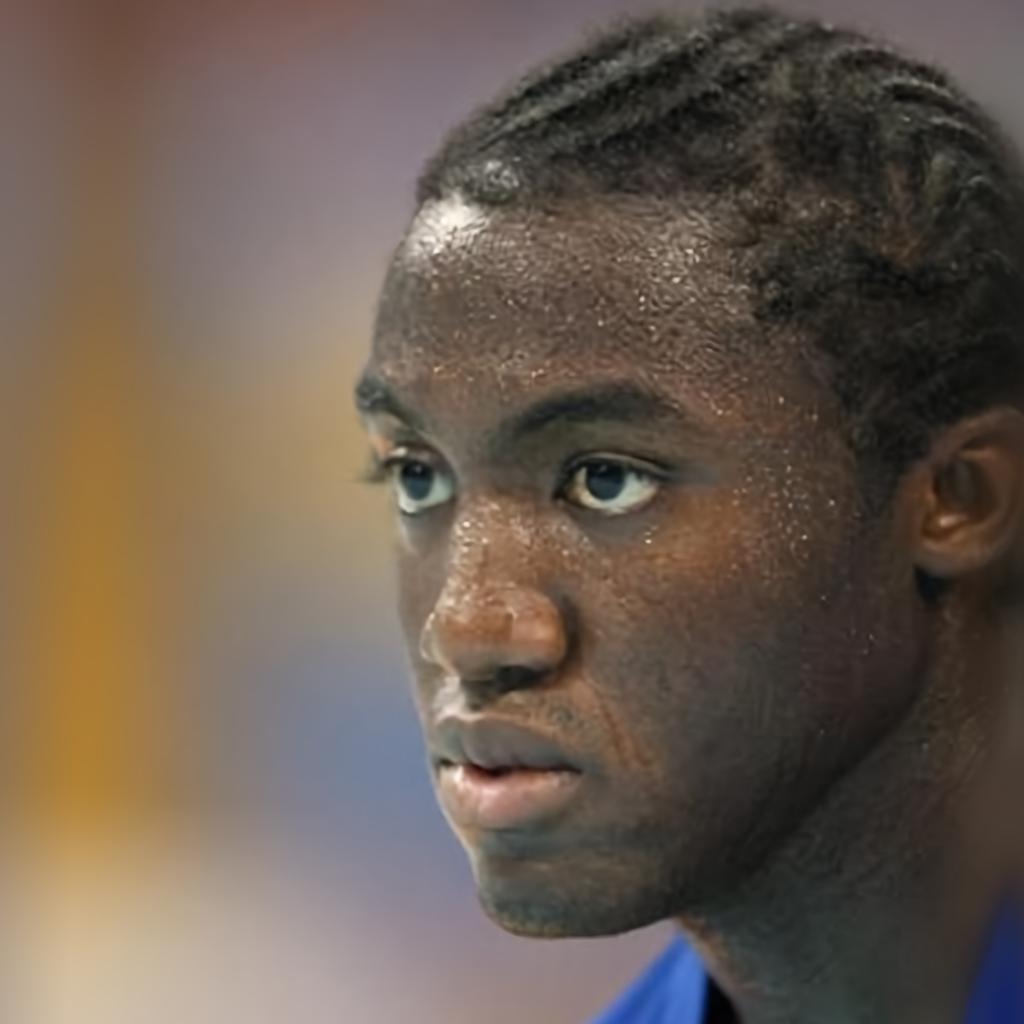}
            $\boldsymbol{a} = [0, 1, 0, 0]$
        \end{subfigure}
        \begin{subfigure}{0.32\linewidth}
            \includegraphics[width=\linewidth]{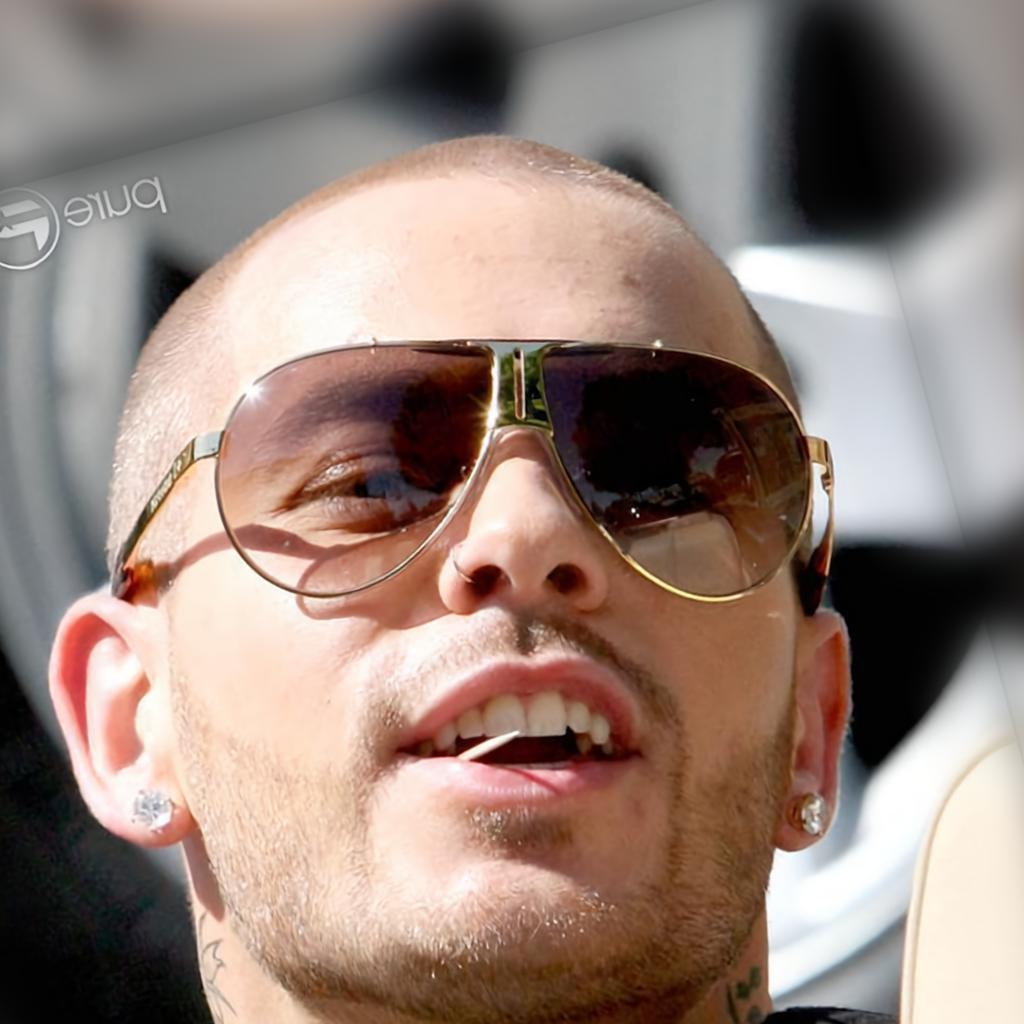}
            $\boldsymbol{b}=[0,1,1,1]$
        \end{subfigure}
        \begin{subfigure}{0.32\linewidth}
            \includegraphics[width=\linewidth]{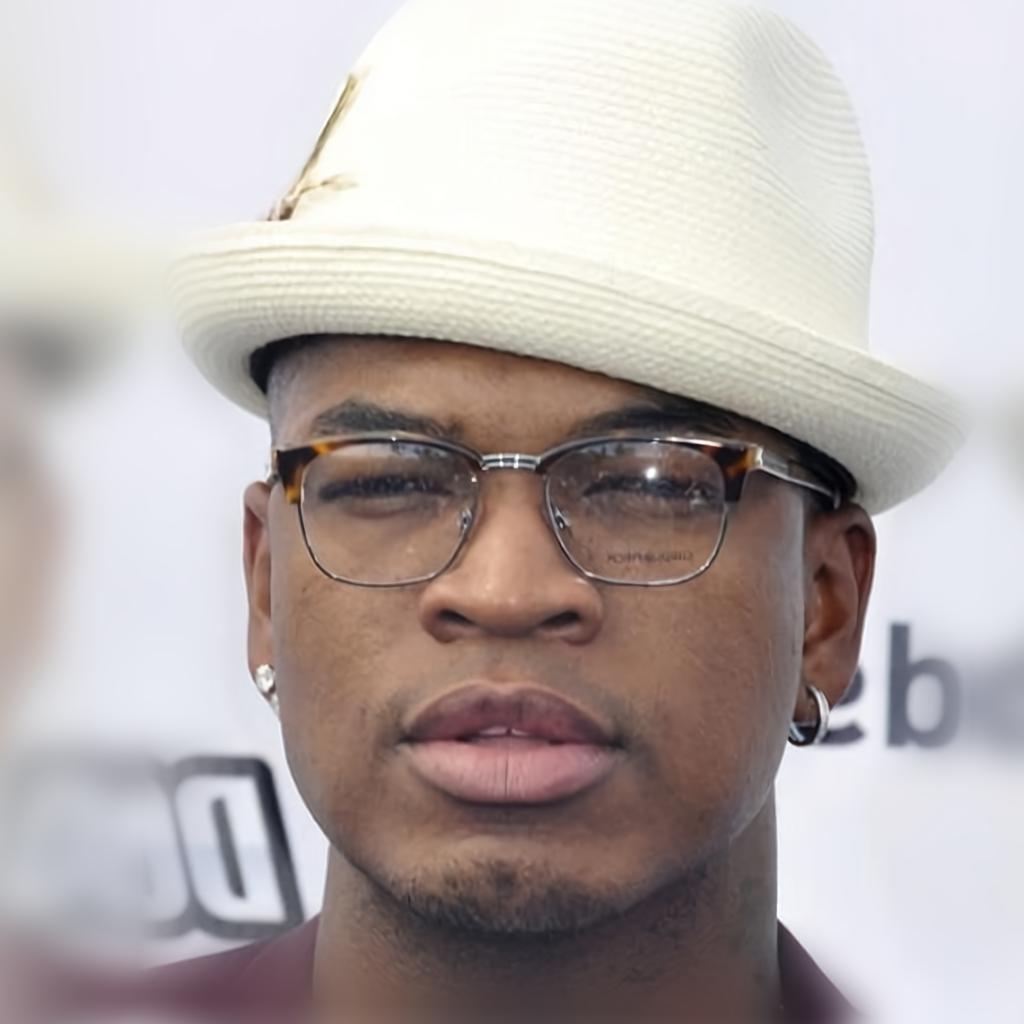}
            $\boldsymbol{s}=[0,1,1,1]$
        \end{subfigure}
    \end{subfigure}\hfill
    \begin{subfigure}{0.485\linewidth}
        \begin{subfigure}{0.32\linewidth}
            \includegraphics[width=\linewidth]{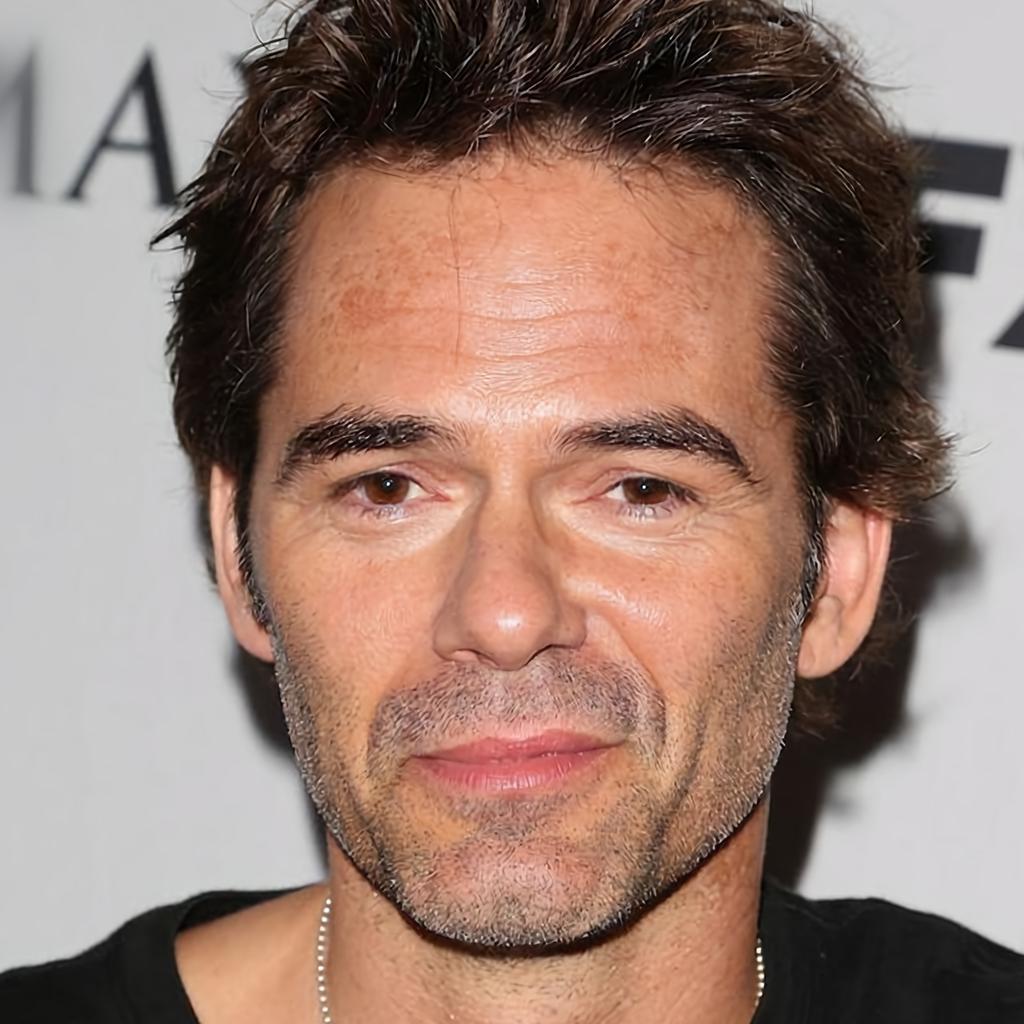}
            $\boldsymbol{a} = [0,1,0,1]$
        \end{subfigure}
        \begin{subfigure}{0.32\linewidth}
            \includegraphics[width=\linewidth]{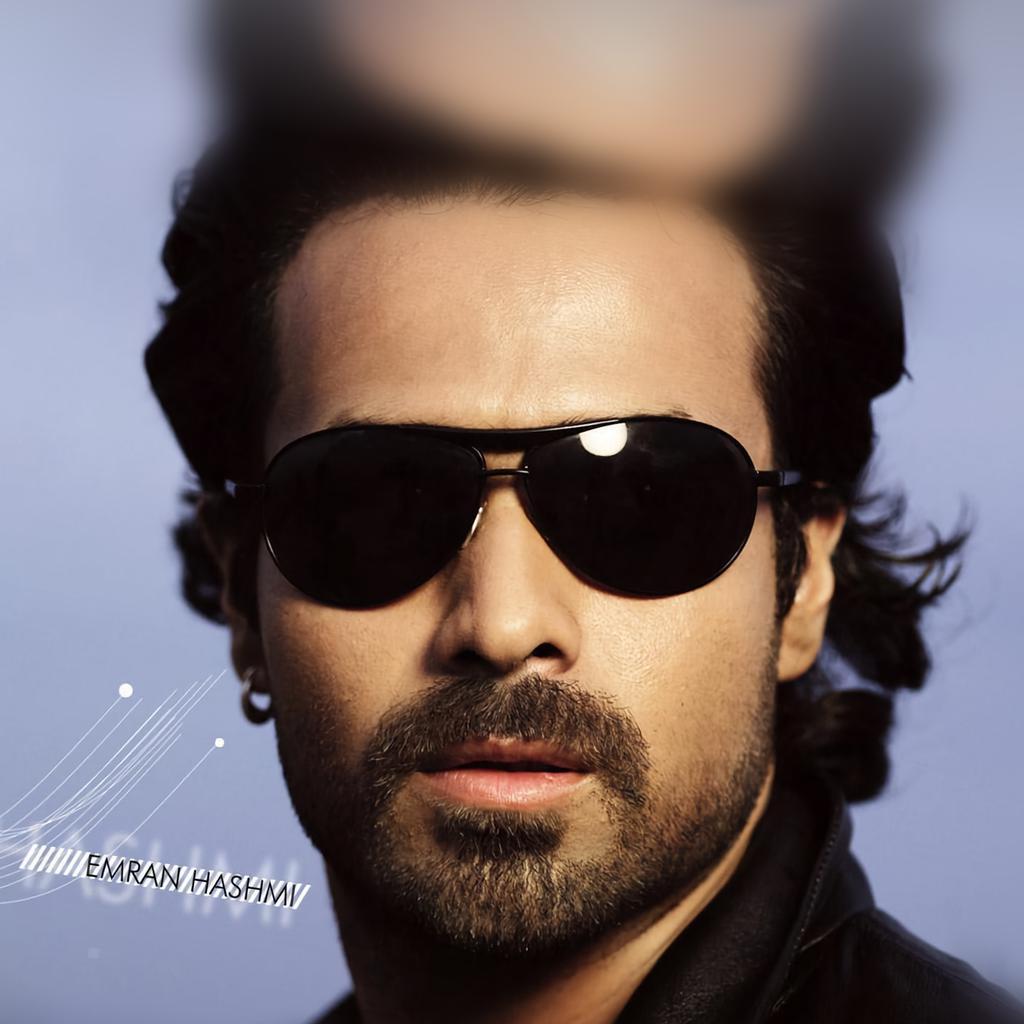}
            $\boldsymbol{b}=[0,1,1,0]$
        \end{subfigure}
        \begin{subfigure}{0.32\linewidth}
            \includegraphics[width=\linewidth]{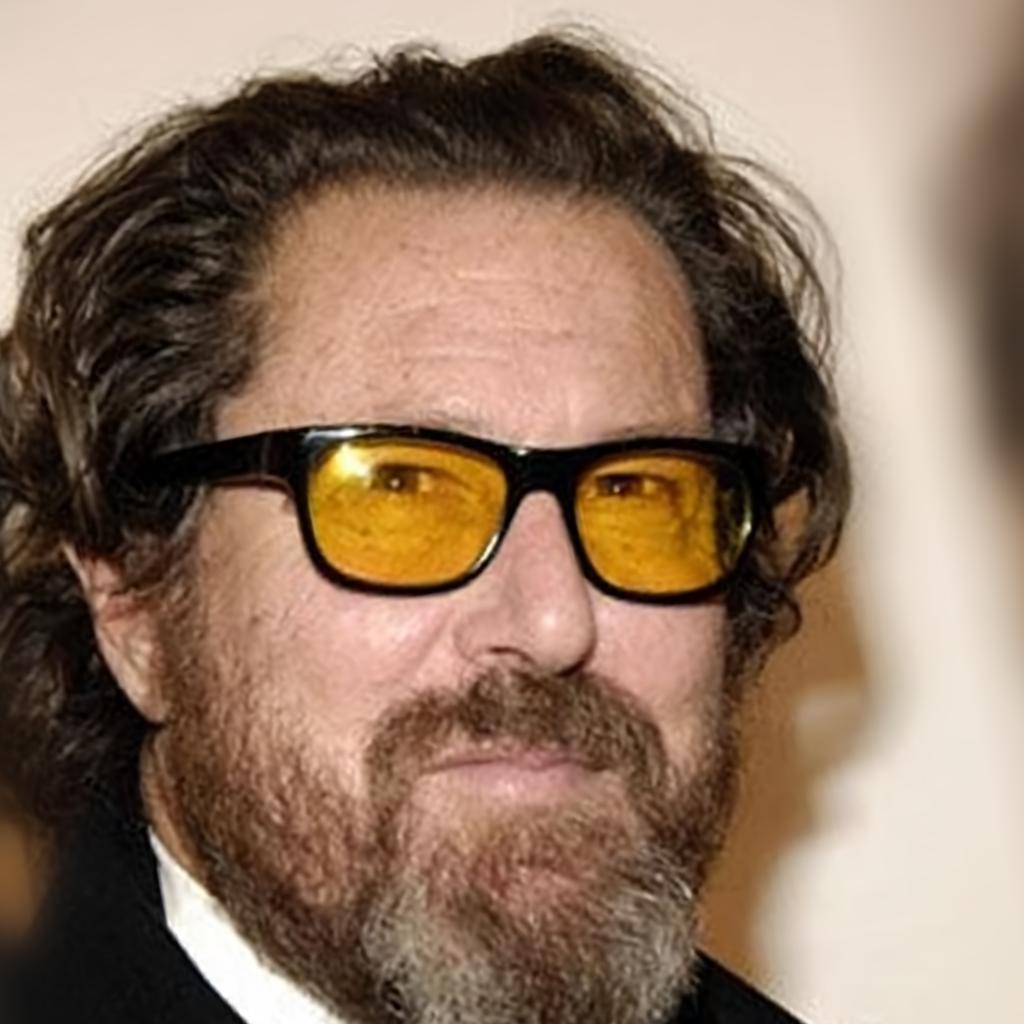}
            $\boldsymbol{s}=[0,1,1,0]$
        \end{subfigure}
    \end{subfigure}
    \caption{\LogicFace examples from the training set with the four binary features: \textit{smiling} (\texttt{OR}), \textit{male} (\texttt{AND}), \textit{glasses} (\texttt{XOR}), \textit{young} (\texttt{IMPLIES}).}
\end{figure}

\textbf{\LogicFace statistics:}
We selected the attributes of the \LogicFace puzzle after a manual examination of the CelebA annotations, where we found \textit{smiling}, \textit{male}, \textit{glasses} and \textit{young} to be less noisy than the other labels, as well as uncorrelated. Attributes like \textit{smiling} (48.21\%) and \textit{male} (41.67\%) are balanced, while \textit{glasses} and \textit{young} constitute the 6.5\%  and 77.36\% of the CelebA-HQ dataset, respectively.
To ensure a balanced number of \LogicFace configurations, we first uniformly sample one million \LogicFace configurations, and then select images whose attributes matched the specific statements.
Note that with this process, the \LogicFace dataset is not balanced in terms of images (indeed, faces with \emph{glasses} are considerably repeated since they need to cover the missing ratio of the CelebA-HQ dataset to create statements of the \texttt{XOR} operator).
However, since our purpose is to abstract on the combinations of glasses rather than the diversity of the generation, we opted to use data augmentation methods (e.g., horizontal flipping) to expand the variety of the samples, but acknowledging that our diffusion models might be recreating training set replicas, especially for unbalanced attributes like (not) \textit{young} and \textit{glasses}.

\textbf{Evaluation of \LogicFace:}
We trained a ResNet-18~\citep{he-etal-2016-resnet} with a binary multi-label objective corresponding to the selected attributes (\textit{smiling}, \textit{male}, \textit{glasses} and \textit{young}) on the CelebA-HQ dataset ($>98\%$ test accuracy).
At inference time, given a generated $\hat{S}$ and its predicted binary value $\hat{\boldsymbol{s}}$, the evaluator uses Table~\ref{tab:logic-statements} to validate the constraints.
Specifically, it individually evaluates $f_j(\boldsymbol{a}_j,\boldsymbol{b}_j)=\hat{\boldsymbol{s}}_j$ for the $j$-th feature and logical operator, and the overall correctness of a sample.

\section{Details on the Model Architectures}\label{ap:RDM}

Our spatial reasoning framework is composed of two models:
(i) the \emph{discoverer}, which is independently trained to disentangle objects in the scene through slot attention,
(ii) the \emph{reasoner}, which is a diffusion model tasked to model the conditional distribution $q(\mathbf{x}_0|\mathbf{x}_c)$ guided via \emph{relational embeddings} learned by a \emph{relational abstraction} module.
In what follows, we describe each module in detail, as well as the specific hyper-parameters used to train them.

\subsection{Architecture of the \emph{Discoverer}}\label{ap:discoverer_config}
The discoverer follows a similar architecture to the one proposed by~\citet{locatello-etal-2020-object}. 

\begin{itemize}[left=5pt]
    \item The slot encoder $\mathcal{E}:\mathbb{R}^{\texttt{H}\times \texttt{W}\times \texttt{c}}\to (\mathbb{R}^{m\times d}, \mathbb{R}^{m\times d})$ maps the input image to a set of hidden features $\mathbf{f}_{1:m}\in\mathbb{R}^{m\times d}$ through convolutional layers, alongside positional embeddings $\mathbf{p}_{1:m}\in\mathbb{R}^{m\times d}$, where $m=\texttt{hw}$ corresponds to the reduced spatial resolution. 
    \item The slot attention module, $\mathcal{S}:(\mathbb{R}^{m\times d},\mathbb{R}^{m\times d})\to \mathbb{R}^{k\times d}$, uses the extracted features ($\mathbf{f}_{1:m}$) and positions ($\mathbf{p}_{1:m}$) from the encoder to compute the slot embeddings ($\mathbf{s}_{1:k}$). This process begins by initializing the slots as $\mathbf{s}_{1:k}\sim\mathcal{N}(0,\mathbf{I})$, and iteratively refining them over $\eta=3$ iterations using a GRU conditioned on the input features and positions. The output is an updated set of slots $\mathbf{s}_{1:k}$, whose information is decoupled into object-specific features ($\tilde{\mathbf{f}}_{1:k}$) and positions ($\tilde{\mathbf{p}}_{1:k}$) by computing the dot product of $\mathbf{f}_{1:m}$ and $\mathbf{p}_{1:m}$ with the final attention matrix of the refinement, respectively. 
    This is formally outlined in Algorithm \ref{alg:slot-attention}.
    \item The slot decoder $\mathcal{D}:\mathbb{R}^{k\times d} \to \mathbb{R}^{\texttt{H}\times\texttt{W}\times\texttt{c}}$ is fed with the slot embeddings $\mathbf{s}_{1:k}$ to generate individual reconstructions and their corresponding importance masks. The decoder spatially broadcasts each slot embedding onto a 2D grid, augmented with positional embeddings $\mathbf{g}\in\mathbb{R}^{\texttt{h}\times\texttt{w}\times d}$. This is processed by convolutional layers to produce a candidate reconstruction per slot, named $\{\hat{\mathbf{x}}_{1:k}\}\in\mathbb{R}^{k\times\texttt{H}\times\texttt{W}\times\texttt{c}}$, with an associated mask, $\boldsymbol{\alpha}_i\in\mathbb{R}^{\texttt{H}\times\texttt{W}}$. The final output $\hat{\mathbf{x}}$ is computed as the weighted sum of the candidate reconstructions, where the weights are determined by a softmax normalization over the masks: $\smash{\hat{\mathbf{x}} = \sum_{i=1}^k\text{softmax}(\boldsymbol\alpha)_i \hat{\mathbf{x}}_i}$.
\end{itemize}

The full architecture is trained end-to-end with the mean squared error between the original input ${\mathbf{x}}$ and the final reconstruction $\hat{\mathbf{x}}$. 

\begin{figure}\scriptsize\centering
    \fontsize{8}{8}\selectfont 
    \begin{minipage}[t]{0.46\linewidth}
        \begin{algorithm}[H]
        \caption{\label{alg:slot-attention}Slot Attention module ($\mathcal{S}$).}
          \SetKwInOut{Input}{Input}
          \SetKwInOut{Parameters}{Params}
          \Input{$(\mathbf{f}_{1:m}, \mathbf{p}_{1:m})$ from $\mathcal{E}$.} 
          \Parameters{Projections $k, q, v$, GRU, MLP, LayerNorm.}
          $\mathbf{h}_{1:m} \leftarrow \text{LayerNorm}(\mathbf{f}_{1:m}+\mathbf{p}_{1:m})$\;
          $\mathbf{s}_{1:k}\sim\mathcal{N}(0,\mathbf{I})$\;
          \For{$i=1,\ldots, \eta$}{
            $\mathbf{s}_{1:k} \leftarrow \text{LayerNorm}(\mathbf{s}_{1:k})$\;
            $A \leftarrow \text{softmax}(k(\mathbf{h}_{1:m}) \cdot q(\mathbf{s}_{1:k})^{\scriptscriptstyle\top})\in\mathbb{R}^{m\times k}$\;
            $\mathbf{s}_{1:k}\leftarrow \text{GRU}(\mathbf{s}_{1:k};  A^\top\cdot v(\mathbf{h}_{1:m}))$\;
            $\mathbf{s}_{1:k} \leftarrow \mathbf{s}_{1:k} + \text{MLP}(\text{LayerNorm}(\mathbf{s}_{1:k}))$
          }
          $\tilde{\mathbf{f}}_{1:k}\leftarrow A^\top \mathbf{f}_{1:m}$\;
          $\tilde{\mathbf{p}}_{1:k}\leftarrow A^\top \mathbf{p}_{1:m}$\;
        \Return $\mathbf{s}_{1:k}$
        \end{algorithm}
    \end{minipage}
    \hfill
    \begin{minipage}[t]{0.52\linewidth}
    \begin{algorithm}[H]
      \caption{\label{alg:abstractor}Relational Abstraction ($\mathcal{R}$).}
      \SetKwInOut{Input}{Input}
      \SetKwInOut{Parameters}{Params}
      \Input{Latents $\mathbf{z} =[\tilde{\mathbf{f}}_{1:K};\tilde{\mathbf{p}}_{1:K}]$ from $\mathcal{S}$.} 
      \Parameters{Projections $k_\ell, q_\ell, v_\ell$ and $\text{FFN}_\ell$ per layer $\ell$.}
           $\mathbf{r}_{1:K} \leftarrow \tilde{\mathbf{p}}_{1:K}$\;
       \For{$\ell=1,\ldots,L$}{
            $\mathbf{v}_{1:K}\leftarrow  v_\ell(\mathbf{r}_{1:K})$\;
            $\mathbf{v}_{1:K} \leftarrow \text{softmax}(k_\ell(\tilde{\mathbf{f}}_{1:K})\cdot q_\ell(\tilde{\mathbf{f}}_{1:K})^{\scriptscriptstyle\top}) \mathbf{v}_{1:K} + \mathbf{v}_{1:K}$\;
            $\mathbf{v}_{1:K} \leftarrow\text{FFN}_\ell(\mathbf{v}_{1:K}) + \mathbf{v}_{1:K}$\;
            $\mathbf{v}_{1:K} \leftarrow \text{SelfAttention}(\mathbf{v}_{1:K}) + \mathbf{v}_{1:K}$\;
            $\mathbf{r}_{1:K} \leftarrow\text{FFN}_\ell(\mathbf{v}_{1:K}) + \mathbf{v}_{1:K}$\;
       }
       \Return $\mathbf{r}_{1:k}$
        \vspace{11.5pt}
    \end{algorithm}
    \end{minipage}
\end{figure}

\textbf{Scalable decoder:} 
To scale the number of slots $k$ to hundreds, we propose a \textit{scalable decoder} that avoids the memory-intensive process of generating full candidate reconstructions for every slot. 
Instead, this decoder estimates the attention masks (alpha weights) prior to the broadcasting step.

The mechanism uses a grid of positional embeddings, $\mathbf{g}\in\mathbb{R}^{\texttt{h}\times\texttt{w}\times d}$. Unlike the standard  decoder that broadcasts every slot to the full grid resolution, our approach computes the dot product between the grid and the slot embeddings $\mathbf{s}_{1:k}$ to derive an estimated mask: $\smash{\tilde{\boldsymbol{\alpha}}=\text{softmax}(\mathbf{g}\cdot \mathbf{s}_{1:k}^\top)}$. The positional grid is then updated by aggregating the slots according to these weights: $\smash{\tilde{\mathbf{s}}=\sum_{i=1...k}\tilde{\boldsymbol\alpha_i}\mathbf{s}_i}$. The weighted feature map $\tilde{\mathbf{s}}$ serves as the sole input for the subsequent decoder layers, effectively decoupling the reconstruction from the number of slots $k$.

\textbf{Training details:} 
We trained our \textit{Discoverer} module with AdamW  \citep{loshchilov-etal-2018-decoupled}, learning rate of $4\times 10^{-4}$, batch size of 64, and linear warmup of 10,000 steps followed by exponential decay; using mixed precision (bfloat) for accelerated training.

\textbf{Evaluation:}
We assess disentanglement of the \textit{Discoverer} by manually inspecting predicted samples, and additionally measure the reconstruction performance\footnote{Note that this accuracy is not correlated with the final reasoning accuracy since it only measures the pixel-level reconstruction quality of the autoencoder. In some cases, like \Tangram, the accuracy is low since most of the shapes are not sharply reconstructed, causing the evaluator to not properly estimate the polygons.} with the task-specific evaluators.
For \Akari, \Coldoku, \Tildoku and MNIST-Sudoku we use the cell accuracy (ratio of cells with matched predictions with the evaluator).
For \Tangram, \LogicFace and Counting Polygons we run the evaluator on the reconstructed image and measure if the output satisfies the logical constraints.

\subsection{Architecture of the \emph{Relational Abstraction} module}\label{ap:abstractor_config}

The relational abstraction module is designed to enrich slot representations with decoupled positional and relational information via attention mechanisms.
It leverages Slot Abstractors~\citep{mondal-etal-2024-slot} to map object-level information onto learnable \textit{relational embeddings}.
These embeddings are iteratively refined alongside the slots to produce high-level, position-aware representations.

The relational abstraction module, $\mathcal{R}: (\mathbb{R}^{K\times d}, \mathbb{R}^{K\times d}) \to \mathbb{R}^{K\times d}$, builds upon the output of the slot autoencoder.
As described in \S\ref{ap:discoverer_config}, the slots from the encoder $\mathcal{E}$ can be decoupled into object-specific features ($\tilde{\mathbf{f}}_{1:k}$) and positional embeddings ($\tilde{\mathbf{p}}_{1:k}$), which together compose the conditional latents for the diffusion model $\mathbf{z}^*=(\tilde{\mathbf{f}}^*_{1:k},\tilde{\mathbf{p}}^*_{1:k})$, where $*=(0,c)$.
This operation re-projects the consolidated slot information back into separate features and position subspaces. 

The slot abstractor module (Algorithm \ref{alg:abstractor}) learns relational information between slots by passing their feature and positional information through $L$ abstractor layers \citep{mondal-etal-2024-slot}, consisting of interleaved cross-attention, self-attention and MLP blocks with residual connections.
Through consecutive layers, the positional embeddings ($\tilde{\mathbf{p}}^*_{1:k}$) are updated by hierarchically pooling feature relations ($\tilde{\mathbf{f}}^*_{1:k}$), enabling the model to capture interactions between the discovered objects and learn contextualized positional-aware representations of their spatial dependencies.

\subsection{Configuration of the \emph{Reasoner}}\label{ap:reasoner_config}

We define the reasoning task  as modeling the conditional distribution $q(\mathbf{x}_0|\mathbf{x}_c)$, where $\mathbf{x}_0$ is a valid solution for the reasoning  task and $\mathbf{x}_c$ is the initial clue or statement.

The reasoner is a diffusion model that simultaneously learns two distributions: (i) the \emph{unguided} conditional distribution  $q(\mathbf{x}_0|\mathbf{x}_c, \mathcal{R}(\mathbf{z}^c))$, and (ii) the \emph{solution-guided} conditional distribution $q(\mathbf{x}_0|\mathbf{x}_c,\mathcal{R}(\mathbf{z}^c,\mathbf{z}^0))$, where the latent variables $\mathbf{z}^c$ and $\mathbf{z}^0$ correspond to the feature-position embeddings obtained with the discoverer module. Since the goal is to model $q(\mathbf{x}_0|\mathbf{x}_c)$ at inference time, ancestral sampling on $\mathbf{z}^0$ is not possible, so the \emph{unguided} distribution is the one sampled to generate a solution from the clue $\mathbf{x}_c$ and the latents $\mathbf{z}^c$.

\textbf{Forward process:} We define the diffusion process of the target distributions with continuous normalizing flows (CNFs), where the trajectory from data distribution $q_0(\mathbf{x})$ to the Gaussian prior $q_1=\mathcal{N}(0,\mathbf{I})$ is governed by Gaussian probability density paths \cite[Eq.~\eqref{eq:prob-density-path}]{lipman-etal-2023-flow}, whose parameters depend on $\mathbf{x}_0$ and a continuous time $t\in[0,1]$ such that convergence from $\mathbf{x}_0$ to the Gaussian prior is  guaranteed by fixing $\mu_0(\mathbf{x}_0) = \mathbf{x}_0$, $\sigma_0(\mathbf{x}_0) = \sigma_{\min}$, $\mu_1(\mathbf{x}_0)=0$ and $\sigma_1(\mathbf{x}_0)=1$. 
The \emph{flow} (Eq.~\eqref{eq:flow}) is a differentiable transformation to reshape samples from the prior to $p_t(\mathbf{x}|\mathbf{x}_0)$, and its associated conditional \emph{vector field} (Eq.~\eqref{eq:vector-field}) is approximated by a neural network.
\begin{equation}\label{eq:prob-density-path}
    p_t(\mathbf{x}|\mathbf{x}_0) = \mathcal{N}(\mathbf{x}|\mu_t(\mathbf{x}_0), \sigma_t(\mathbf{x}_0)^2 \mathbf{I}),
\end{equation}
\begin{equation}\label{eq:flow}
    \phi_t(\boldsymbol{\epsilon}, \mathbf{x}_0)= \mu_t(\mathbf{x}_0) + \sigma_t(\mathbf{x}_0) \boldsymbol{\epsilon},
\end{equation}
    \begin{equation}\label{eq:vector-field}
    v_t(\mathbf{x}|\mathbf{x}_0) =\frac{\sigma'_t(\mathbf{x}_0)}{\sigma_t(\mathbf{x}_0)} (\mathbf{x}-\mu_t(\mathbf{x}_0)) + \mu'_t(\mathbf{x}_0).
\end{equation}

We define the mean and variance of the probability paths as linear functions of $t$ (Eq.~\eqref{eq:mean-variance}):
\begin{equation}\label{eq:mean-variance}
    \mu_t(\mathbf{x}_0)=(1-t)\mathbf{x}_0, \quad \sigma_t(\mathbf{x}_0)=t,
\end{equation}
which effectively sets $\sigma_{\min}=0$.
Given this choice of linear functions, the training objective based on flow matching~\citep{lipman-etal-2023-flow} can be written as follows\footnote{We exclude in this notation the conditioning on the clue $\mathbf{x}_c$ and latents $\mathbf{z}^c$, $\mathbf{z}^0$ for better readability.}:
\begin{equation}\label{eq:loss-function}
    \mathcal{L}(\theta) = \mathbb{E}
    _{\mathbf{x}_0\sim q_0,\,\boldsymbol{\epsilon}\sim q_1,\,\mathbf{x}_t\sim p_t}
    \| p_\theta(\mathbf{x}_t, t) -  (\boldsymbol{\epsilon}-\mathbf{x}_0)\|^2
\end{equation}

\textbf{Sampling:} Since the transformation is governed by a continuous-time vector field, sampling can be framed as solving an ordinary differential equation (ODE). Specifically, once the neural network $p_\theta(\mathbf{x}_t, t)$ is trained to approximate the target vector field $v_t(\mathbf{x}|\mathbf{x}_0)$, we generate new samples by evolving the Gaussian noise $\boldsymbol{\epsilon} \sim \mathcal{N}(0, \mathbf{I})$ backward in time to $t=0$. The trajectory follows the ODE: 
\begin{equation}
    \frac{\mathrm{d}\mathbf{x}_t}{\textrm{d}t} = p_\theta(\mathbf{x}_t, t).
\end{equation}

In practice, this is computed with first-order discretization (Euler method) with step size $\mathrm{d}t$:
\begin{equation}
    \mathbf{x}_{t-\textrm{d}t} \approx \mathbf{x}_t + p_\theta(\mathbf{x}_t,t) \textrm{d}t .
\end{equation}

\textbf{Architecture details:}
Our denoiser module uses a standard attention-based U-Net \citep{ronneberger-etal-2015-unet} comprising residual convolutional blocks where input images are always stacked in the channel dimension. Self-attention is applied to the visual features exclusively within the final two blocks of the encoder. Each residual block is composed of two $3\times 3$ convolutions followed by a 2-factor pooling. The U-Net self-attention layers incorporate an extra cross-attention layer with shared KV projections to integrate the relational embeddings with the hidden features. Skip connections are implemented via channel concatenations between the corresponding encoder and decoder layers.

In preliminary experiments, we also explored the Diffusion Transformer (DiT) architecture~\citep{peebles-etal-2023-dit} as a backbone for the reasoning mechanism.
However, it proved difficult to optimize this architecture even under baseline configurations for these specific spatial tasks.
This observation also aligns with findings by~\citet{wever-etal-2025-spatial}, who noted similar challenges in reasoning-heavy domains.
We leave the development of hybrid architectures, leveraging the inductive biases of convolutional layers alongside the scalability of high-complexity Transformers, for future work.

\textbf{Training details:}
We use AdamW optimizer \citep{loshchilov-etal-2018-decoupled}, learning rate of $1\times 10^{-4}$, linear warm-up of 5,000 steps, and batch size of 128. For accelerated training, all denoisers are trained with the mixed precision (bfloat16). We train all models on the same number of steps.

\subsection{Model complexity and parameters}\label{ap:model-complexity-parameters}

We provide three size configurations for RDM to broadly investigate the impact of training small to large diffusion models for spatial reasoning tasks.
Previous work like SRM~\citep{wever-etal-2025-spatial} rely on large U-Net backbones (120M parameters) for puzzles like Sudoku, while others~\citep{goren-etal-2026-visual,lee-etal-2025-adaptive} provide empirical evidence that visual reasoning can still be unlocked with small architectures ($<$50M parameters).
We show two key findings through our experiments: (1) RDM is compatible with multiple backbone sizes, scaling from 30M to 120M parameters, and (2) larger architectures do not necessarily benefit reasoning and smaller architectures are preferable for faster convergence.
We highlight these findings as a positive result for the AI community in terms of reproducibility and feasibility.

We set RDM and capacity-matched baselines to three size configurations: small (\texttt{S}), medium (\texttt{M}) and large (\texttt{L}).
In both \texttt{S} and \texttt{M}, RDM maintains a compact discoverer module of $<$8M parameters (see Table \ref{tab:architecture-details-discoverer}) while the reasoner leverages 25M and 40M parameters, respectively (see Table \ref{tab:reasoner-architecture}).
In the \texttt{L} setting, we expanded the \textit{discoverer} dimension to $d=512$ to study the impact of richer slot embeddings, and matched the diffusion backbone in \texttt{L} to exactly mirror the model from~\citep{wever-etal-2025-spatial}.
Due to the computational overhead, we adapted the training limit to each task (see Table \ref{tab:reasoner-architecture}).

\begin{table}[h]\centering\footnotesize
    \caption{\label{tab:architecture-details-discoverer}Architecture, training and performance details of the \textit{Discoverer} in  configurations \texttt{S} and \texttt{M}.}
    \begin{tabular}{lccccccc}
        \toprule 
         & $k$ & $d$ & channels &  \#conv/layer & \#params. & \#steps & Accuracy \\
         \midrule 
         \textit{Slot-Attention} \\ 
         \Akari & 16 & 128 & [16, 32, 64, 128] & 1 &  1.4M & 100K & 99.58 \\ 
         \Coldoku & 16 & 128 & [32, 64, 128, 256] & 1 & 5.4M & 400K & 88.19\\ 
         \Tangram & 7  & 128 & [16, 32, 64, 128] & 1 & 1.4M & 300K & 16.34\\
         \LogicFace & 6 & 256 & [32, 64, 128, 256] & 2 & 7.6M & 200K & 72.54 \\
         \Tildoku & 20 & 128 & [16, 32, 64, 128] & 2 & 7.6M & 400K & 61.15\\
         MNIST-Sudoku & 16 & 128 & [16, 32, 64, 128] & 1 & 1.4M & 200K & 84.73 \\
         Counting Polygons  & 12 & 256 & [32, 64, 128, 256] & 1 & 5.4M & 300K & 80.67\\
         \midrule 
         \textit{Scalable Slot-Attention} \\
         \Akari & 121 & 128 & [16, 32, 64, 128] & 1 & 1.4M & 100K & 100\\
         \Coldoku & 81 & 128 & [16, 32, 64, 128] & 1 & 1.4M & 200K & 99.52\\
         \bottomrule
    \end{tabular}
\end{table}

\begin{table}[h]\centering\footnotesize
    \caption{\label{tab:reasoner-architecture}Architecture and training details of the \textit{Reasoner}.}
    \setlength{\tabcolsep}{3pt}
    \resizebox{\linewidth}{!}{
    \begin{tabular}{lccccccc}
        \toprule 
         &  $d$ & channels & \#conv/layer & \#params of $\mathcal{D}$ & \#params of $\mathcal{R}$ & \#params of U-Net &  \# steps \\
         \midrule 
         RDM\textsuperscript{\texttt{[S]}} & 128 & [64, 128, 256, 512] & [2, 2, 1, 1] & (Table \ref{tab:architecture-details-discoverer})& 4M  & 25M & 300K\\
         RDM\textsuperscript{\texttt{[M]}} &  256 & [64, 128, 256, 512] & [2, 2, 2, 2] &(Table \ref{tab:architecture-details-discoverer}) & 4M  & 35M & 300K\\
         \midrule 
        RDM\textsuperscript{\texttt{[L]}} \\
        \Akari & 512 & [128, 256, 512, 512] & 2 & 40M & 15M & 120M & 40K \\
        \Coldoku & 512 & [128, 256, 512, 512] & 2 & 40M & 15M & 120M & 45K  \\
        \Tangram & 512 & [128, 256, 512, 512] & 2 & 40M & 15M & 120M & 80K  \\
        \LogicFace & 512 & [128, 256, 512, 512] & 2 & 40M & 15M & 120M & 40K  \\
        \Tildoku & 512 & [128, 256, 512, 512] & 2 & 40M & 15M & 120M & 45K  \\
        \bottomrule
    \end{tabular}}\\
\end{table}

\textbf{Computational resources}: All models were implemented in PyTorch 2.9 and trained using the CUDA 13.0 toolkit.
Our experiments were conducted across multiple compute clusters (e.g., Austrian Scientific Computing infrastructure), consisting of NVIDIA L40, A40, A100 and H100 GPU nodes.

\section{Details on the Experimental Results}\label{ap:results}

\subsection{Results of the \emph{Discoverer}}

The reconstruction quality of the discoverer module is evaluated with the \textit{cell accuracy} in the grid-like puzzles (\Akari and \Coldoku).
For \Tangram, the evaluation pipeline described in \S\ref{ap:tangram} is also used to evaluate the discoverer.
For \LogicFace, we use the ResNet-18 classifier (\S\ref{ap:logicface}) on the reconstructed samples to match the real binary vector with the one predicted by the classifier.

\begin{figure}[h!]\centering\small
    \begin{subfigure}{0.1\linewidth}\centering
        \caption*{Real}
    \end{subfigure}
    \begin{subfigure}{0.1\linewidth}\centering
        \caption*{Output}
    \end{subfigure}
    \begin{subfigure}{0.1\linewidth}\centering
        \caption*{Slot 1}
    \end{subfigure}
    \begin{subfigure}{0.1\linewidth}\centering
        \caption*{Slot 2}
    \end{subfigure}
    \begin{subfigure}{0.1\linewidth}\centering
        \caption*{Slot 3}
    \end{subfigure}
        \begin{subfigure}{0.1\linewidth}\centering
        \caption*{Slot 4}
    \end{subfigure}
        \begin{subfigure}{0.1\linewidth}\centering
        \caption*{Slot 5}
    \end{subfigure}
    \begin{subfigure}{0.1\linewidth}\centering
        \caption*{Slot 6}
    \end{subfigure}
    \begin{subfigure}{0.1\linewidth}\centering
        \caption*{Slot 7}
    \end{subfigure}
    \includegraphics[width=\linewidth]{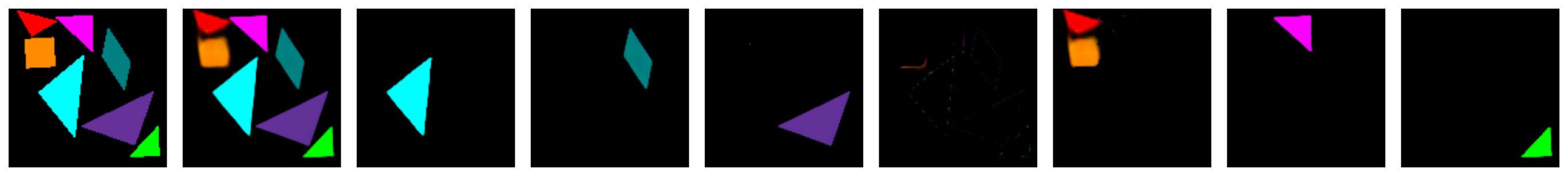}
    \includegraphics[width=\linewidth]{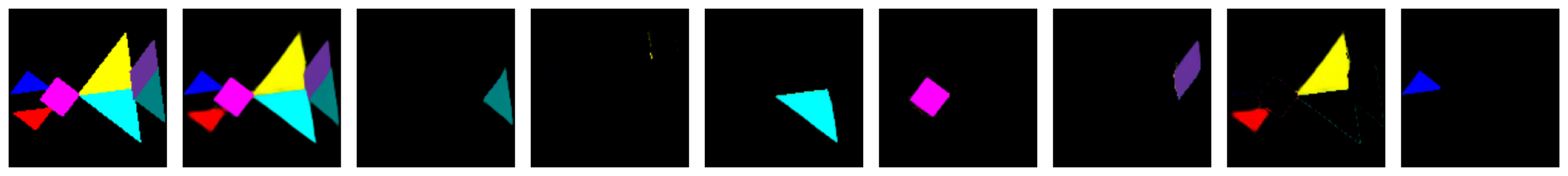}
    \includegraphics[width=\linewidth]{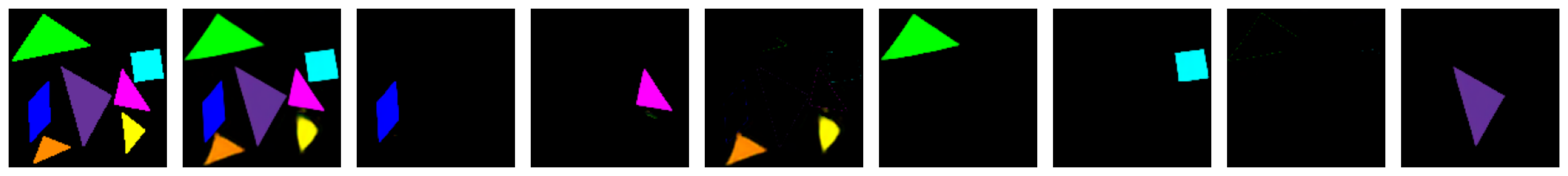}
    \includegraphics[width=\linewidth]{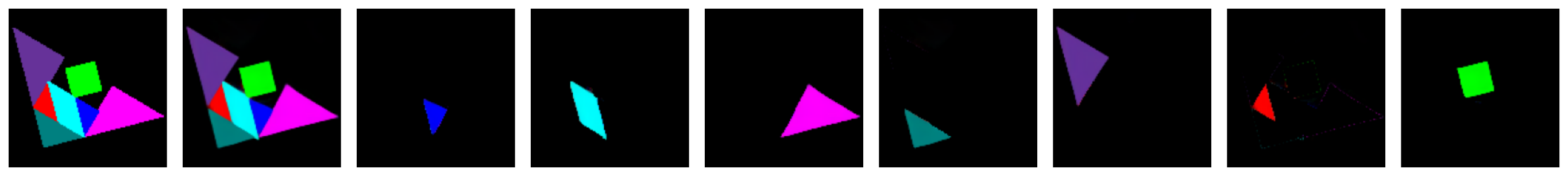}
    \includegraphics[width=\linewidth]{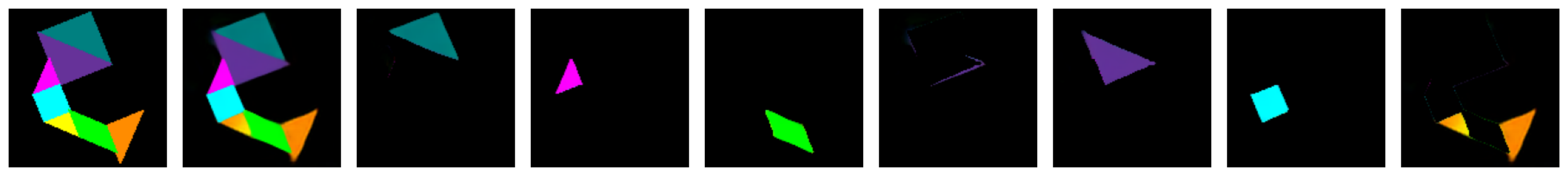}
    \includegraphics[width=\linewidth]{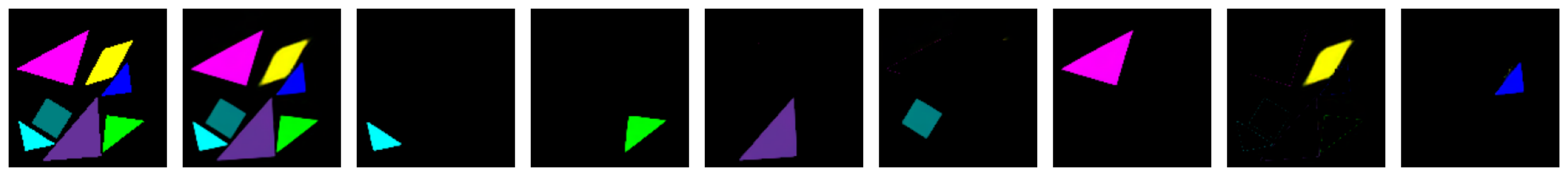}
    \caption{\label{fig:slot-tangram}Disentanglement of the tans on \Tangram inputs.}
\end{figure}

In \Akari reconstruction accuracies achieved 97\% and 98\% using 16 and 121 slots, respectively.
For \Coldoku, digit reconstruction accuracy reached 82.64\% with 16 slots and 99.79\% with 81 slots.
However, color accuracy was significantly lower at 40.41\% and 72.05\%.
We observed that the gap in color reconstruction correlates with observed blurriness in the outputs.
Because the slot autoencoder uses a variational architecture with a simple convolutional decoder, it is prone to producing blurred digits with imprecise color values.
Consequently, more errors occurred when reconstructed colors fell between two palette values, leading to misclassification by evaluator.
We observed a similar behavior for \Tangram, where the discoverer only achieved 57\% of accuracy with our evaluation pipeline (\S\ref{ap:tangram}).
In practice the model showed real disentanglement (Figure \ref{fig:slot-tangram}), although reconstruction struggled to define sharp borders for the figures, leading to errors when estimating polygons with the deterministic evaluator.
For \LogicFace, we evaluated the discoverer's reconstructions and obtained 95\% of accuracy in the selected features (\textit{smiling}, \textit{male}, \textit{glasses} and \textit{young}). 

\begin{figure}[h!]\centering\small
    \begin{subfigure}{0.1\linewidth}
        \caption*{Output}
        \includegraphics[width=\linewidth]{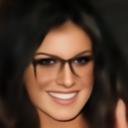}
        \includegraphics[width=\linewidth]{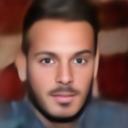}
        \includegraphics[width=\linewidth]{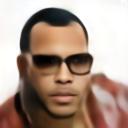}
        \includegraphics[width=\linewidth]{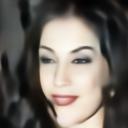}
    \end{subfigure}
    \begin{subfigure}{0.1\linewidth}
        \caption*{Slot 1}
        \includegraphics[width=\linewidth]{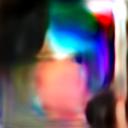}
        \includegraphics[width=\linewidth]{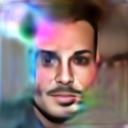}
        \includegraphics[width=\linewidth]{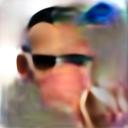}
        \includegraphics[width=\linewidth]{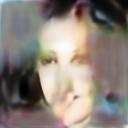}
    \end{subfigure}
        \begin{subfigure}{0.1\linewidth}
        \caption*{Slot 2}
        \includegraphics[width=\linewidth]{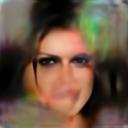}
        \includegraphics[width=\linewidth]{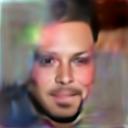}
        \includegraphics[width=\linewidth]{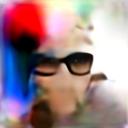}
        \includegraphics[width=\linewidth]{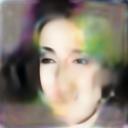}
    \end{subfigure}
        \begin{subfigure}{0.1\linewidth}
        \caption*{Slot 3}
        \includegraphics[width=\linewidth]{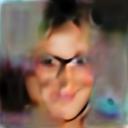}
        \includegraphics[width=\linewidth]{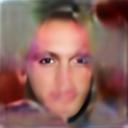}
        \includegraphics[width=\linewidth]{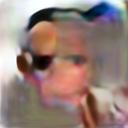}
        \includegraphics[width=\linewidth]{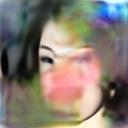}
    \end{subfigure}
    \begin{subfigure}{0.1\linewidth}
        \caption*{Slot 4}
        \includegraphics[width=\linewidth]{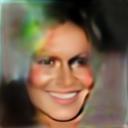}
        \includegraphics[width=\linewidth]{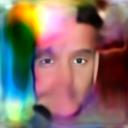}
        \includegraphics[width=\linewidth]{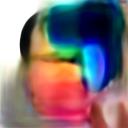}
        \includegraphics[width=\linewidth]{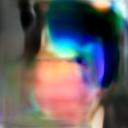}

    \end{subfigure}
        \begin{subfigure}{0.1\linewidth}
        \caption*{Slot 5}
        \includegraphics[width=\linewidth]{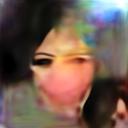}
        \includegraphics[width=\linewidth]{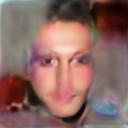}
        \includegraphics[width=\linewidth]{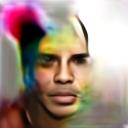}
        \includegraphics[width=\linewidth]{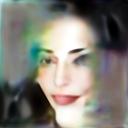}
    \end{subfigure}
        \begin{subfigure}{0.1\linewidth}
        \caption*{Slot 6}
        \includegraphics[width=\linewidth]{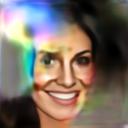}
        \includegraphics[width=\linewidth]{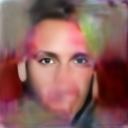}
        \includegraphics[width=\linewidth]{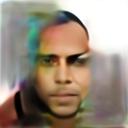}
        \includegraphics[width=\linewidth]{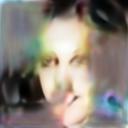}
    \end{subfigure}
    \caption{\label{fig:slot-logicface}Disentanglement of the facial attributes on \LogicFace inputs.}
\end{figure}

\textbf{Quality of object disentanglement:} Figures \ref{fig:slot-tangram} and \ref{fig:slot-logicface} illustrate the decomposition performance of the trained slot autoencoders on the \Tangram and \LogicFace datasets. While Slot Attention is theoretically designed to ensure perfect object disentanglement, we observed that for \Tangram, certain slots remained inactive while others encapsulated multiple tans. We hypothesize that because pieces vary in color and exhibit slight scale variances across samples, the slots may have fit specific chromatic patterns rather than individual geometric entities. Consequently, the model occasionally groups co-occurring tans into a single slot instead of maintaining a strictly one-to-one object correspondence.

Slot visualizations for \LogicFace (Figure \ref{fig:slot-logicface}) depicts the inherent difficulty of object disentanglement in real images.
We observed that salient, well-defined features from the CelebA-HQ dataset—such as glasses, smiling, and facial hair, are consistently isolated within specific slot embeddings.
Conversely, the remaining slots struggle to decouple diffuse attributes, including skin tone and hair textures, or to model the high-variance backgrounds.
This suggests that while Slot Attention effectively identifies discrete semantic anchors, it faces challenges in partitioning continuous or highly irregular global structures into distinct object-centric representations.
We still find slot embeddings beneficial for reasoning, since key features were already represented into slots (glasses and smiling), while the others could be associated with other attributes (e.g., beard with \textit{male}, gray hair with \textit{not young}).

\subsection{Results of the \emph{Reasoner}}\label{ap:reasoner_results}

\begin{figure}[h]\centering
    \begin{subfigure}[t]{\linewidth}
        \begin{subfigure}[t]{0.49\linewidth}
            \begin{subfigure}[t]{0.66\linewidth}
                \begin{center}\textit{clues}\end{center}
                \includegraphics[width=0.49\linewidth]{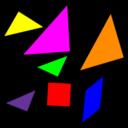}
                \includegraphics[width=0.49\linewidth]{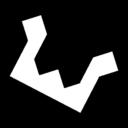}
            \end{subfigure}
            \hfill
            \begin{subfigure}[t]{0.33\linewidth}
                \begin{center}Generation\end{center}%
                \includegraphics[width=0.98\linewidth]{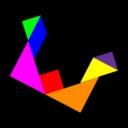}
            \end{subfigure}
        \end{subfigure}
        \hfill 
        \begin{subfigure}[t]{0.49\linewidth}
            \begin{subfigure}[t]{0.66\linewidth}
                \begin{center}\textit{clues}\end{center}
                \includegraphics[width=0.49\linewidth]{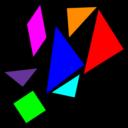}
                \includegraphics[width=0.49\linewidth]{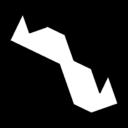}
            \end{subfigure}
            \hfill
            \begin{subfigure}[t]{0.33\linewidth}
                \begin{center}Generation\end{center}
                \includegraphics[width=0.98\linewidth]{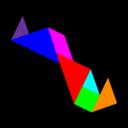}
            \end{subfigure}
        \end{subfigure}
        \\
        \begin{subfigure}[t]{0.49\linewidth}
            \begin{subfigure}[t]{0.66\linewidth}
                \includegraphics[width=0.49\linewidth]{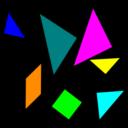}
                \includegraphics[width=0.49\linewidth]{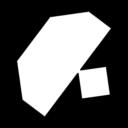}
            \end{subfigure}
            \hfill
            \begin{subfigure}[t]{0.33\linewidth}
                \includegraphics[width=0.98\linewidth]{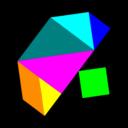}
            \end{subfigure}
        \end{subfigure}
        \hfill 
        \begin{subfigure}[t]{0.49\linewidth}
            \begin{subfigure}[t]{0.66\linewidth}
                \includegraphics[width=0.49\linewidth]{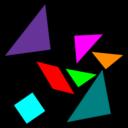}
                \includegraphics[width=0.49\linewidth]{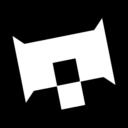}
            \end{subfigure}
            \hfill
            \begin{subfigure}[t]{0.33\linewidth}
                \includegraphics[width=0.98\linewidth]{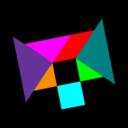}
            \end{subfigure}
        \end{subfigure}
        \caption{Examples from the \textit{Baseline}.}
        \vspace{10pt}
    \end{subfigure}
    \begin{subfigure}[t]{\linewidth}
        \begin{subfigure}[t]{0.49\linewidth}
            \begin{subfigure}[t]{0.66\linewidth}
                \begin{center}\textit{clues}\end{center}
                \includegraphics[width=0.49\linewidth]{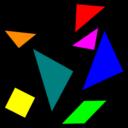}
                \includegraphics[width=0.49\linewidth]{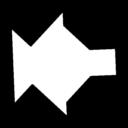}
            \end{subfigure}
            \hfill
            \begin{subfigure}[t]{0.33\linewidth}
                \begin{center}Generation\end{center}
                \includegraphics[width=0.98\linewidth]{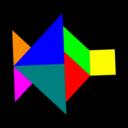}
            \end{subfigure}
        \end{subfigure}
        \hfill 
        \begin{subfigure}[t]{0.49\linewidth}
            \begin{subfigure}[t]{0.66\linewidth}
                \begin{center}\textit{clues}\end{center}
                \includegraphics[width=0.49\linewidth]{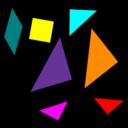}
                \includegraphics[width=0.49\linewidth]{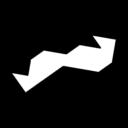}
            \end{subfigure}
            \hfill
            \begin{subfigure}[t]{0.33\linewidth}
                \begin{center}Generation\end{center}
                \includegraphics[width=0.98\linewidth]{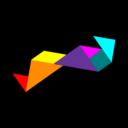}
            \end{subfigure}
        \end{subfigure}
        \\
        \begin{subfigure}[t]{0.49\linewidth}
            \begin{subfigure}[t]{0.66\linewidth}
                \includegraphics[width=0.49\linewidth]{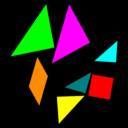}
                \includegraphics[width=0.49\linewidth]{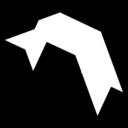}
            \end{subfigure}
            \hfill
            \begin{subfigure}[t]{0.33\linewidth}
                \includegraphics[width=0.98\linewidth]{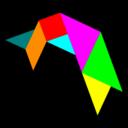}
            \end{subfigure}
        \end{subfigure}
        \hfill 
        \begin{subfigure}[t]{0.49\linewidth}
            \begin{subfigure}[t]{0.66\linewidth}
                \includegraphics[width=0.49\linewidth]{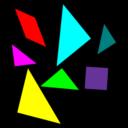}
                \includegraphics[width=0.49\linewidth]{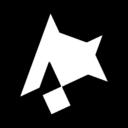}
            \end{subfigure}
            \hfill
            \begin{subfigure}[t]{0.33\linewidth}
                \includegraphics[width=0.98\linewidth]{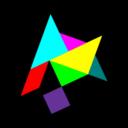}
            \end{subfigure}
        \end{subfigure}
        \\
        \begin{subfigure}[t]{0.49\linewidth}
            \begin{subfigure}[t]{0.66\linewidth}
                \includegraphics[width=0.49\linewidth]{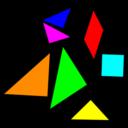}
                \includegraphics[width=0.49\linewidth]{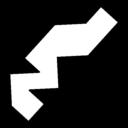}
            \end{subfigure}
            \hfill
            \begin{subfigure}[t]{0.33\linewidth}
                \includegraphics[width=0.98\linewidth]{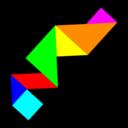}
            \end{subfigure}
        \end{subfigure}
        \hfill 
        \begin{subfigure}[t]{0.49\linewidth}
            \begin{subfigure}[t]{0.66\linewidth}
                \includegraphics[width=0.49\linewidth]{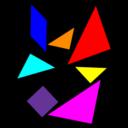}
                \includegraphics[width=0.49\linewidth]{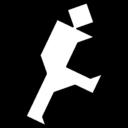}
            \end{subfigure}
            \hfill
            \begin{subfigure}[t]{0.33\linewidth}
                \includegraphics[width=0.98\linewidth]{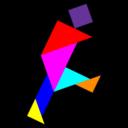}
            \end{subfigure}
        \end{subfigure}
        \\
        \begin{subfigure}[t]{0.49\linewidth}
            \begin{subfigure}[t]{0.66\linewidth}
                \includegraphics[width=0.49\linewidth]{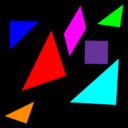}
                \includegraphics[width=0.49\linewidth]{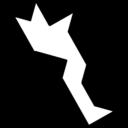}
            \end{subfigure}
            \hfill
            \begin{subfigure}[t]{0.33\linewidth}
                \includegraphics[width=0.98\linewidth]{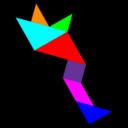}
            \end{subfigure}
        \end{subfigure}
        \hfill 
        \begin{subfigure}[t]{0.49\linewidth}
            \begin{subfigure}[t]{0.66\linewidth}
                \includegraphics[width=0.49\linewidth]{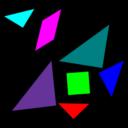}
                \includegraphics[width=0.49\linewidth]{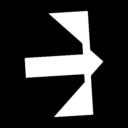}
            \end{subfigure}
            \hfill
            \begin{subfigure}[t]{0.33\linewidth}
                \includegraphics[width=0.98\linewidth]{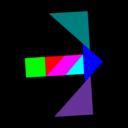}
            \end{subfigure}
        \end{subfigure}
        \\
        \begin{subfigure}[t]{0.49\linewidth}
            \begin{subfigure}[t]{0.66\linewidth}
                \includegraphics[width=0.49\linewidth]{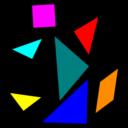}
                \includegraphics[width=0.49\linewidth]{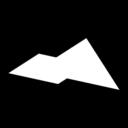}
            \end{subfigure}
            \hfill
            \begin{subfigure}[t]{0.33\linewidth}
                \includegraphics[width=0.98\linewidth]{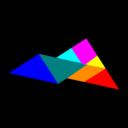}
            \end{subfigure}
        \end{subfigure}
        \hfill 
        \begin{subfigure}[t]{0.49\linewidth}
            \begin{subfigure}[t]{0.66\linewidth}
                \includegraphics[width=0.49\linewidth]{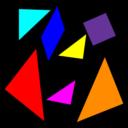}
                \includegraphics[width=0.49\linewidth]{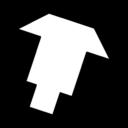}
            \end{subfigure}
            \hfill
            \begin{subfigure}[t]{0.33\linewidth}
                \includegraphics[width=0.98\linewidth]{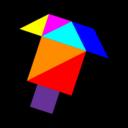}
            \end{subfigure}
        \end{subfigure}
        \\
        \begin{subfigure}[t]{0.49\linewidth}
            \begin{subfigure}[t]{0.66\linewidth}
                \includegraphics[width=0.49\linewidth]{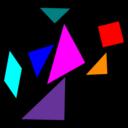}
                \includegraphics[width=0.49\linewidth]{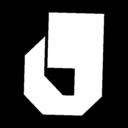}
            \end{subfigure}
            \hfill
            \begin{subfigure}[t]{0.33\linewidth}
                \includegraphics[width=0.98\linewidth]{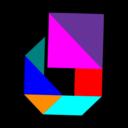}
            \end{subfigure}
        \end{subfigure}
        \hfill 
        \begin{subfigure}[t]{0.49\linewidth}
            \begin{subfigure}[t]{0.66\linewidth}
                \includegraphics[width=0.49\linewidth]{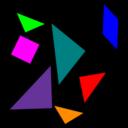}
                \includegraphics[width=0.49\linewidth]{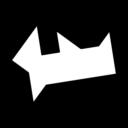}
            \end{subfigure}
            \hfill
            \begin{subfigure}[t]{0.33\linewidth}
                \includegraphics[width=0.98\linewidth]{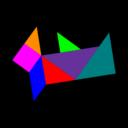}
            \end{subfigure}
        \end{subfigure}
        \caption{Examples from RDM\textsuperscript{\texttt{[L]}} with the \textit{clue} based abstraction conditioning approach.}
    \end{subfigure}
    \caption{\label{fig:tangram-generations}Examples of generated \Tangram solutions.}
\end{figure}

\begin{figure}\centering\small 
    \begin{subfigure}[t]{\linewidth}
        \begin{subfigure}[t]{0.16\linewidth}\centering 
            \caption*{$A$}
            \includegraphics[width=\linewidth]{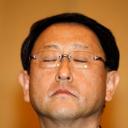}
            $\boldsymbol{a} =[0, 1, 1, 0]$
        \end{subfigure}
        \begin{subfigure}[t]{0.16\linewidth}\centering 
            \caption*{$B$}
            \includegraphics[width=\linewidth]{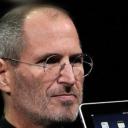}
            $\boldsymbol{b} = [0,1,1,0]$
        \end{subfigure}
        \begin{subfigure}[t]{0.16\linewidth}\centering 
            \caption*{$S$}
            \includegraphics[width=\linewidth]{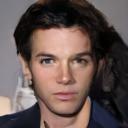}
            $\boldsymbol{s} = [0,1,0,1]$
        \end{subfigure}
        \hfill 
        \begin{subfigure}[t]{0.16\linewidth}\centering 
            \caption*{$A$}
            \includegraphics[width=\linewidth]{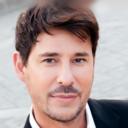}
            $\boldsymbol{a} =[0, 1, 0, 1]$
        \end{subfigure}
        \begin{subfigure}[t]{0.16\linewidth}\centering 
            \caption*{$B$}
            \includegraphics[width=\linewidth]{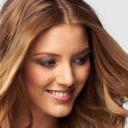}
            $\boldsymbol{b} = [1,0,0,1]$
        \end{subfigure}
        \begin{subfigure}[t]{0.16\linewidth}\centering 
            \caption*{$S$}
            \includegraphics[width=\linewidth]{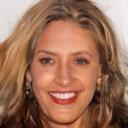}
            $\boldsymbol{s} = [1,0,0,1]$
        \end{subfigure}
        \\[8pt]
        \begin{subfigure}[t]{0.16\linewidth}\centering 
            \caption*{$A$}
            \includegraphics[width=\linewidth]{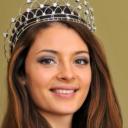}
            $\boldsymbol{a} =[1, 0, 0, 1]$
        \end{subfigure}
        \begin{subfigure}[t]{0.16\linewidth}\centering 
            \caption*{$B$}
            \includegraphics[width=\linewidth]{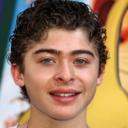}
            $\boldsymbol{b} = [1,1,0,1]$
        \end{subfigure}
        \begin{subfigure}[t]{0.16\linewidth}\centering 
            \caption*{$S$}
            \includegraphics[width=\linewidth]{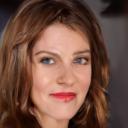}
            $\boldsymbol{s} = [1,0,0,1]$
        \end{subfigure}
        \hfill 
        \begin{subfigure}[t]{0.16\linewidth}\centering 
            \caption*{$A$}
            \includegraphics[width=\linewidth]{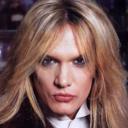}
            $\boldsymbol{a} =[0, 1, 0, 0]$
        \end{subfigure}
        \begin{subfigure}[t]{0.16\linewidth}\centering 
            \caption*{$B$}
            \includegraphics[width=\linewidth]{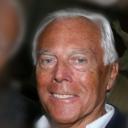}
            $\boldsymbol{b} = [1,1,0,0]$
        \end{subfigure}
        \begin{subfigure}[t]{0.16\linewidth}\centering 
            \caption*{$S$}
            \includegraphics[width=\linewidth]{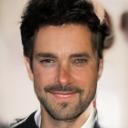}
            $\boldsymbol{s} = [1,1,0,1]$
        \end{subfigure}
        \caption{Examples from the \textit{Baseline}.}
        \vspace{10pt}
    \end{subfigure}\\
    \begin{subfigure}[t]{\linewidth}
        \begin{subfigure}[t]{0.16\linewidth}\centering 
            \caption*{$A$}
            \includegraphics[width=\linewidth]{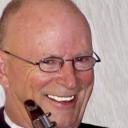}
            $\boldsymbol{a} =[1, 1, 1, 0]$
        \end{subfigure}
        \begin{subfigure}[t]{0.16\linewidth}\centering 
            \caption*{$B$}
            \includegraphics[width=\linewidth]{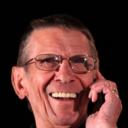}
            $\boldsymbol{b} = [1,1,1,0]$
        \end{subfigure}
        \begin{subfigure}[t]{0.16\linewidth}\centering 
            \caption*{$S$}
            \includegraphics[width=\linewidth]{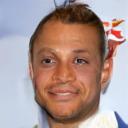}
            $\boldsymbol{s} = [1,1,0,1]$
        \end{subfigure}
        \hfill 
        \begin{subfigure}[t]{0.16\linewidth}\centering 
            \caption*{$A$}
            \includegraphics[width=\linewidth]{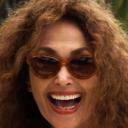}
            $\boldsymbol{a} =[1, 0,1,1]$
        \end{subfigure}
        \begin{subfigure}[t]{0.16\linewidth}\centering 
            \caption*{$B$}
            \includegraphics[width=\linewidth]{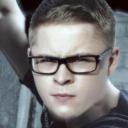}
            $\boldsymbol{b} = [0,1,1,1]$
        \end{subfigure}
        \begin{subfigure}[t]{0.16\linewidth}\centering 
            \caption*{$S$}
            \includegraphics[width=\linewidth]{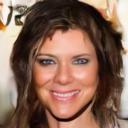}
            $\boldsymbol{s} = [1,0,0,1]$
        \end{subfigure}
        \\
        \begin{subfigure}[t]{0.16\linewidth}\centering 
            \caption*{$A$}
            \includegraphics[width=\linewidth]{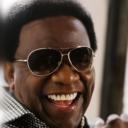}
            $\boldsymbol{a} =[1,1,1,0]$
        \end{subfigure}
        \begin{subfigure}[t]{0.16\linewidth}\centering 
            \caption*{$B$}
            \includegraphics[width=\linewidth]{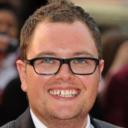}
            $\boldsymbol{b} = [1,1,1,1]$
        \end{subfigure}
        \begin{subfigure}[t]{0.16\linewidth}\centering 
            \caption*{$S$}
            \includegraphics[width=\linewidth]{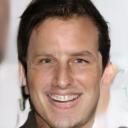}
            $\boldsymbol{s} = [1,1,0,1]$
        \end{subfigure}
        \hfill 
        \begin{subfigure}[t]{0.16\linewidth}\centering 
            \caption*{$A$}
            \includegraphics[width=\linewidth]{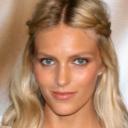}
            $\boldsymbol{a} =[0,0,0,1]$
        \end{subfigure}
        \begin{subfigure}[t]{0.16\linewidth}\centering 
            \caption*{$B$}
            \includegraphics[width=\linewidth]{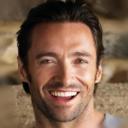}
            $\boldsymbol{b} = [1,1,0,1]$
        \end{subfigure}
        \begin{subfigure}[t]{0.16\linewidth}\centering 
            \caption*{$S$}
            \includegraphics[width=\linewidth]{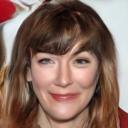}
            $\boldsymbol{s} = [1,0,0,1]$
        \end{subfigure}
        \\
        \begin{subfigure}[t]{0.16\linewidth}\centering 
            \caption*{$A$}
            \includegraphics[width=\linewidth]{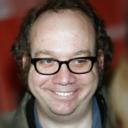}
            $\boldsymbol{a} =[1, 1, 1, 0]$
        \end{subfigure}
        \begin{subfigure}[t]{0.16\linewidth}\centering 
            \caption*{$B$}
            \includegraphics[width=\linewidth]{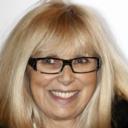}
            $\boldsymbol{b} = [1,0,1,0]$
        \end{subfigure}
        \begin{subfigure}[t]{0.16\linewidth}\centering 
            \caption*{$S$}
            \includegraphics[width=\linewidth]{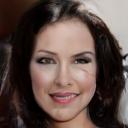}
            $\boldsymbol{s} = [1,0,0,1]$
        \end{subfigure}
        \hfill 
        \begin{subfigure}[t]{0.16\linewidth}\centering 
            \caption*{$A$}
            \includegraphics[width=\linewidth]{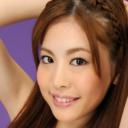}
            $\boldsymbol{a} =[1,0,0,1]$
        \end{subfigure}
        \begin{subfigure}[t]{0.16\linewidth}\centering 
            \caption*{$B$}
            \includegraphics[width=\linewidth]{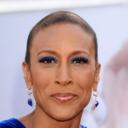}
            $\boldsymbol{b} = [1,0,0,0]$
        \end{subfigure}
        \begin{subfigure}[t]{0.16\linewidth}\centering 
            \caption*{$S$}
            \includegraphics[width=\linewidth]{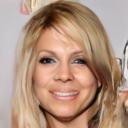}
            $\boldsymbol{s} = [1,0,0,0]$
        \end{subfigure}
        \\ 
        \begin{subfigure}[t]{0.16\linewidth}\centering 
            \caption*{$A$}
            \includegraphics[width=\linewidth]{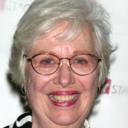}
            $\boldsymbol{a} =[1, 0, 1, 0]$
        \end{subfigure}
        \begin{subfigure}[t]{0.16\linewidth}\centering 
            \caption*{$B$}
            \includegraphics[width=\linewidth]{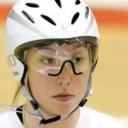}
            $\boldsymbol{b} = [0,0,1,1]$
        \end{subfigure}
        \begin{subfigure}[t]{0.16\linewidth}\centering 
            \caption*{$S$}
            \includegraphics[width=\linewidth]{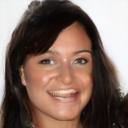}
            $\boldsymbol{s} = [1,0,0,1]$
        \end{subfigure}
        \hfill 
        \begin{subfigure}[t]{0.16\linewidth}\centering 
            \caption*{$A$}
            \includegraphics[width=\linewidth]{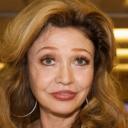}
            $\boldsymbol{a} =[0, 0, 0, 1]$
        \end{subfigure}
        \begin{subfigure}[t]{0.16\linewidth}\centering 
            \caption*{$B$}
            \includegraphics[width=\linewidth]{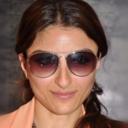}
            $\boldsymbol{b} = [0,0,1,0]$
        \end{subfigure}
        \begin{subfigure}[t]{0.16\linewidth}\centering 
            \caption*{$S$}
            \includegraphics[width=\linewidth]{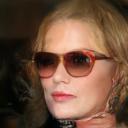}
            $\boldsymbol{s} = [0,0,1,0]$
        \end{subfigure}
        \caption{Examples from RDM\textsuperscript{\texttt{[L]}} with the \textit{clue} based abstraction conditioning approach.}
    \end{subfigure}
    \caption{\label{fig:logicface-generations}Examples of generated \LogicFace solutions shown in $S$, when conditioned on $A$ and $B$.}
\end{figure}

Tables \ref{tab:ap-akari-results}--\ref{tab:ap-logicface-results} present detailed results for our reasoning benchmark with different sampling steps ($\tau\in\{10,100\}$) in \Akari, \Tildoku, \Tangram and \LogicFace. We exclude sampling steps $\tau=100$ from \Tildoku since we did not appreciate significant changes from $\tau=10$; and SRM performance in \Tangram, since performance was significantly degraded in Table \ref{tab:tangram-logicface-results}. For \Akari, we also report the number of unlit cells (\textsc{unl}), the absolute error between the adjacent bulbs and annotated walls (\textsc{dst}), and the overlap (\textsc{ovp}, bulbs that mirror each other). 
We provide visualization of results obtained with our models on \Tangram and \LogicFace datasets in Figures~\ref{fig:tangram-generations} and~\ref{fig:logicface-generations}.

\begin{table}[ht!]\footnotesize
    \caption{\label{tab:ap-akari-results}Detailed results on \Akari.}\vspace{0.1cm}
    \setlength{\tabcolsep}{2pt}
    \renewcommand{\arraystretch}{1}
    \resizebox{\linewidth}{!}{
    \begin{tabular}{lllccccccccccccccc}
        \toprule
        &&& \multicolumn{5}{c}{\multirow{2}{*}{ID}} & \multicolumn{10}{c}{OoD}\\ 
        \cmidrule(l{.5em}r{.5em}){9-18}
        & &&&&& &&  \multicolumn{5}{c}{\textit{5-10\% (low)}} & \multicolumn{5}{c}{\textit{35-40\% (high)}}\\
        \cmidrule(l{.5em}r{.5em}){4-8}\cmidrule(l{.5em}r{.5em}){9-13}\cmidrule(l{.5em}r{.5em}){14-18}
         \multicolumn{3}{l}{} & \textsc{unl} & \textsc{dst} & \textsc{ovp} & \textsc{blb} &\underline{\textsc{acc}}  & \textsc{unl} & \textsc{dst} & \textsc{ovp} & \textsc{blb} &\underline{\textsc{acc}}  & \textsc{unl} & \textsc{dst} & \textsc{ovp} & \textsc{blb} &\underline{\textsc{acc}} \\ 
         \midrule
        \parbox[t]{4pt}{\multirow{11}{*}{\rotatebox[origin=c]{90}{\scriptsize $\tau=10$}}}
        & \multicolumn{2}{l}{Base} & 0.46 & 0.10 & 0.03 & 74.08 & 67.35 & 3.22 & 0.22 & 0.48 & 17.02 & 13.80 & 0.17 & 0.04 & 0.01 & 89.78 & 86.80\\
        \cmidrule(l{.5em}r{.5em}){2-18}
        & \multicolumn{2}{l}{SRM} \\ 
        & \multicolumn{2}{l}{\textit{parallel}} & 2.38 & 0.81 & 0.15 & 26.52 & 12.26 & 15.43 & 0.63 & 0.77 & 1.04 & 0.54 & 1.27 & 0.63 & 0.06 & 45.42 & 27.22 \\ 
        & \multicolumn{2}{l}{\textit{sequential}} & 7.01 & 9.81 & 17.79 & 0.00 & 0.00 &  0.26 & 4.64 & 87.15 & 0.00 & 0.00 & 13.87 & 13.47 & 6.28 & 0.00 & 0.00 \\
        \cmidrule(l{.5em}r{.5em}){2-18}
         &\multicolumn{2}{l}{RDM\textsuperscript{\texttt{[S]}}} \\ 
        & \parbox[t]{4pt}{\multirow{3}{*}{\rotatebox[origin=c]{90}{\scriptsize \textit{16 slots}}}}
         & \textit{clue}  & 0.40 & 0.10 & 0.01 & 77.14 & 70.20 & 0.54 & 0.29 & 0.22 & 55.48 & 42.30 & 0.16 & 0.04 & 0.01 & 90.92 & 87.88\\
        && \textit{solution}& 0.45 & 0.11 & 0.02 & 75.32 & 67.90 & 0.68 & 0.24 & 0.15 & 54.28 & 42.58 & 0.17 & 0.07 & 0.00 & 90.02 & 84.62\\
        && \textit{full} & 0.30 & 0.13 & 0.04 & 79.14 & 69.98 & 0.45 & 0.21 & 0.41 & 50.06 & 41.32 & 0.12 & 0.07 & 0.00 & 92.64 & 87.02\\

        & 
        \parbox[t]{4pt}{\multirow{3}{*}{\rotatebox[origin=c]{90}{\scriptsize \textit{121 slots}}}}
         & \textit{clue}   & 0.42 & 0.09 & 0.02 & 77.73 & 72.17 & 1.69 & 0.59 & 0.09 & 42.57 & 24.96 & 0.15 & 0.04 & 0.01& 91.00 & 87.94\\
         && \textit{solution} & 0.52 & 0.13 & 0.02 & 71.70 & 63.02 & 1.19 & 0.22 & 0.14 & 42.74 & 34.16 & 0.29 & 0.07 & 0.01 & 85.16 & 80.78\\
         && \textit{full}&  0.61 & 0.17 & 0.04 & 66.49 & 57.44 & 1.77 & 0.71 & 0.13 & 35.00 & 18.05 & 0.25 & 0.08 & 0.01 & 84.68 & 79.01 \\
        \midrule
        \parbox[t]{4pt}{\multirow{11}{*}{\rotatebox[origin=c]{90}{\scriptsize $\tau=100$}}}
        & \multicolumn{2}{l}{Base} & 0.10 & 0.08 & 0.01 & 86.02 & 80.04 & 1.61 & 0.11 & 0.16 & 43.62 & 39.48 & 0.10 & 0.04 & 0.01 & 92.22 & 88.92 \\
        \cmidrule(l{.5em}r{.5em}){2-18}
        & \multicolumn{2}{l}{SRM} \\ 
        & \multicolumn{2}{l}{\textit{parallel}} & 1.86 & 0.68 & 0.11 & 35.82 & 18.80  & 15.43 & 0.57 & 0.40 & 1.60 & 0.94 & 0.98 & 0.58 & 0.06 & 54.18 & 33.34\\ 
        & \multicolumn{2}{l}{\textit{sequential}$^*$} & 6.53 & 9.66 & 16.89 & 0.00 & 0.00 & -- &--  &-- &-- &-- & --&-- & -- & -- &--  \\
        \cmidrule(l{.5em}r{.5em}){2-18}
         &\multicolumn{2}{l}{RDM\textsuperscript{\texttt{[S]}}} \\ 
        & \parbox[t]{4pt}{\multirow{3}{*}{\rotatebox[origin=c]{90}{\scriptsize \textit{16 slots}}}}
         & \textit{clue}  & 0.14 & 0.08 &  0.01 & 88.30 & 82.06 & 0.15 & 0.18 & 0.02 & 85.98 & 72.48 & 0.09 & 0.05 & 0.01 & 92.90 & 88.94\\
        && \textit{solution}& 0.18 & 0.09 &  0.01 & 86.10 & 78.68 & 0.19 & 0.15 & 0.03 & 82.20 & 70.73 & 0.09 & 0.07 & 0.01 & 93.33 & 87.96 \\ 
        && \textit{full} & 0.12 & 0.10 & 0.01 & 89.44 & 80.42 & 0.13 & 0.11 & 0.08 & 83.32 & 75.06 & 0.07 & 0.08 & 0.01 & 93.80 & 87.48\\
        & \parbox[t]{4pt}{\multirow{3}{*}{\rotatebox[origin=c]{90}{\scriptsize \textit{121 slots}}}}
         & \textit{clue} &0.15& 0.07 &0.00 & 88.88 & 83.12 & 0.26 & 0.49 & 0.01 & 83.10 & 54.58 & 0.08 & 0.05 & 0.00 & 93.62 & 89.56 \\
         && \textit{solution} & 0.17 & 0.08 & 0.01 & 86.36 & 80.28 & 0.33 & 0.13 & 0.02 & 75.34 & 74.58 &  0.12 & 0.00 & 0.00 & 91.14 & 88.94 \\
         && \textit{full}& 0.18 &0.09 & 0.01 & 85.84 &  77.34 & 0.27 & 0.34 & 0.04 & 77.30 & 56.20 & 0.10 & 0.07 & 0.01 & 98.76 & 87.40\\
         \bottomrule
    \end{tabular}}
    \\
    $^*$We do not report results of SRM (\textit{sequential}) with $\tau=100$ in \textit{low} and \textit{high} regimes due to already low performance in ID evaluation and the high computational overhead of inference.
\end{table}

\begin{table}[ht]\footnotesize\centering
    \caption{\label{tab:ap-tildoku-results}Detailed results on the \Tildoku dataset.}\vspace{0.1cm}
    \setlength{\tabcolsep}{3pt}
    \renewcommand{\arraystretch}{1}
    \begin{tabular}{lccccccccc}
        \toprule
         \multicolumn{1}{c}{\multirow{2}{*}{\textsc{acc}} }&  \multicolumn{3}{c}{\textit{Easy (50-60)}} & \multicolumn{3}{c}{\textit{Medium (40-50)}} & \multicolumn{3}{c}{\textit{Hard (20-30)}}\\
         & \textsc{dgt}  &  \textsc{clr} & \underline{\textsc{acc}} & \textsc{dgt}  &  \textsc{clr} & \underline{\textsc{acc}} & \textsc{dgt}  &  \textsc{clr}  & \underline{\textsc{acc}} \\
         \midrule
         Base & 59.20 & 66.84 & 40.28 & 22.62 & 45.92 & 12.48 & 1.50 & 9.42 & 0.24 \\ 
         SRM (\textit{seq.}) & 44.87 & 25.87 & 13.80 & 24.31 & 21.31 & 7.08 & 3.62 & 5.62 & 0.62\\
         \midrule 
         RDM\texttt{[L]} \\ 
         \textit{clue} & 62.52 & 79.74 & 50.28 & 33.02 & 52.84 & 19.18 & 3.78 & 10.12 & 0.64\\ 
         \textit{solution} & 65.72 & 80.92 & 53.26 & 43.70 & 60.12 & 28.40 & 8.34 & 14.48 & 1.92\\
         \textit{full} & 55.56 & 69.52 & 39.30 & 22.24 & 31.96 & 8.46 & 1.66 & 2.80 & 0.14\\
         \bottomrule
    \end{tabular}
\end{table}

\begin{table}[ht!]\footnotesize\centering 
    \caption{\label{tab:ap-tangram-results}Detailed results on \Tangram.}\vspace{0.1cm}
    \setlength{\tabcolsep}{2pt}
    \begin{tabular}{llccccccc|ccccccc}
        \toprule 
        && \multicolumn{7}{c|}{Configuration \texttt{[S]}} & 
        \multicolumn{7}{c}{Configuration \texttt{[L]}} \\
        \multicolumn{2}{c}{$(^*\uparrow)$}& \parallelogramm & $\square$ & $\triangle^S$ & $\triangle^M$ & $\triangle^L$ & \textsc{shape} & \underline{\textsc{acc}}& \parallelogramm & $\square$ & $\triangle^S$ & $\triangle^M$ & $\triangle^L$ & \textsc{shape} & \underline{\textsc{acc}} \\
        \midrule 
         \parbox[t]{2pt}{\multirow{5}{*}{\rotatebox[origin=c]{90}{\scriptsize $\tau=10$}}} 
         & Base & 17.96 & 17.82 & 29.78 & 34.98 & 39.33 & 40.09 & 12.37 & 16.08 & 15.24 & 21.78 & 24.64 & 28.08 & 28.64 & 11.46 \\
         \cmidrule(l{.5em}r{.5em}){2-16}
         & RDM \\
         &\textit{clue} & 24.84 & 25.37 & 37.55 & 40.71 & 41.90 & 42.41  &22.20 & 34.72 & 34.26 & 41.80 & 43.34 & 44.88 & 44.98 & 31.96 \\
         & \textit{solution} & 31.43 & 31.80 & 36.73 & 38.69 & 40.06 & 40.51 & 28.45 & 33.34 & 32.66 & 39.90 & 41.64 & 43.16 & 43.52 & 29.94\\
         & \textit{full} & 34.49 & 34.69 & 38.65 & 42.67 & 43.82 & 4.31 & 30.18 & 22.20 & 21.38 & 33.26 & 36.28 & 40.48 & 41.44 & 16.74\\ 
        \midrule 
         \parbox[t]{2pt}{\multirow{5}{*}{\rotatebox[origin=c]{90}{\scriptsize $\tau=100$}}} 
         & Base & 23.06 & 24.22 & 42.82 & 41.50 & 55.60 & 57.72 & 15.46  & 24.74 & 23.06 & 40.34 & 38.62 & 47.74 & 48.02 & 19.00 \\
         \cmidrule(l{.5em}r{.5em}){2-16}
         & RDM \\
         &\textit{clue} & 36.71 & 36.96 & 59.59 &  60.33 &  63.29  & 58.41 & 34.37 & 42.62 & 41.54 & 60.78 & 59.46 & 63.02 & 63.04 & 40.04\\ 
         & \textit{solution} & 41.12 &  43.94 & 56.49 & 53.06 & 59.94& 60.25& 38.57 & 42.18 & 40.18 & 40.44 & 60.04 & 58.50 & 62.74 & 38.84  \\ 
         & \textit{full} & 49.75 & 49.98 & 61.55 & 61.25 & 65.00 & 65.12 & 46.76  & 31.44 & 30.32 & 53.18 & 52.34 & 61.54 & 61.96 & 25.00\\ 
         \bottomrule
    \end{tabular}
\end{table}

\begin{table}[ht]\footnotesize\centering
    \caption{\label{tab:ap-logicface-results}Detailed results on \LogicFace.}\vspace{0.1cm}
    \setlength{\tabcolsep}{2pt}
    \renewcommand{\arraystretch}{1}
    \resizebox{\linewidth}{!}{
    \begin{tabular}{llccccccc|ccccccc}
        \toprule
        \multicolumn{2}{c}{\multirow{2}{*}{$(^*\uparrow)$}} &  \multicolumn{7}{c|}{Configuration \texttt{[S]}} & \multicolumn{7}{c}{Configuration \texttt{[L]}}\\
        & &  \makecell[c]{\textit{smiling}\\ (\texttt{OR})} & \makecell[c]{\textit{male}\\ (\texttt{AND})} & \makecell[c]{\textit{glasses}\\ (\texttt{XOR})} & \makecell[c]{\textit{young} \\ (\texttt{IMP.})} & \textsc{avg} & \underline{\textsc{acc}} & \textsc{fid}&\makecell[c]{\textit{smiling}\\ (\texttt{OR})} & \makecell[c]{\textit{male}\\ (\texttt{AND})} & \makecell[c]{\textit{glasses}\\ (\texttt{XOR})} & \makecell[c]{\textit{young} \\ (\texttt{IMP.})} & \textsc{avg} &  \underline{\textsc{acc}} & \textsc{fid}\\ 
        \midrule 
        \parbox[t]{3pt}{\multirow{5}{*}{\rotatebox[origin=c]{90}{\scriptsize $\tau=10$}}} &
        Base & 71.28& 79.39& 60.77& 73.50 & 71.24& 26.03 & 44.95 & 55.08 & 55.42 & 39.24 & 70.28 & 54.51 & 8.62 & 35.92\\
        \cmidrule(l{.5em}r{.5em}){2-16}
        & RDM \\
        & \textit{clue} & 87.74 & 96.66 & 93.61 & 81.67 & 89.92 & 66.08 & 37.66 & 83.34 & 94.52 & 49.60 & 70.02 & 74.37 & 27.98 & 37.18 \\
        & \textit{solution} & 68.53 & 78.34 & 49.81 & 72.10 & 67.19 & 18.98& 50.21&65.41 & 66.44 & 49.41 & 66.05 & 61.83 & 14.95 & 39.71\\
        & \textit{full} & 84.97 & 94.74 & 90.81 & 78.55 & 87.27 & 59.57 & 36.37  & 78.38 & 91.74 & 50.68 & 65.66 & 71.61 & 25.08 & 38.04 \\ 
        \midrule 
        \parbox[t]{3pt}{\multirow{8}{*}{\rotatebox[origin=c]{90}{\scriptsize $\tau=100$}}}
        & Base & 68.18 & 76.31 & 64.86 & 71.09 & 70.11 & 25.13 & 21.00 & 54.50 & 52.06 & 39.70 & 69.52 & 53.94 & 8.36 & 33.08 \\
        \cmidrule(l{.5em}r{.5em}){2-16}
        & SRM \\
        & \textit{parallel} & 37.00 & 38.16 & 49.26 & 39.44 & 40.97 & 2.60 & 109.48 & -- & -- & -- & -- & -- & -- & -- \\
        & \textit{sequential} &  25.04 & 58.38 & 50.58 & 73.86 & 51.94 & 5.70 & 191.03 & -- & -- & -- & -- & -- & -- & -- \\ 
        \cmidrule(l{.5em}r{.5em}){2-16}
        & RDM \\
        & \textit{clue} & 90.19 & 97.00 & 97.26 & 87.27 & 92.93 & 74.74 & 13.34 &  82.50 & 94.30 & 49.74 & 66.58 & 73.28 & 26.60 & 35.70\\
        & \textit{solution} & 64.87 & 73.20 & 50.16 & 67.78 & 64.00 & 16.27 & 19.14 &  65.32 & 66.16 & 49.38 & 66.02 & 61.72 & 14.98 & 37.19 \\
        & \textit{full} & 83.16 & 94.00 & 94.76 & 79.81 & 87.93 & 60.76 & 14.72 &  82.16 & 92.12 & 50.00 & 68.84 & 73.23 & 25.66 & 36.03 \\ 
     \bottomrule 
    \end{tabular}}
\end{table}

\textbf{Discussion on \Tangram complexity:} 
The majority of existing generative modeling literature focuses on Sudoku-style reasoning constraints for diffusion models~\citep{wever-etal-2025-spatial,avdeyev-etal-2023-dirichlet,lee-etal-2025-adaptive}.
These works typically frame the task as an inpainting problem, where the model learns to satisfy localized relational rules within a fixed coordinate system.
In our \Tangram benchmark, generating valid solutions represents a significantly more complex challenge with fewer precedents in deep learning.
Prior work, such as TANGAN~\citep{yamada-etal-2025-tangan}, used adversarial networks to partition silhouettes into geometric components.
However, our benchmark requires the model to go beyond structural partitioning, and requires simultaneously perform geometric induction, maintain color-consistency across dynamic pieces and infer latent spatial transformations to align tans with a target configuration.
The low baseline accuracy reported in Table \ref{tab:tangram-logicface-results} clearly reflects this increased dimensionality.

We conducted preliminary experiments to better understand the complexity of \Tangram. 
We designed two intermediate level benchmarks to isolate the challenges introduced in the main task (geometric partitioning and color consistency).
The first level resembles the TANGAN approach~\citep{yamada-etal-2025-tangan}, where the model is tasked with generating a valid partition of a given silhouette.
We evaluated both a standard convolutional autoencoder (AE) and our baseline diffusion model on this task, using the silhouette as the sole conditional input and using solutions where a unique color was always associate to the same tan.
Under this setup, the AE achieved only 1.93\% of accuracy, whereas the diffusion model reached 82\%.
These results show that while traditional reconstruction-based architectures struggle with geometric arrangements, diffusion models are highly capable of learning complex partitions. 

In our second level, we introduced the additional challenge of chromatic modeling.
Here, the model is provided with the target silhouette alongside \textit{a fixed piece view} of the tans, where only the colors vary across samples.
The diffusion model is tasked with generating a valid partition of the silhouette, while matching the specific colors provided in the tan view.
Here, the model only needs to learn the fixed proportions of the pieces and the mapping of their associated colors.
We consider this level simpler than our \Tangram task, since the tan view does not remain fixed across samples in our task. 

Under this configuration, the diffusion model's partitioning accuracy dropped to 54\%, with a total accuracy of 52\%.
Accordingly, we infer that the key challenge in \Tangram appears when different color permutations are associated with the same silhouette partition. 
This suggests that the model struggles to establish a robust cross-modal correspondence between the conditional input and the denoising target. 
Specifically, the increased difficulty stems from the model's inability to consistently correlate the arbitrary colors in the tan view with the final spatial arrangement of the pieces.

\subsubsection{Additional Experiments}\label{ap:additional_experiments}

We present further experimental analyses by: (1) isolating the impact of relational abstractions via an ablation study where we evaluate the reasoning capabilities of purely slot-conditioned diffusion guidance, (2) investigating empirical evidence of symbolic rule learning by the relational abstractions module, and (3) evaluations on previous spatial reasoning benchmarks MNIST-Sudoku and Counting Polygons~\citep{wever-etal-2025-spatial} for completeness.

\textbf{Impact of relational abstractions:}
We demonstrate that diffusion is not only compatible with relational abstractions over object-centric embeddings, but also benefits from these contextual representations in solving reasoning problems.
Previous works by~\citet{wu-etal-2023-slotdiffusion} and~\citet{jiang2023object} have proposed slot-conditioned diffusion models for image synthesis, although these were not adapted to CFG or object disentanglement for higher-level abstraction and reasoning.
To isolate the impact of \textit{relational abstractions}, we also benchmarked a purely \textit{slot-based} conditioned diffusion model, i.e., $p_{\theta}(\mathbf{x}_0|\mathbf{x}_c,\mathbf{s}^c_{1:k})$ at training and inference.
In Table~\ref{tab:slot-diffusion-comparison}, we demonstrate an interesting result which was not investigated in previous works, that purely slot conditioning alone can
\begin{wraptable}{r}{0.49\linewidth}\footnotesize 
    \vspace{-5pt}
    \setlength{\tabcolsep}{4pt}
    \caption{\label{tab:slot-diffusion-comparison}RDM vs slot-based diffusion.}\vspace{-.2cm}
    \begin{tabular}{lcccc}
        \toprule 
        \multicolumn{1}{c}{\multirow{2}{*}{\textsc{acc}}}& \multicolumn{3}{c}{\Coldoku ($\tau=10$)}  & \Tangram \\ 
        & \textit{Easy} & \textit{Medium} & \textit{Hard} & ($\tau=100$) \\
        \midrule 
        Base & 90.84 & 48.72 & 3.43 & 15.46\\
        \textit{Slot-based} &  93.76 & 54.94 & 4.74 & 27.09 \\ 
        RDM$^*$ (\textit{clue}) & \bf 96.66 & \bf 65.62 & \bf 7.76 & \bf 34.37 \\ 
        \bottomrule 
    \end{tabular}
    $^*$RDM\textsuperscript{\texttt{[S]}} for \Tangram, RDM\textsuperscript{\texttt{[L]}} for \Coldoku.
    \vspace{-10pt}
\end{wraptable}
already improve reasoning upon a standard baseline.
Nevertheless, our results show that the slot-based approach still underperforms relative to RDM.
This performance gap is particularly pronounced in the \Tangram dataset, where RDM outperforms slot-conditioning (\textit{clue}: 34.37 vs 27.09).
These results suggest that explicit relational abstraction between objects and their spatial configurations is critical for complex reasoning.

\textbf{Empirical evidence of rule learning:}
Beyond our general evaluations of reasoning performance based on rule-consistent image synthesis, we further explore whether the learned embeddings by the relational abstraction modules truly capture the logical rules required for reasoning.
We investigate this aspect in three different experiments with RDM.
In our first experiment, we perform sampling from $p_\theta(\mathbf{x}_0|\mathbf{x}_c,\mathcal{R}(\mathbf{z}^0))$ with RDM (\textit{solution}, by enabling access to the relational solution embeddings at inference time.
Under this setting, accuracy boosted to 93\% in \Tangram, and 95\% in \Coldoku (\textit{hard}), evidencing that RDM learns to abstract representations to almost perfectly solve both puzzles.
In the second experiment, we used the \Tangram dataset and removed the raw visual clues $\mathbf{x}_c$ while training RDM to model $p_\theta(\mathbf{x}_0|\mathcal{R}(\mathbf{z}^c))$.
This approach yielded \textsc{acc}: 37.55 by surpassing again the Base model (15.46), and showing that the relational abstractions alone are powerful enough to replace the clue itself and improve reasoning.
In our third experiment, we conducted linear probing tests on \Coldoku.
We froze the abstractor stack and trained a linear classifier to distinguish the relational embeddings produced by 6,400 valid and 6,400 invalid puzzles, where invalid puzzles were artificially generated to violate \Coldoku constraints.
In the first setting, invalid puzzles were obtained by arranging the same digit in all columns.
In the second setting, the digit arrangements were randomly shuffled across the grid.
We trained simple linear classifiers on these high-dimensional ($k\times d$) embeddings, and achieved 89.98\% in the first setting, and 82.59\% in the second setting.
These results indicate that the learned relational representations encode information that makes valid versus rule-violating puzzles linearly separable.
Altogether, we treat these results as additional evidence that learned relational representations encode meaningful logical rule information for reasoning, and justify their effectiveness when guiding diffusion models.

\paragraph{Evaluations on previous benchmarks:}
Table~\ref{tab:previous-benchmarks} presents the evaluation   of  RDM on previous spatial reasoning benchmarks~\citep{wever-etal-2025-spatial}: MNIST-Sudoku (\textit{medium} and \textit{hard} regimes), and Counting Polygons. 
In MNIST-Sudoku we report the original metrics and the evaluation of SRM under our setup ($\tau=100$ and 5,000 test samples). 
In Counting Polygons, we report metrics based on the original neural evaluator and our unbiased neural evaluator\footnote{We observed that the original evaluator for Counting Polygons~\citep{wever-etal-2025-spatial} was biased to always predict a positive number of polygons and exactly two digits, even when no clear shape or digit was present, leading to optimistic evaluations. We therefore trained an \textit{unbiased} evaluator that handles cases when no digit or shape appears in the image, and report evaluations with both.}. 
With less than half of parameters, RDM outperforms SRM in the \textit{medium} and \textit{hard} regimes, while slightly surpasses the Base model in the \textit{medium} regime (55.80 vs 54.18). 
We noticed that diffusion forcing only benefits
\begin{wraptable}{r}{0.55\linewidth}\footnotesize
    \vspace{-6pt}
    \caption{\label{tab:previous-benchmarks}Results on previous benchmarks.}
    \setlength{\tabcolsep}{3pt}
    \begin{tabular}{lcc|cc} 
        \toprule
         & \multicolumn{2}{c|}{MNIST-Sudoku} & \multicolumn{2}{c}{Counting Polygons} \\
         &  \textit{Medium} & \textit{Hard} & \textit{Original} & \textit{Unbiased}\\
         \midrule 
         Base & 54.18 \reported{53.6} & \textbf{1.9} \reported{0.8}\phantom{0} & 13.2 & 9.8 \\ 
         \midrule 
         SRM \\
         \textit{parallel} & 31.04 \reported{59.0} & 0.00 \reported{1.0}\phantom{0} & 16.6 &  10.1 \\
         \textit{sequential} & 40.02 \reported{75.4} & 0.01 \reported{51.6} & 18.6 & \phantom{0}5.4 \\
         \midrule 
         \multicolumn{5}{l}{Non-object-centric features} \\
        DINOv2  & 44.26 & 0.18 & -- & -- \\
        ViT-MAE & 44.60 & 0.08 & -- & --\\
         \midrule 
         RDM\textsuperscript{\texttt{[M]}}\\ 
         \textit{clue} &47.34 & 0.22 & -- & --\\
         \textit{solution} & 47.14& 0.06 & \textbf{51.2} & \textbf{30.7} \\ 
         \textit{full} &\bf 55.80 & 0.76  & -- & --\\
         \bottomrule
    \end{tabular}
    Reported results from~\citet{wever-etal-2025-spatial} in parentheses.
    \vspace{-10pt}
\end{wraptable}
performance with a high number of sampling steps, as the accuracy gap disappears when sampling with 100 steps.
In contrast, RDM considerably outperforms SRM and the Base diffusion model in Counting Polygons (30.7 vs 10.1), where pure diffusion forcing (\textit{parallel}) and SRM \textit{sequential} no longer benefit the denoising process.

Finally, we conducted additional ablations on MNIST-Sudoku to demonstrate the benefits of object-centric representations for diffusion guidance.
We perform conditioning via non-object-centric visual features from DINOv2~\citep{oquab2024dinov} and ViT-MAE~\citep{he-etal-2021-mae}.
Table~\ref{tab:previous-benchmarks} shows these representations do not lead to substantial improvements in either the \textit{medium} or the \textit{hard} regime.
Overall, performance is still worse than the Base model and RDM, demonstrating that object-centric relational representations are superior to non-disentangled image features for reasoning.

\end{document}